\documentclass{article}

\usepackage{microtype}
\usepackage{graphicx}
\usepackage{subfigure}
\usepackage{booktabs} 

\usepackage{hyperref}

\usepackage{amsmath,amsthm,amssymb}
\usepackage{algorithm}
\usepackage{algpseudocode}

\usepackage[accepted]{icml2021}

\icmltitlerunning{Large-Scale Meta-Learning with Continual Trajectory Shifting}

\begin{document}

\twocolumn[
\icmltitle{Large-Scale Meta-Learning with Continual Trajectory Shifting}



\icmlsetsymbol{equal}{*}

\begin{icmlauthorlist}
\icmlauthor{JaeWoong Shin}{equal,kaist}
\icmlauthor{Hae Beom Lee}{equal,kaist}
\icmlauthor{Boqing Gong}{google,kaist}
\icmlauthor{Sung Ju Hwang}{kaist,aitrics}
\end{icmlauthorlist}

\icmlaffiliation{kaist}{Graduate School of AI, KAIST, South Korea}
\icmlaffiliation{google}{Google, LA}
\icmlaffiliation{aitrics}{AITRICS, South Korea}

\icmlcorrespondingauthor{Sung Ju Hwang}{sjhwang82@kaist.ac.kr}

\icmlkeywords{Meta-learning, Large-scale}

\vskip 0.3in
]



\printAffiliationsAndNotice{\icmlEqualContribution} 





\newtheorem{thm}{Theorem}[section]
\newtheorem{lem}{Lemma}[section]
\newtheorem{cor}{Corollary}

\newcommand{\dee}{\mathrm{d}}
\newcommand{\grad}{\nabla}

\newcommand{\normal}{\mathcal{N}}
\newcommand{\bbE}{\mathbb{E}}
\newcommand{\calL}{\mathcal{L}}
\newcommand{\calM}{\mathcal{M}}
\newcommand{\calD}{\mathcal{D}}
\newcommand{\card}[1]{\vert {#1} \vert}
\newcommand{\distiid}{\overset{iid}{\dist}}
\newcommand{\distind}{\overset{ind}{\dist}}
\newcommand{\Indicator}[1]{\mathds{1}_{\{#1\}}}
\newcommand{\II}{\mathbb{I}}
\newcommand{\defas}{\vcentcolon=}  
\newcommand{\iid}{i.i.d.}

\newcommand{\bs}[1]{{\boldsymbol{#1}}}
\newcommand{\bepsilon}{{\boldsymbol{\epsilon}}}
\newcommand{\btheta}{{\boldsymbol{\theta}}}
\newcommand{\bTheta}{{\boldsymbol{\Theta}}}
\newcommand{\bbeta}{{\boldsymbol{\beta}}}
\newcommand{\Real}{{\mathbb{R}}}
\newcommand{\calP}{{\mathcal{P}}}
\newtheorem{observation}{Observation}

\newcommand{\D}{\mathcal{D}}
\newcommand{\tr}{\text{tr}}
\newcommand{\va}{\text{va}}
\newcommand{\te}{\text{te}}
\newcommand{\bx}{\mathbf{x}}
\newcommand{\bz}{\mathbf{z}}
\newcommand{\by}{\mathbf{y}}
\newcommand{\ba}{\mathbf{a}}
\newcommand{\bX}{\mathbf{X}}
\newcommand{\bY}{\mathbf{Y}}

\newcommand{\bzero}{\mathbf{0}}
\newcommand{\bone}{\mathbf{1}}
\newcommand{\bb}{\mathbf{b}}
\newcommand{\bH}{\mathbf{H}}
\newcommand{\bof}{\mathbf{f}}
\newcommand{\bhy}{\widehat{\by}}
\newcommand{\bw}{\mathbf{w}}
\newcommand{\bI}{\mathbf{I}}
\newcommand{\be}{\mathbf{e}}
\newcommand{\bG}{\mathbf{G}}
\newcommand{\bh}{\mathbf{h}}
\newcommand{\bhW}{{\widehat{\bW}}}
\newcommand{\btW}{{\widetilde{\bW}}}
\newcommand{\bV}{\mathbf{V}}
\newcommand{\bv}{\mathbf{v}}
\newcommand{\bphi}{{\boldsymbol{\phi}}}
\newcommand{\bpi}{{\boldsymbol{\pi}}}
\newcommand{\noise}{{\boldsymbol{\varepsilon}}}

\newcommand{\ts}{{(t)}}

\newcommand{\bee}{\begin{eqnarray}}
\newcommand{\eee}{\end{eqnarray}}

\begin{abstract}
Meta-learning of shared initialization parameters has shown to be highly effective in solving few-shot learning tasks. However, extending the framework to many-shot scenarios, which may further enhance its practicality, has been relatively overlooked due to the technical difficulties of meta-learning over long chains of inner-gradient steps. In this paper, we first show that allowing the meta-learners to take a larger number of inner gradient steps better captures the structure of heterogeneous and large-scale task distributions, thus results in obtaining better initialization points. Further, in order to increase the frequency of meta-updates even with the excessively long inner-optimization trajectories, we propose to estimate the \emph{required shift} of the task-specific parameters with respect to the change of the initialization parameters. By doing so, we can arbitrarily increase the frequency of meta-updates and thus greatly improve the meta-level convergence as well as the quality of the learned initializations. We validate our method on a heterogeneous set of large-scale tasks and show that the algorithm largely outperforms the previous first-order meta-learning methods in terms of both generalization performance and convergence, as well as multi-task learning and fine-tuning baselines.

\end{abstract}

\section{Introduction}
\label{introduction}


Meta-learning~\cite{schmidhuber1987evolutionary,thrun98} is a framework for learning a learning process itself by extracting common knowledge over a task distribution. As this meta-knowledge allows task learners to adapt to newly given tasks in a sample efficient manner, meta-learning has frequently been used for solving few-shot learning problems where each of the task learners is given only a few training examples~\cite{lake2015human,vinyals2016matching,santoro2016meta,snell2017prototypical,finn2017model}. 
While there exists a vast literature on meta-learning methods that tackle few-shot learning, one of the most popular approaches is the optimization-based method such as Model Agnostic Meta-Learning (MAML) \cite{finn2017model}, which aims to improve the generalization ability of few-shot learners by learning good initialization parameters, from which the model can rapidly adapt to novel tasks within only a few gradient steps.

Then, a natural question is if the same meta-learning strategy is applicable to tasks with a larger number of examples, for instance STL10~\cite{coates2011analysis} and Stanford Cars~\cite{stanford_cars}. It is well known that such standard learning tasks with a large number of training examples also benefit from good initialization parameters for better convergence and generalization, when compared with random initializations~\cite{kornblith2019do}. A prevalent approach to enhance generalization for large-scale tasks is to pre-train the model with a large dataset such as ImageNet~\cite{russakovsky2015imagenet}, and further finetune the pretrained model parameters with the target dataset. This demonstrates that knowledge transfer is also highly beneficial for tasks with larger training sets.


\begin{figure*}[t]
    \vspace{-0.1in}
	\centering
	\hfill
	\subfigure[Previous meta-learning]{
	\includegraphics[height=2.9cm]{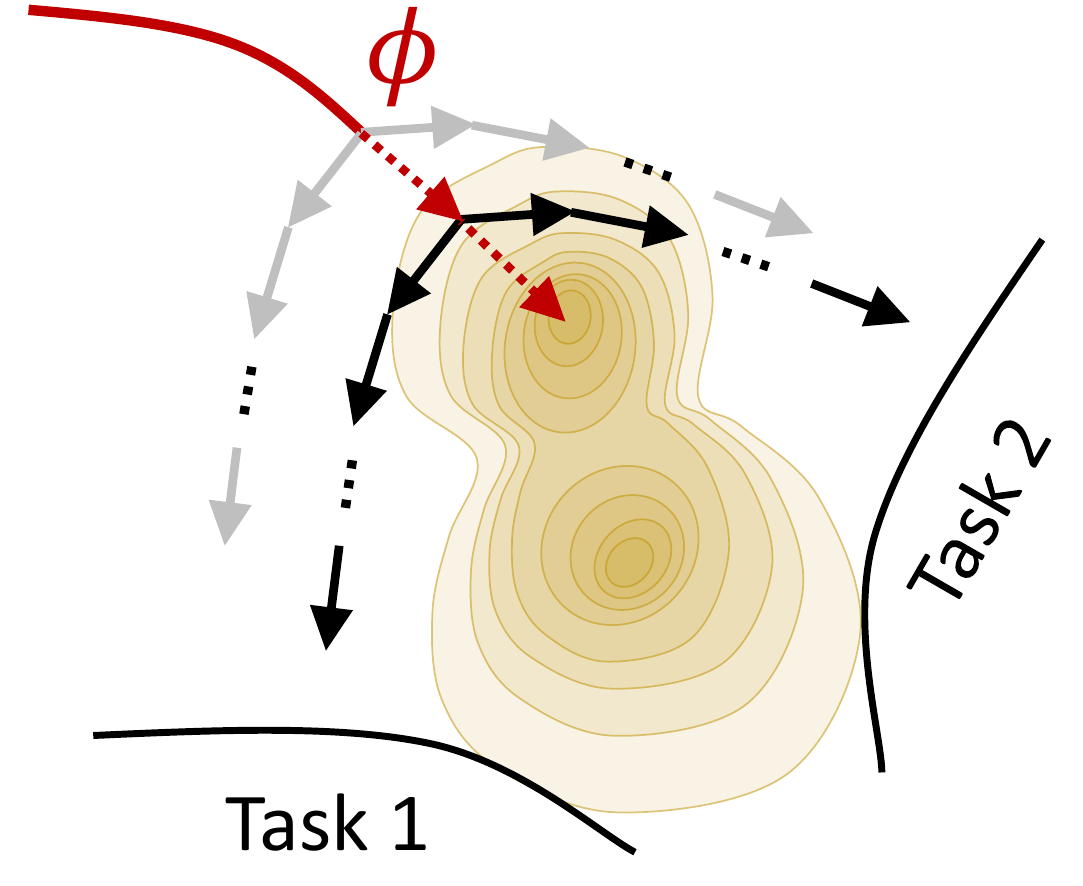}
	\label{fig:concept_prev}
	}
	\hfill
	\subfigure[Meta-learning with Continual Trajectory Shifting]{
	\includegraphics[height=2.9cm]{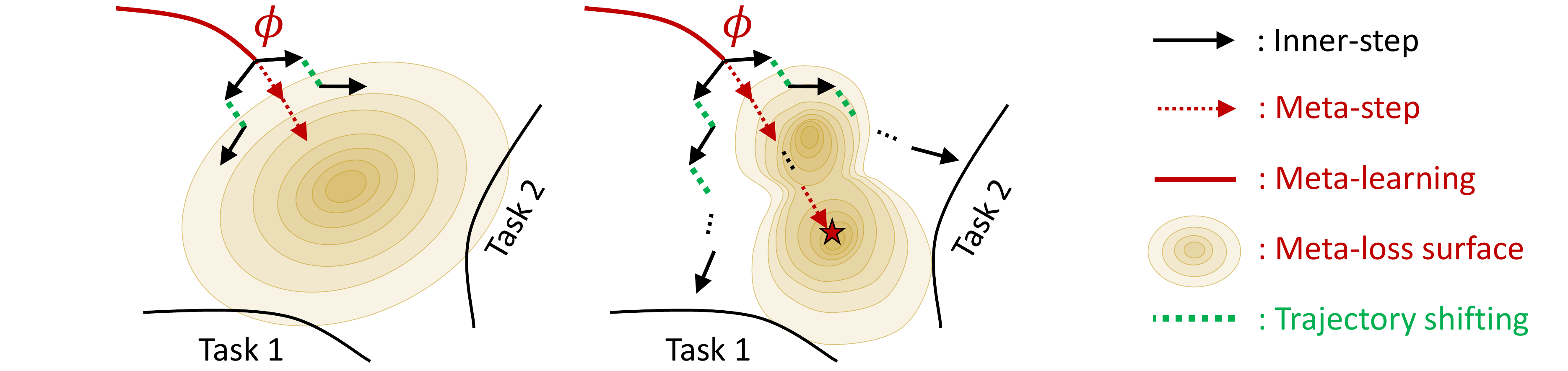}
	\label{fig:concept_ours}
	}
	\vspace{-0.13in}
	\caption{\small \textbf{Concepts of large-scale meta-learning,} whose inner learning trajectories have to be long enough to fit each large-scale individual task. (a) Previous meta-learning which is vulnerable to bad meta-level local minima and waits for the excessively long inner learning trajectories between two meta-updates. (b) Our method that performs frequent meta-updates by interleaving them with the inner-learning trajectories, plus continual trajectory shiftings, and is less prone to bad local minima by gradually growing the trajectory length $k$. }
	\vspace{-0.15in}
	\label{fig:concept}
\end{figure*}

However, the meta-learning of shared initialization parameters for many-shot learning problems has not received much attention. One reason may be that ImageNet pretraining has been practically effective for most of the standard object classification tasks or other computer vision problems. However, \citet{kornblith2019do} empirically show that ImageNet pretraining may not obtain meaningful performance gains on fine-grained classification tasks. This is because fine-grained classification tasks may require features that are more domain-specific or local for the discrimination of highly similar visual classes, for which the ImageNet features learned for general object classification may be ineffective. In other words, pretraining neural networks only on a single large-scale dataset will not sufficiently cover the heterogeneity of datasets and tasks that the model needs to handle at inference time. One of the effective ways to handle such heterogeneity is to train the model via meta-learning over a heterogeneous task distribution.

There have been several attempts to apply meta-learning to large-scale settings, where the training set consists of a large number of instances~\cite{reptile,leap,Flennerhag2020Meta-Learning}.
While these methods alleviate the computational cost for large-scale meta-learning, they are not truly scalable to tasks that are considered in conventional learning scenarios. The main difficulty of meta-learning with shared initialization for large-scale tasks, is that they require a large number of gradient steps to converge, since otherwise the meta-learner would suffer from the short horizon bias problem~\cite{wu2018understanding}. Note that the computational cost of a single meta update increases linearly with respect to the number of inner gradient steps. Therefore, a single meta-gradient update for gradient-based meta-learning algorithms (e.g. MAML, Reptile~\cite{reptile}) would require probably thousands of subsequent inner-gradient steps for the given tasks (See Figure~\ref{fig:concept_prev}).

The key to this challenging problem of large-scale meta-learning, is how to perform frequent meta-updates for meta-convergence while allowing the learning trajectories of inner optimization problems to become sufficiently long. However, due to the strong dependency between the initialization parameters and learning trajectories for each task, naively updating the initialization parameters without correcting the learning trajectories may be suboptimal. We tackle this by proposing a novel idea: estimating the corresponding change of the task-specific parameters with respect to the change of the initialization point. If  we can estimate such an update direction with a reasonable accuracy, then we will be able to arbitrarily increase the frequency of meta updates with the corresponding shifting for the task learning trajectories (See Figure~\ref{fig:concept_ours}).

In this paper, we show that first-order Taylor expansion together with the first-order approximation of Jacobian over the learning trajectories~\cite{finn2017model,reptile,leap} yields a surprisingly simple but effective shifting rule, shifting the entire learning trajectories to the direction and the amount for each meta update. By doing so, we can perform more frequent meta-updates compared to existing optimization-based meta-learning algorithms, while preserving the connection between the initialization point and the task learning trajectories up to the approximation error.
Our method enjoys significantly faster convergence over existing first-order meta-learning algorithms, and the learned initialization by our method leads to better generalization performance as well.

We validate our method by meta-training over a heterogeneous set of standard, many-shot learning tasks such as Aircraft~\cite{maji2013fine}, CUB~\cite{WahCUB_200_2011}, and Fashion MNIST~\cite{xiao2017} that require at least a thousand of gradient steps for an accurate estimation of meta gradients. We then meta-test the learned initial model parameters by finetuning with a similar set of diverse datasets such as Stanford Cars~\cite{stanford_cars} and STL10~\cite{coates2011analysis}. 

We summarize our contributions as follows:
\vspace{-0.1in}
\begin{itemize}
\vspace{-0.025in}
    \item We show that large-scale meta-learning requires substantially a larger number of inner gradient steps than what are reqruied for few-shot learning.
    \vspace{-0.025in}
    \item We show that gradually extending the length of inner-learning trajectories lowers the risk of converging to poor meta-level local optima.
    \vspace{-0.025in}
    \item To this end, we propose a novel and an efficient algorithm for large-scale meta-learning that frequently performs meta-optimization even with excessively long inner-learning trajectories. 
    \vspace{-0.025in}
    \item We verify our algorithm on a heterogeneous set of tasks, 
    on which it achieves significant improvements over existing meta-learning algorithms in terms of meta-convergence and generalization performance.
    \vspace{-0.025in}
\end{itemize}

\section{Related Work}
\paragraph{Meta-learning}
Meta-learning~\cite{schmidhuber1987evolutionary,thrun98} aims to learn how to learn on novel tasks, without overfitting to seen tasks. Meta-learning is usually done by assuming a task distribution~\cite{vinyals2016matching,ravi2016optimization} from which tasks are sampled and a meta-learner which solves them by extracting common meta-knowledge among the given tasks. Many recent works have demonstrated the effectiveness of such a strategy in few-shot learning settings where the learner should adapt to novel tasks with few training samples for each task~\cite{lee2018gradient,mishra2018a,rusu2018meta,liu2018learning,Lee_2019_CVPR}. A popular approach to meta-learning is to learn a common metric space over a task distribution~\cite{vinyals2016matching,snell2017prototypical,yang2017learning,oreshkin18}, that can be used for the prediction for a novel task. For classification, the space could be learned to map each instance (query instance) closer to either another instance from the same class, or the prototype of the class. However, 
in many-shot scenario we target, it is not trivial to fully exploit the task information without taking a sufficient number of gradient steps. Therefore, we focus more on optimization-based meta-learning methods~\cite{finn2017model} that are model-agnostic, whose goal is to learn a shared initialization parameters from which each of the target tasks can adapt after taking some amount of gradient steps. The shared initialization parameters are meta-learned by backpropagating through the learning trajectories. 



\vspace{-0.07in}
\paragraph{Efficient Meta-learning} Early optimization-based meta-learning algorithms usually require computing the second-order derivatives in order to obtain the meta-gradients~\cite{finn2017model}. Yet, due to the heavy computational cost in computing them, many prior works propose to use a first order approximation to obtain the meta-gradient~\cite{finn2017model,reptile,leap}, based on the empirical observation that given a sufficiently small stepsize for the inner-gradient steps, curvature information around a local region, i.e. Hessian, can be safely ignored. Other ways to efficiently compute meta-gradients include~\citet{rajeswaran2019meta} and \citet{Song2020ES-MAML}. However, despite their computational efficiency, none of the existing gradient-based meta-learning methods are truly scalable to large-scale meta-learning that involves a large number of inner-gradient steps, since this will slow down the meta-convergence as the meta-update frequency will decrease as the trajectory lengths for the inner-gradient steps increase. In this paper, we propose a novel algorithm for effectively \emph{increase the frequency of meta-updates} while preserving the connection between the shared initialization and task learning trajectories.

\vspace{-0.07in}
\paragraph{Transfer and multi-task learning} It is possible to use transfer learning as an alternative of meta-learning for large-scale sceanarios to avoid the excessive computational cost associated with it. Specifically, finetuning from a pretrained network on a large dataset such as ImageNet~\cite{russakovsky2015imagenet} is a simple yet an effective method that is known to perform well in practice. \citet{Dhillon2020A} recently showed that a simple variant of finetuning strategy in transductive setting outperforms most of the current sophisticated meta-learning algorithms. Yet, finetuning strategies are based on a strong assumption that a feature extractor learned from a single big dataset is beneficial in boosting the generalization performance of the target datasets, which may not hold when the target tasks have largely different distributions from the source task~\cite{kornblith2019do}. While more sophisticated transfer learning or domain adaptation methods can tackle this problem~\cite{Jang2019LearningWA}, it remains an important question whether we can learn intialization parameters that can generalize well even to tasks with large distributional shifts, such as fine-grained classification. Also, although there exist abundant datasets that may contribute to meta-knowledge, it is not trivial to decide which datasets to use for pretraining. Multi-task learning (MTL) is an effective way to achieve generalization across tasks. However a naive MTL approach with joint training of multiple tasks is vulnerable to negative transfer under a heterogeneous task distribution, degrading the quality of the learned feature extractor that will be used for finetuning the target tasks. Optimization based meta-learning can be a natural solution to tackle the negative transfer problem, since it finds an intialization point that can lead to optimal solutions for heterogeneous tasks, rather than trying to find a solution that are jointly optimal for all tasks.


\vspace{-0.05in}
\section{Approach}
We start by describing our problem setup. Our goal is to learn shared initialization parameters $\phi$ that can lead to good solutions for diverse tasks after task-specific adaptation. Suppose that we have $T$ tasks (or datasets) $\D^{(1)},\dots,\D^{(T)}$ that will be used for meta-training, and each of the tasks has a large number of training examples. We further assume that there exist substantial distributional discrepancies among the tasks. In this large-scale heterogeneous meta-learning setup, it is crucial for the task-specific model parameters $\theta^{(t)}$ to fully adapt to a given task $\D^{(t)}$ by taking a sufficient number ($K$) of gradient steps (e.g. $K=1,000$ steps) from the shared initialization $\phi$. 
We also let the inner-optimization processes repeat $M$ times in order to make the initialization parameters $\phi$ fully converge. As a result, we expect the meta-learned $\phi$ to work well on new tasks or datasets.

\vspace{-0.025in}
\subsection{Limitations of previous methods}
\vspace{-0.025in}
Naturally, backpropagating through a learning process requires to compute the second-order derivatives such as Hessians~\cite{finn2017model}. Due to its heavy computational cost, we focus on the first-order meta-learning algorithms such as FOMAML~\cite{finn2017model}, Reptile~\cite{reptile}, or Leap~\cite{leap} which are more suitable for large-scale meta-learning. We sketch the method in Algorithm~\ref{algo:previous}. For instance, the meta-gradient of FOMAML is $\mathsf{MetaGrad}(\phi;\theta_K^{(t)})=\grad_{\theta} \mathcal{L}_K^{(t)}|_{\theta=\theta_K^{(t)}}$ where $\mathcal{L}_k^{(t)}$ denotes loss of task $t$ at step $k$, and the Reptile gradient is $\mathsf{MetaGrad}(\phi;\theta_K^{(t)})=\phi-\theta_K^{(t)}$. They consist of only the first-order terms and thus are cheaper to compute than meta-gradients with the second-order derivatives.

However, the previous meta-learning methods have to reinitiate each inner-optimization process right after every single meta-update, because the inner-learning process should remain consistent with the updated initialization point and start a new learning trajectory from there (see Figure~\ref{fig:concept}). This makes the large-scale meta-learning inefficient. We see from Algorithm~\ref{algo:previous} that the interval between the current meta-update and the previous one (in red) linearly increases with respect to $K$, the total length of every inner-optimization trajectory.
For example, if we have $T=10$ tasks and $K = 1,000$, then we need to take $10 \times 1,000 = 10,000$ gradient steps before making a single meta-gradient step, which quickly becomes computationally expensive no matter how efficient the approximation is for computing each meta gradient. For meta-learning, we often need to perform a large number of meta-updates to ensure the convergence of the meta-model $\phi$, but having long trajectories of inner-gradient steps for large-scale tasks would prevent us from making sufficient meta-updates within a computing budget.



\begin{algorithm}[t]
	\caption{Previous meta-learning algorithms}\label{algo:previous}
	\small
	\begin{algorithmic}[1]
		\State \textbf{Input:} A set of tasks $\D^{(1)},\dots,\D^{(T)}$
		\State \textbf{Input:}
		Inner-learning rate $\alpha$, meta-learning rate $\beta$
		\State \textbf{Output:}
		Meta-learned initialization $\phi$
		\State Randomly initialize $\phi$.
		\For{$m=1$ to $M$} \Comment{Repeating inner-opt. processes}
		\For{$t=1$ to $T$} 
		\State $\theta_0^{(t)} \leftarrow \phi$ \Comment{Resetting each task learner}
		\For{$k=1$ to $K$} \Comment{Inner-optimization}
		\State $\theta^{(t)}_k \leftarrow \theta_{k-1}^{(t)} - \alpha \grad_{\theta}\mathcal{L}_k^{(t)}|_{\theta=\theta_{k-1}^{(t)}}$
		\EndFor
		\EndFor
		\State {\color{red} $\phi \leftarrow \phi - \beta \frac{1}{T}\sum_{t=1}^{T} \mathsf{MetaGrad}(\phi;\theta_K^{(t)})$ \Comment{Meta-update} } 
		\EndFor
	\end{algorithmic}
	\end{algorithm}

\begin{figure}[t]
    \vspace{-0.25in}
	\centering
	\hfill
	\subfigure[$k=1$]{
	    \includegraphics[height=1.9cm]{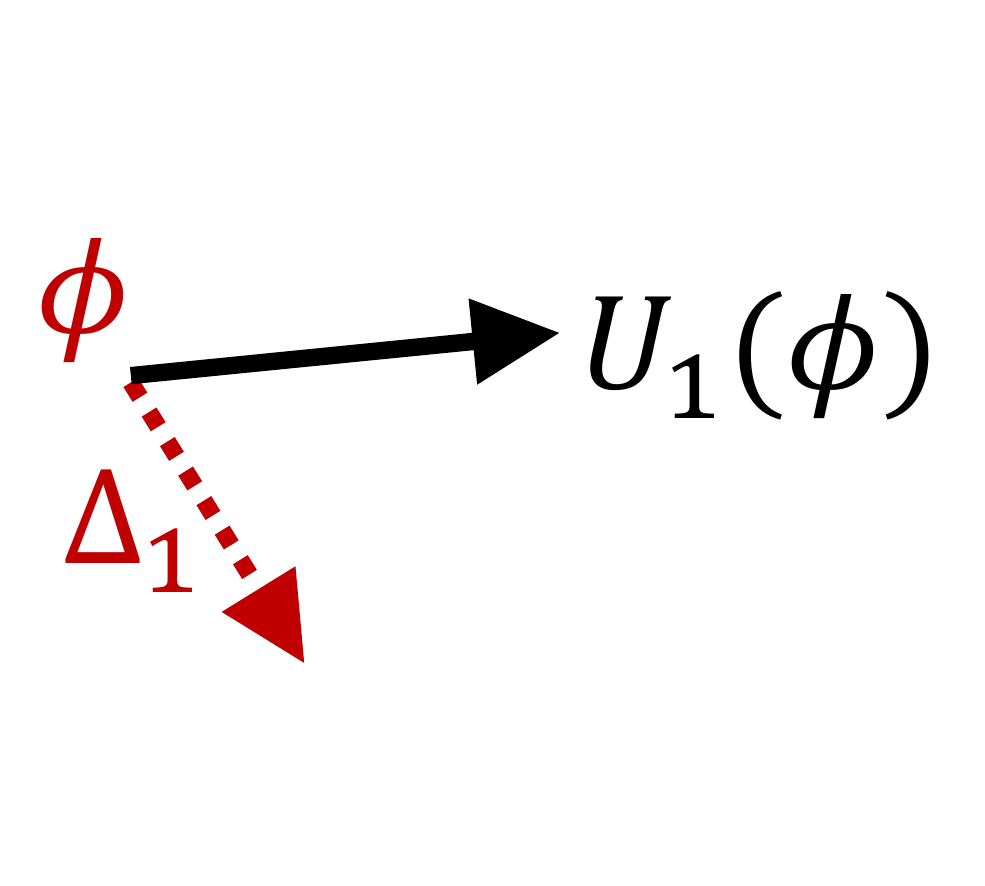}
	    \label{fig:detail_step1}
	}
	\hfill
	\subfigure[$k=2$]{
	    \includegraphics[height=1.9cm]{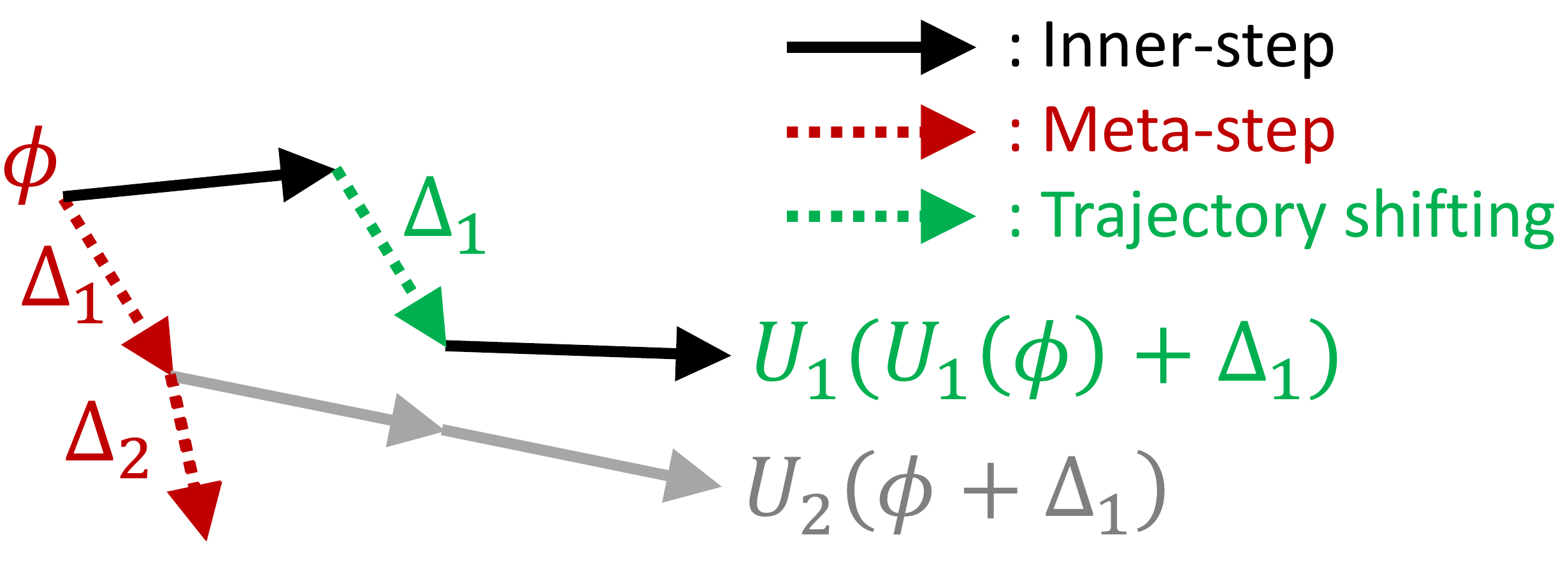}
	    \label{fig:detail_step2}
	}
	\hfill
	\vspace{-0.15in}
	\caption{Illustration of the proposed continual trajectory shifting.}
	\vspace{-0.12in}
	\label{fig:detail}
\end{figure}

\subsection{Continual trajectory shifting}
\label{sec:ours}

Our key idea is to interleave the meta-updates with the inner-optimization processes to reduce the long wait between two adjacent meta-updates. This is made possible by continual trajectory shifting described as follows.
We first introduce some notations. Let $U_k(\phi)$ denote a function that takes the initialization $\phi$ as input and outputs $\theta_k$, the $k$-th step parameters for solving a task. Here we drop the task dependency for notational brevity. For instance, if we use vanilla stochastic gradient descent, then we have $U_k(\phi) := \phi - \alpha \sum_{i=0}^{k-1} \grad_{\theta} \mathcal{L}_{i}|_{\theta=\theta_{i}}$ where $\theta_0 := \phi$ and $\alpha$ is the inner learning rate. \footnote{We do not impose any restrictions on the type of optimizers for $U_k(\phi)$. See \textbf{the supplementary file} for more discussion.} Denote by $\Delta_1,\Delta_2,\dots$ the series of meta-updates induced by all tasks, such that the shared initialization evolves as $\phi, \phi+\Delta_1, \phi + \Delta_1 + \Delta_2, \cdots$.

\begin{algorithm}[t ]
	\caption{Meta-learning with continual shifting}\label{algo:ours}
	\small
	\begin{algorithmic}[1]
		\State \textbf{Input:} A set of tasks $\D^{(1)},\dots,\D^{(T)}$
		\State \textbf{Input:}
		Inner-learning rate $\alpha$, meta-learning rate $\beta$
		\State \textbf{Output:}
		Meta-learned initialization $\phi$
		\State Randomly initialize $\phi$
		
		\For{$m=1$ to $M$} \Comment{ Reapating inner-opt. processes}
		\State $\theta_0^{(t)} \leftarrow \phi$ for $t=1,\dots,T$ \Comment{ Resetting task learners}
		\For{$k=1$ to $K$} \Comment{ Inner-opt. for all tasks}
		\For{$t=1$ to $T$}  \Comment{ Parallel for loop}
		
		\State $\theta^{(t)}_k \leftarrow \theta_{k-1}^{(t)} - \alpha  \grad_{\theta}\mathcal{L}_k^{(t)}|_{\theta=\theta_{k-1}^{(t)}}$
		\EndFor
		
		\State {\color{red}$\Delta_k \leftarrow - \beta \frac{1}{T}\sum_{t=1}^{T} 
		\mathsf{MetaGrad}(\phi;\theta_k^{(t)})$}
		
		\State {\color{red} $\phi \leftarrow \phi + \Delta_k$ \Comment{Meta-update}}
		\State {\color{red} $\theta_k^{(t)} \leftarrow \theta_k^{(t)} + \Delta_k$ for $t=1,\dots,T$ \Comment{Shifting} } 
		\EndFor
		\EndFor
	\end{algorithmic}
	\end{algorithm}

Now we show that \emph{we can perform $k$ meta-updates within $k$ inner-gradient steps}, unlike the previous meta-learning methods. Note that Reptile gradient $\phi-\theta_k$ depends only on the task-specific parameters $\theta_k := U_k(\phi)$. Based on this property\footnote{Note that we can use any meta-gradients with similar property.}, we propose to estimate $U_1(\phi),U_2(\phi+\Delta_1),\dots,U_k(\phi+\Delta_1+ \cdots + \Delta_{k-1})$ from a single inner-optimizatin process, and perform the meta-updates with them for every step up to $k$. 
Specifically, in Figure~\ref{fig:detail_step1}, we compute the first meta-update $\Delta_1$ based on the single-step task-specific parameters $U_1(\phi)$. Then in the next step $k=2$ in Figure~\ref{fig:detail_step2}, in order to compute the next meta-update $\Delta_2$ w.r.t. the new initialization point $\phi + \Delta_1$, we propose to approximate $U_2(\phi+\Delta_1)$ with $U_1(U_1(\phi) + \Delta_1)$, which we can obtain \emph{without actually taking gradient steps at} $\phi + \Delta_1$.
We generalize the approximation as follows:
\begin{align}
    &U_k(\phi + \Delta_1 + \cdots + \Delta_{k-1}) \nonumber \\
    &\approx U_1(\cdots U_1(U_1(\phi)+\Delta_1) \cdots + \Delta_{k-1}) 
    \label{eq:ours}
\end{align}
Eq.~\eqref{eq:ours} means that for every inner-step from $1$ to $k$, we continuously shift the task-specific learning trajectory by \emph{the same direction and amount of each meta-update}, thereby allowing the task-learning trajectory to remain consistent with the series of updates for the initialization parameters. See Figure~\ref{fig:detail_step2} and Algorithm~\ref{algo:ours} for the detailed procedure. We name this method as \emph{Continual Trajectory Shifting}. 

One important aspect of our method is that $k$, the inner-trajectory length used to compute each meta-update, gradually increases from $1$ to maximum $K$. In Figure~\ref{fig:detail_step1}, we compute $\Delta_1$ with the single-step task learner $U_1(\phi)$, and in Figure~\ref{fig:detail_step2} we compute $\Delta_2$ with $U_1(U_1(\phi) + \Delta_1)$, which is an approximation of the two-step task learner $U_2(\phi + \Delta_1)$. Later we will discuss the effect of gradually increasing $k$ as a meta-level regularizer.



\vspace{-0.025in}
\subsection{Approximation error}
\vspace{-0.025in}

We next analyze the approximation error in Eq.~\eqref{eq:ours}, which is central to our continual trajectory shifting method. We first show $U_k(\phi + \Delta) \approx U_k(\phi) + \Delta$ can be derived by applying two different approximations.
The first approximation is Taylor expansion: 
\begin{align}
&U_k(\phi + \Delta) \nonumber \\
&= U_k(\phi) + \frac{\partial U_k(\phi)}{\partial \phi}\Delta + \frac{1}{2} \Delta^\top \frac{\partial^2 U_k(\phi)}{\partial \phi^2} \Delta + \cdots \nonumber\\
&= U_k(\phi) + \frac{\partial U_k(\phi)}{\partial \phi}\Delta + O(\beta^2)
\label{eq:approx1}
\end{align}
\noindent $O( \beta^2)$ is because $\Delta = -\beta \cdot \mathsf{MetaGrad}(\phi;\theta_k) = O(\beta)$. 
Therefore, the first order Taylor approximation with Eq.~\eqref{eq:approx1} is reasonable if $\beta > 0$ is sufficiently small.

Secondly, we apply the Jacobian approximation frequently used by the first-order meta-learning algorithms~\cite{finn2017model,reptile,leap}:
\begin{align}
    \frac{\partial U_k(\phi)}{\partial\phi} &= \frac{\partial U_k(\phi)}{\partial U_{k-1}(\phi)} 
    \cdots
    \frac{\partial U_1(\phi)}{\partial\phi} = \prod_{i=0}^{k-1} \left(I-\alpha H_{i}\right)  \nonumber\\ 
    &= I + O(\alpha h k )
    \label{eq:approx2}
\end{align}
\noindent where we let $\theta_0 := \phi$, $\alpha > 0$ is the inner-learning rate, $H_i$ is the Hessian at step $i$, and $h$ denotes an upper bound of  norm of Hessians (e.g. spectral norm). As long as $\alpha h k > 0$ is significantly smaller than $1$, we can safely approximate Eq.~\eqref{eq:approx2} with the identity matrix $I$. 
Applying Eq.~\eqref{eq:approx1} and Eq.~\eqref{eq:approx2}, we have
\begin{align}
U_k(\phi + \Delta) &=U_k(\phi) + \Delta + O(\beta \alpha h k + \beta^2). \label{eq:error}
\end{align}
Based on Eq.~\eqref{eq:error}, we can derive the complexity of the approximation error caused by Eq.~\eqref{eq:ours}:
\begin{align}
    &U_k\left(\phi + \Delta_1 + \cdots + \Delta_{k-1} \right) \nonumber \\
    &=  U_1(\cdots U_1(U_1(\phi)+\Delta_1) \cdots + \Delta_{k-1})  \nonumber \\
    &\quad + O(\beta \alpha h k^2 + \beta^2 k). \label{eq:error_main} 
\end{align}
See \textbf{the supplementary file} for the derivation.

\begin{figure}[t]
    \vspace{-0.1in}
	\centering
	\hspace{-0.2in}
	\subfigure[$\alpha$]{
	    \raisebox{-0.08in}{
	    \includegraphics[height=3.9cm]{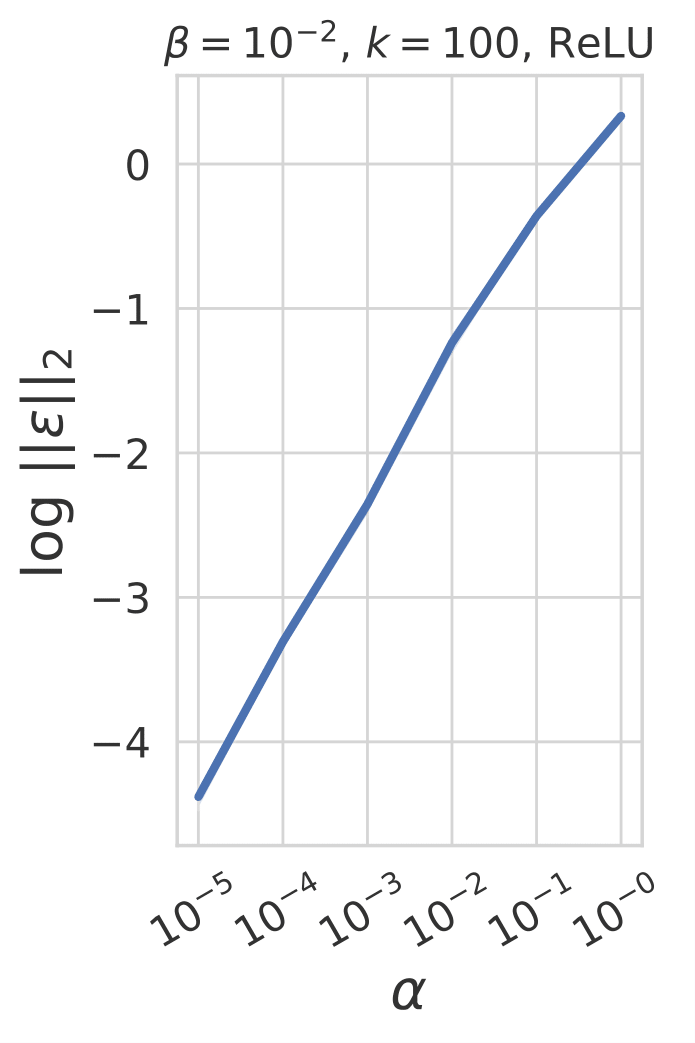}
	    \label{fig:err_alpha}
	    }
	}
	\hspace{-0.2in}
	\subfigure[$\beta$]{
	    \raisebox{-0.08in}{
	    \includegraphics[height=3.9cm]{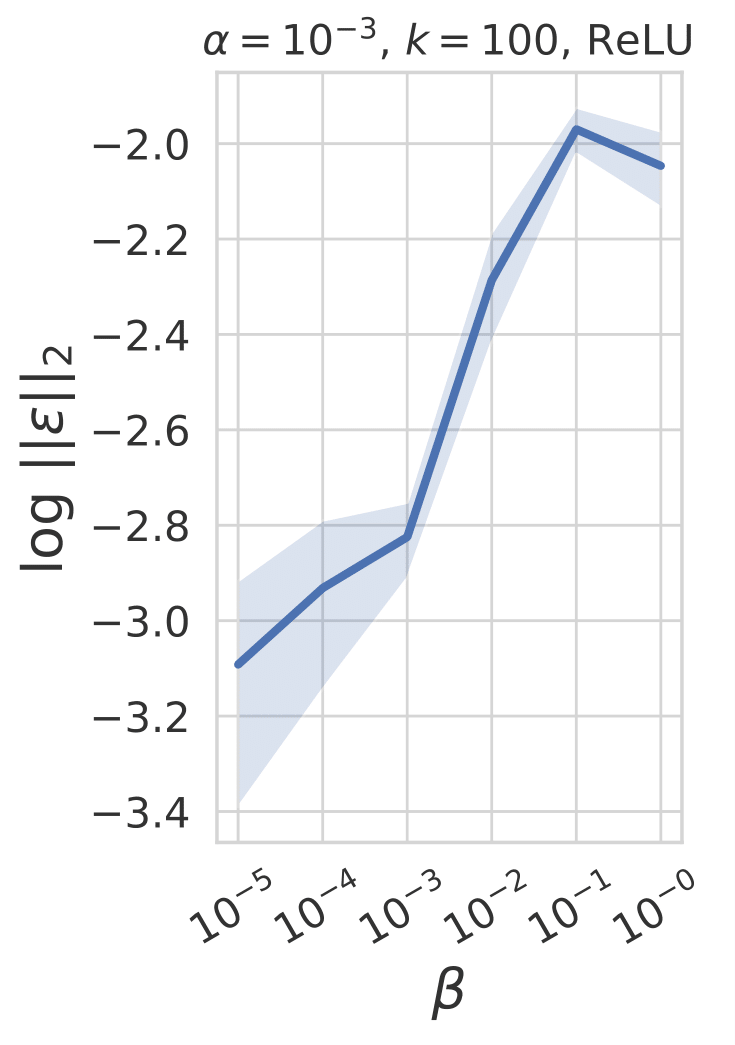}
	    \label{fig:err_beta}
	    }
	}
	\hspace{-0.2in}
	\subfigure[$k$, Activation]{
	    \raisebox{-0.08in}{
	    \includegraphics[height=3.9cm]{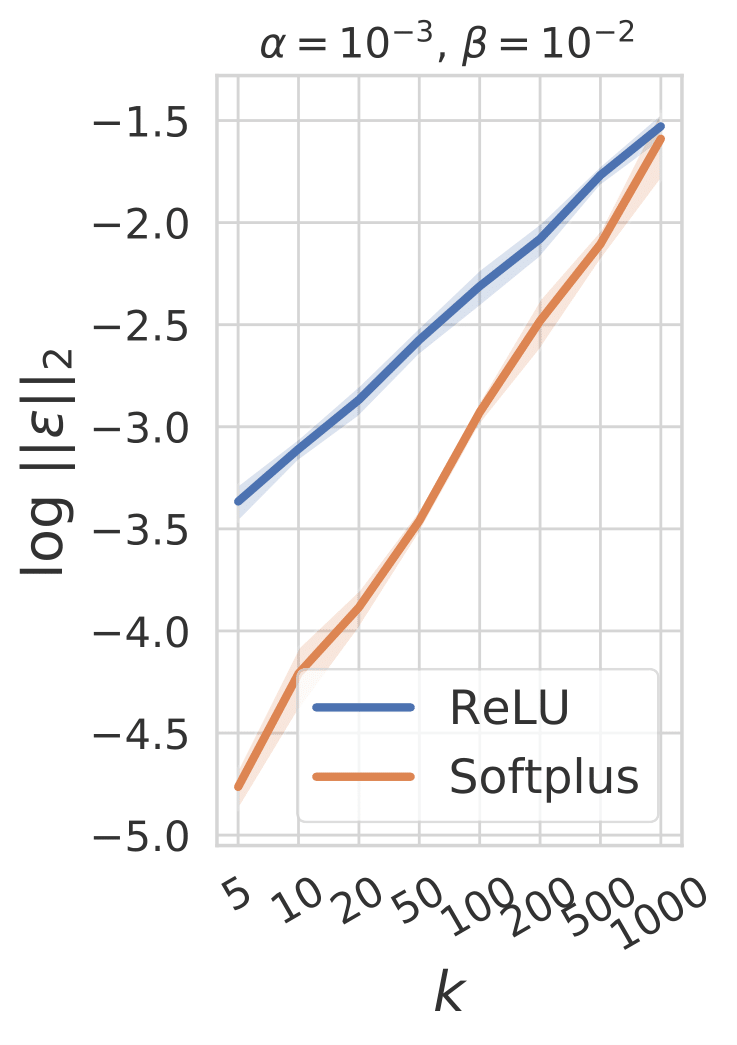}
	    \label{fig:err_k}
	    }
	}
	\hspace{-0.2in}
	\vspace{-0.17in}
	\caption{\small \textbf{Approximation error} versus inner-learning rate $\alpha$, meta-learning rate $\beta$, inner-learning trajectory length $k$, and the type of network activations. 
	We report the mean and $95$\% confidence intervals over $10$ draws of inner-learning trajectories. See \textbf{the supplementary file} for the detailed experimental setup.}
	\vspace{-0.15in}
	\label{fig:error}
\end{figure}

\vspace{-0.05in}
\paragraph{Empirical analysis.} Then, is the approximation error in Eq.~\eqref{eq:error_main} empirically manageable? To answer the question, we define the error $\varepsilon := U_k(\phi + \Delta_1 + \cdots + \Delta_{k-1} ) - U_1(\cdots U_1(U_1(\phi)+\Delta_1) \cdots + \Delta_{k-1}) $ and collect the norm of $\varepsilon$ empirically. We see from Figure~\ref{fig:error} that the error sharply increases in proportion to $\alpha$, $\beta$, and $k$. Especially, Figure~\ref{fig:err_k} shows the difficulty of managing the error for the large-scale tasks that require a large number of gradient steps. Further, the use of ReLU activations and max-pooling introduces additional errors~\cite{balduzzi17b,balduzzi17c}. 
It is because the Taylor expansion assumes infinitely differentiable functions, but ReLU and max-pooling are not differentiable at certain points.
See Figure~\ref{fig:err_k} which shows that the networks with ReLU activations yield more inaccurate approximations over ones with Softplus activations. In conclusion, for most of the modern convolutional networks and large-scale tasks, we cannot guarantee that the proposed approximation will be highly accurate. However, we empirically found that the method still works very well even with the large approximation error. We provide a plausible interpretation about the results in the next subsection.

\begin{figure*}[t]
\vspace{-0.125in}
	\centering
	\subfigure[Template function]{
	    \includegraphics[height=3.6cm]{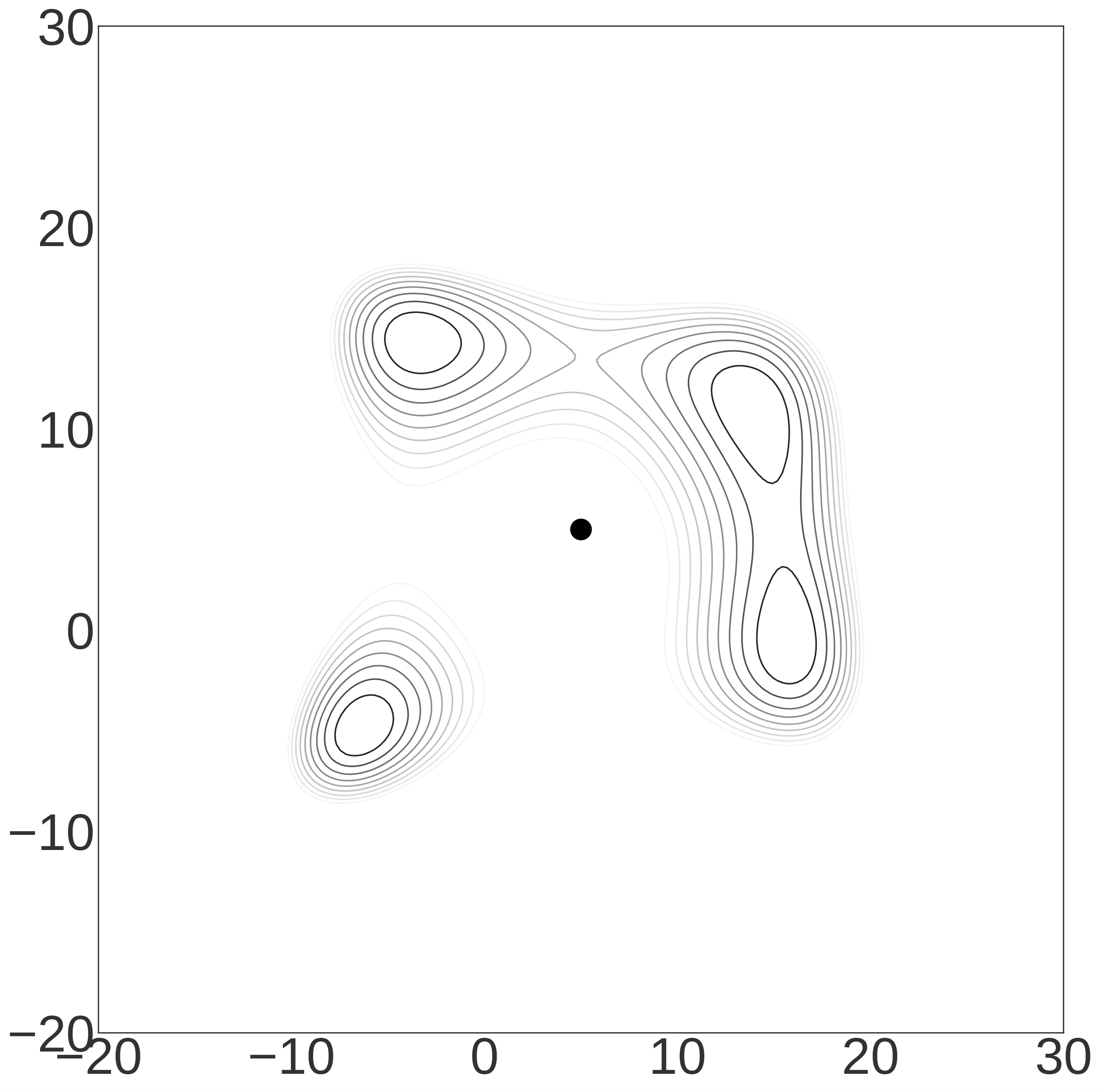}
	    \label{fig:task}
	}
	\subfigure[Task $1$ loss function $\mathcal{L}^{(1)}$]{
	    \includegraphics[height=3.6cm]{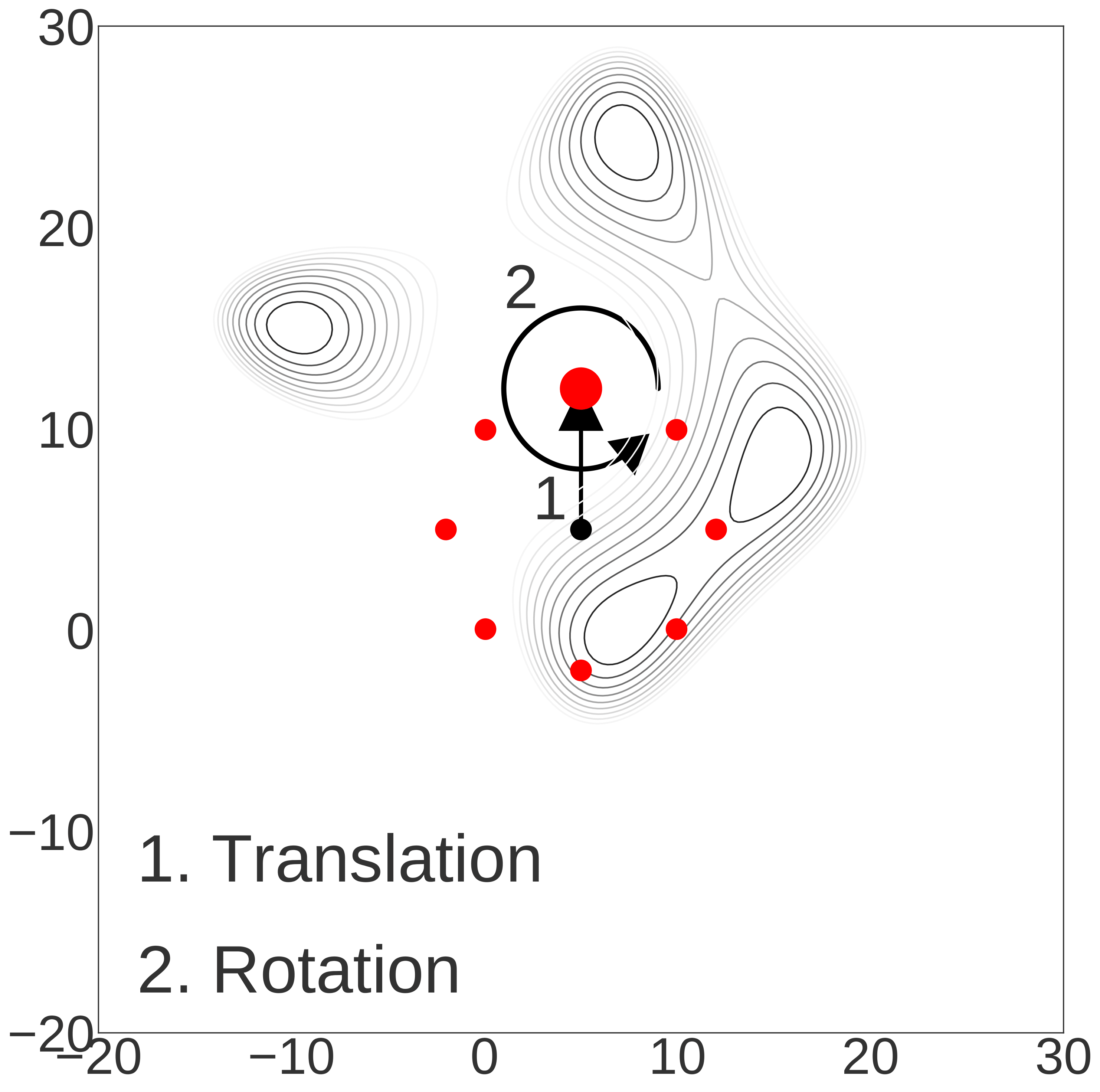}
	    \label{fig:task1}
	}
	\subfigure[Short horizon bias]{
	    \includegraphics[height=3.6cm]{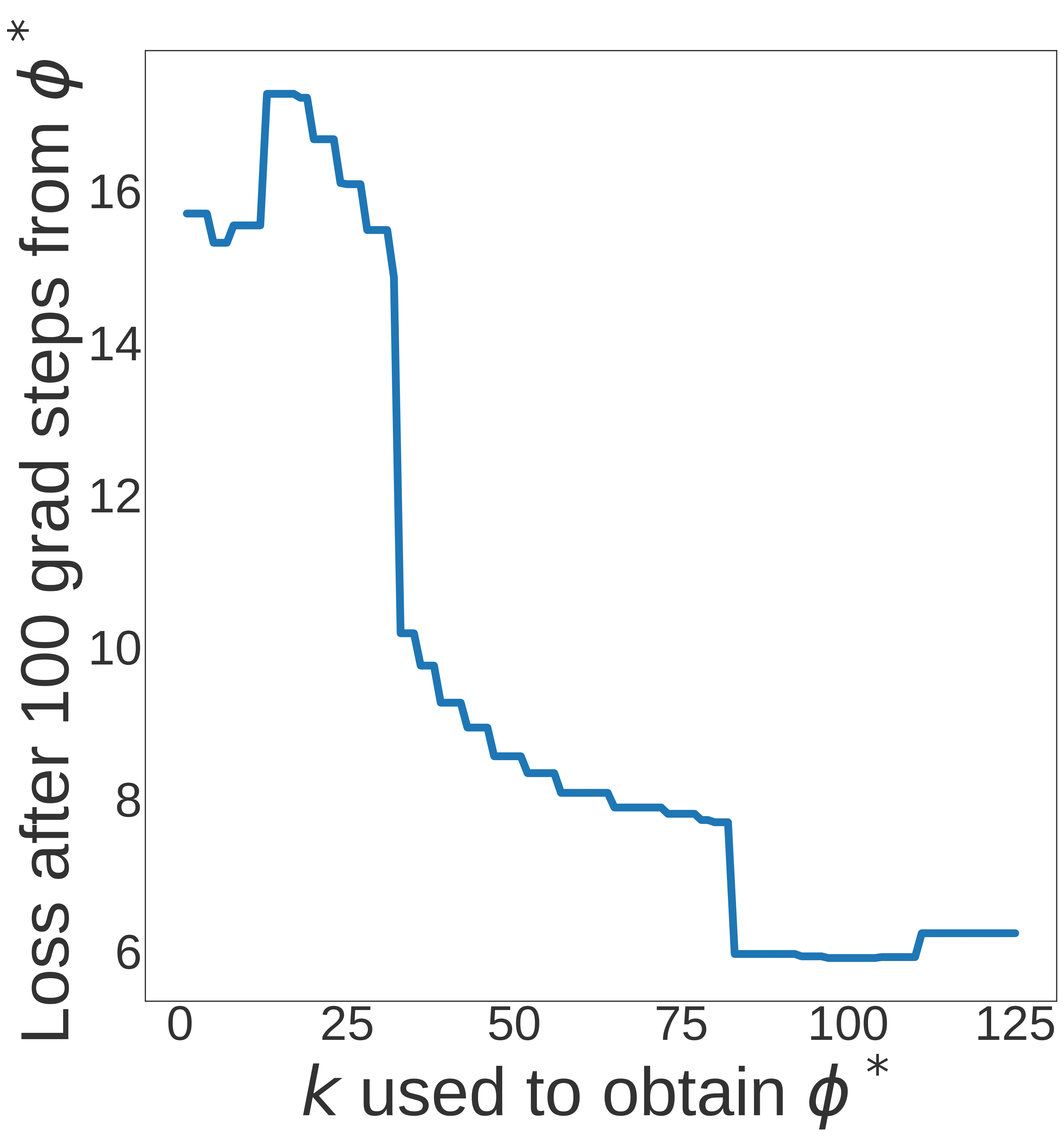}
	    \label{fig:largeK}
	}
	\subfigure[Computational cost]{
	    \includegraphics[height=3.5cm]{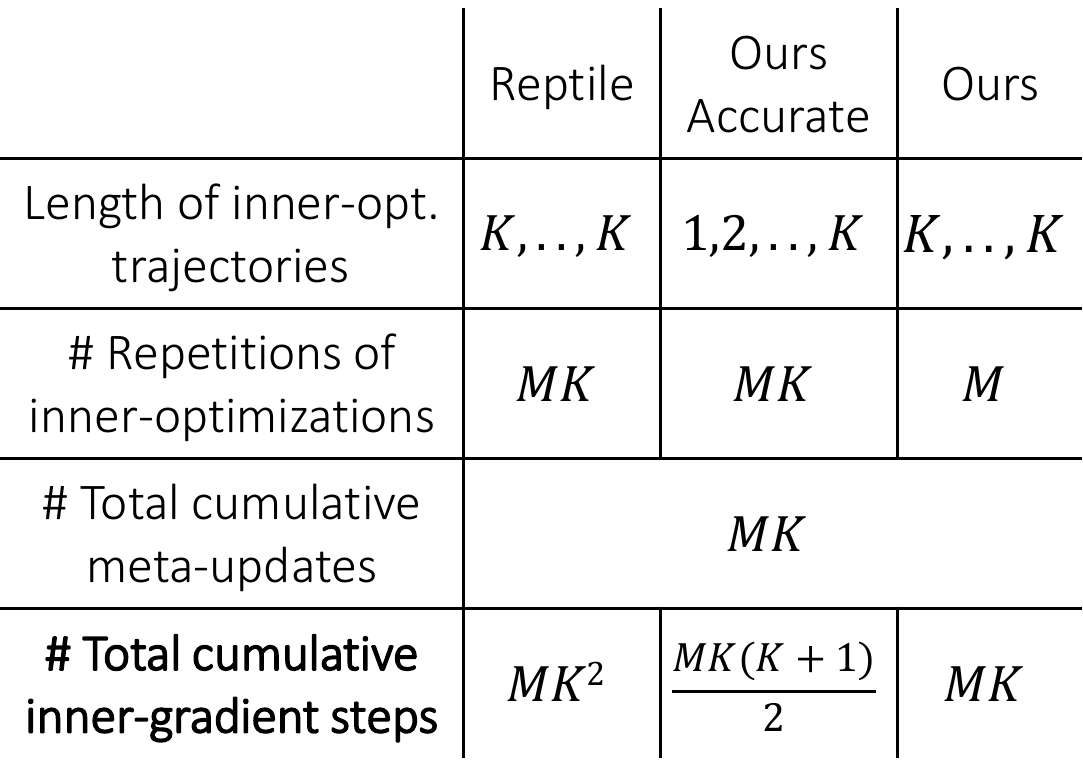}
	    \label{fig:thetable}
	}
	\vspace{-0.18in}
	\caption{\small \textbf{(a)} Template function used to generate the task loss functions. \textbf{(b)} Task $1$ loss function $\mathcal{L}^{(1)}$ obtained by applying translation (straight arrow) and random rotation (round arrow). \textbf{(c)} Task-average loss after $100$ gradient steps from $\phi^*$ vs. $k$ used to obtain the optimal initialization parameters $\phi^*$ over the tasks.  \textbf{(d)} Computational cost in terms of total cumulative number of inner-gradient steps. }
    \label{fig:synthetic_figs1}
    \vspace{-0.15in}
\end{figure*} 
\begin{figure*}[t]
    \vspace{-0.05in}
	\centering
	\subfigure[$k=5$]{
	    \includegraphics[height=3.95cm]{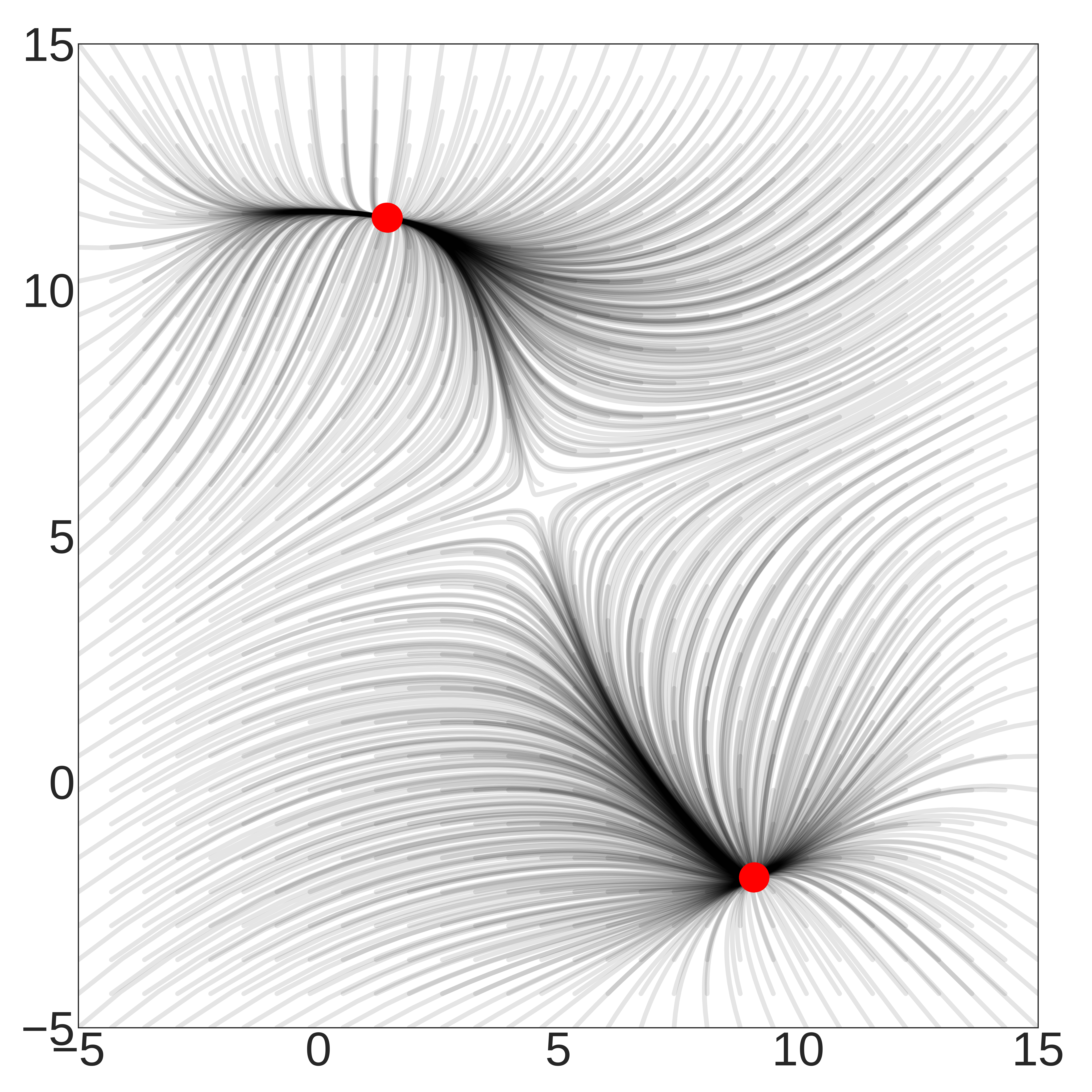}
	    \label{fig:bar_5}
	}
	\subfigure[$k=100$]{
	    \includegraphics[height=3.95cm]{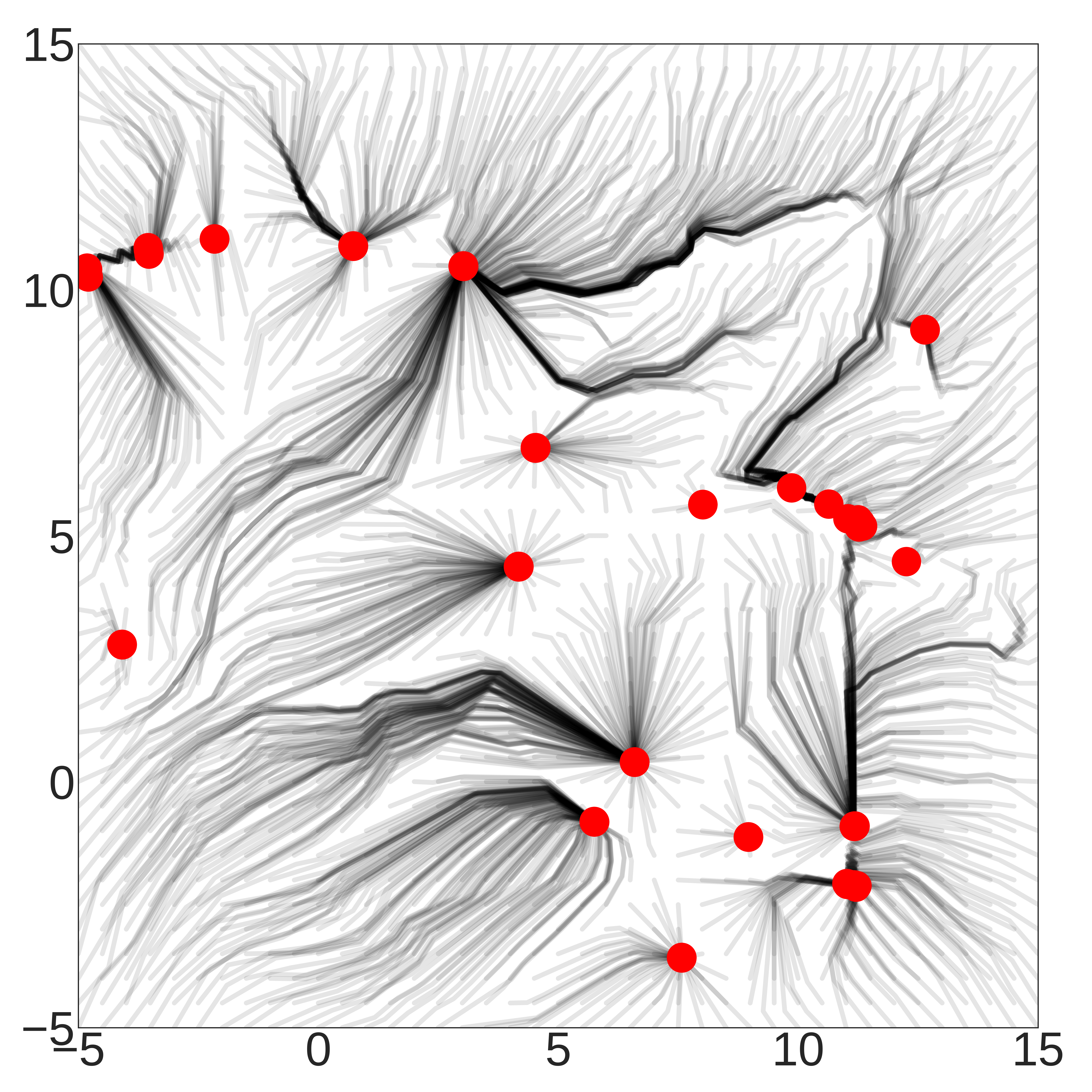}
	    \label{fig:bar_80}
	}
	\subfigure[Starting point: $(-5,5)$]{
	    \includegraphics[height=3.9cm]{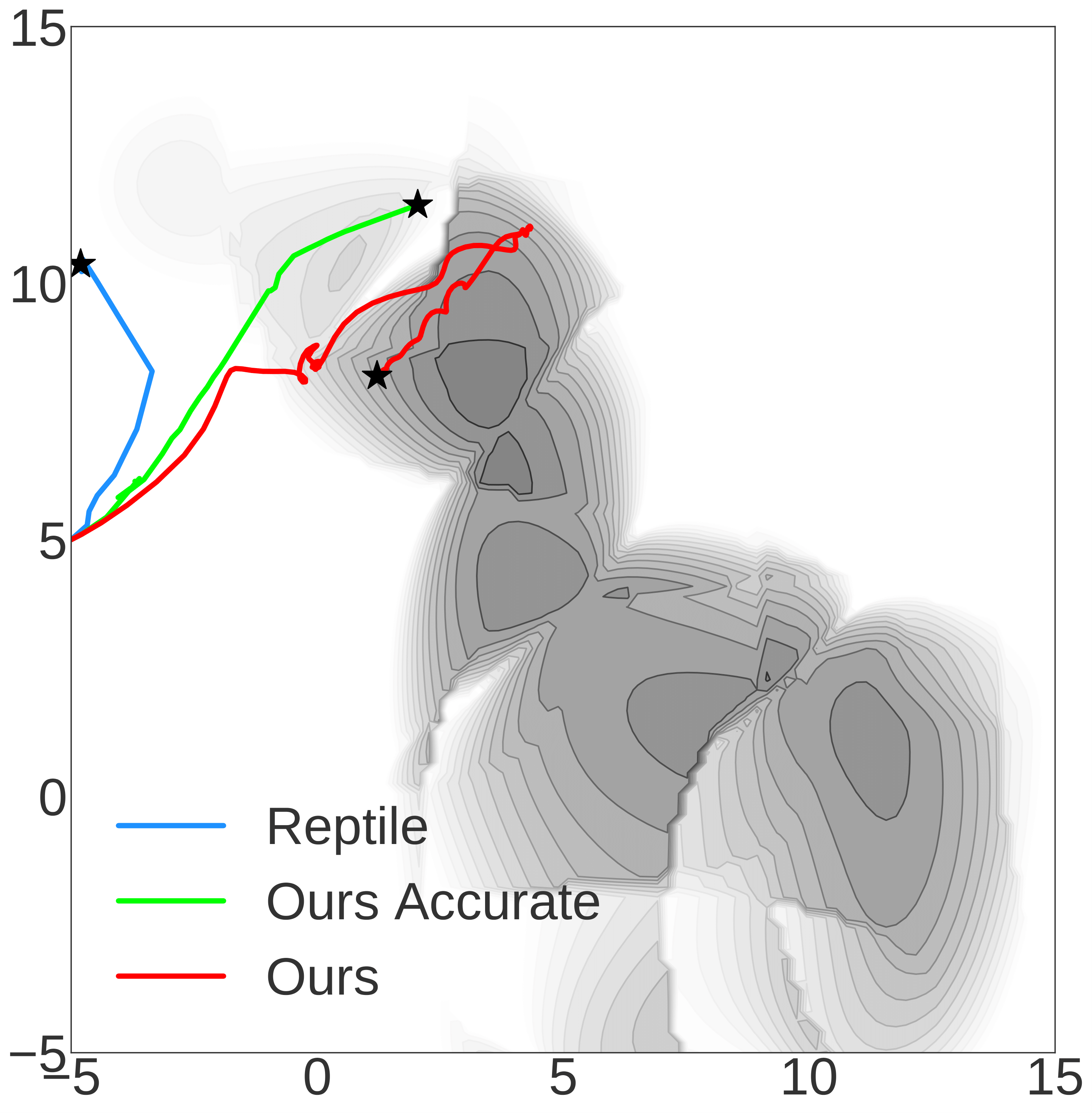}
	    \label{fig:quality_phi1}
	}
	\subfigure[Starting point: $(5,-5)$]{
	    \includegraphics[height=3.9cm]{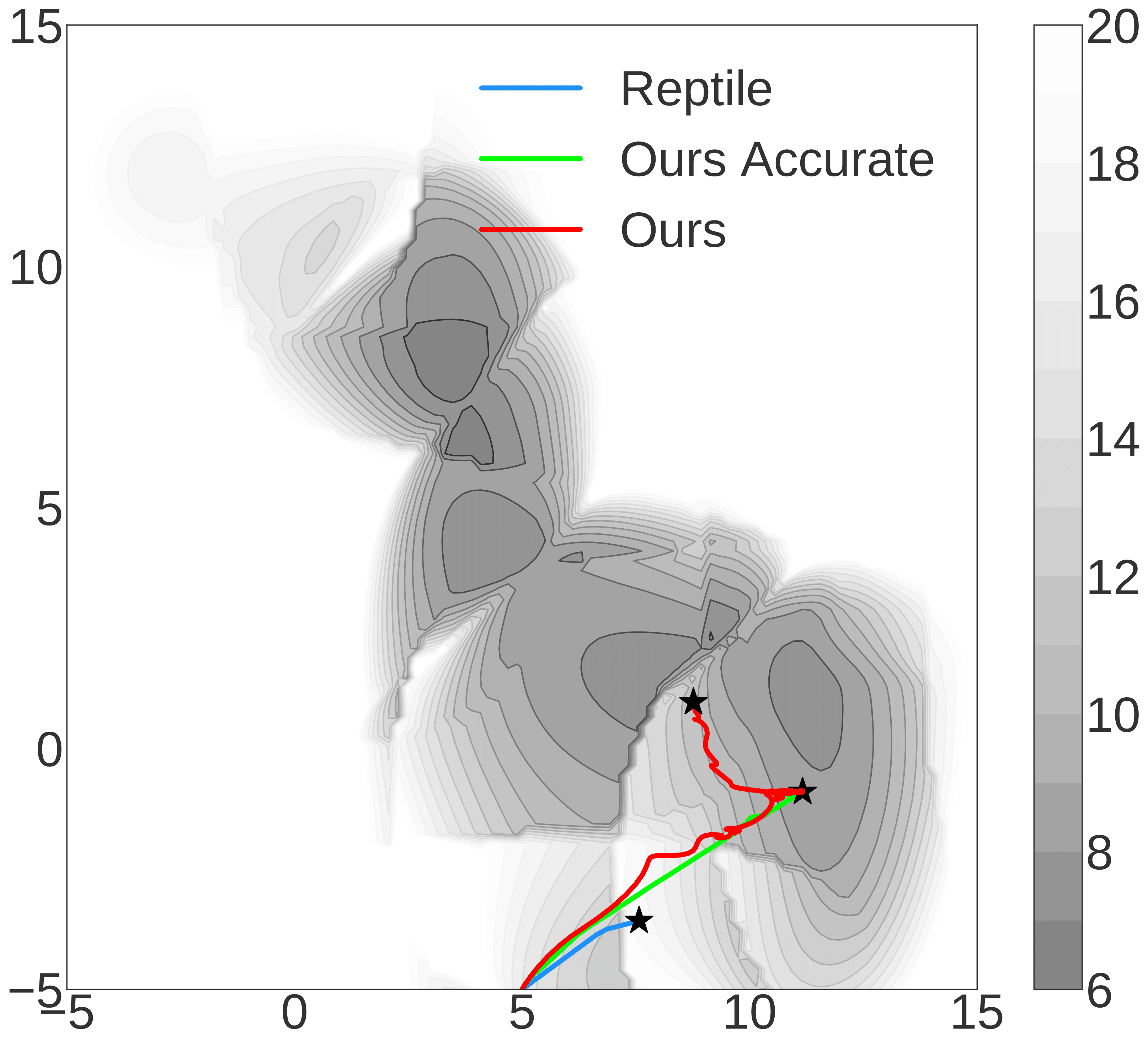}
	    \label{fig:quality_phi2}
	}
	\vspace{-0.18in}
	\caption{\small \textbf{(a,b) Meta-learning trajectories of Reptile} with the length of inner-optimization fixed as $k$. We collect the trajectories by initiating them from the various points in the grid of $\phi$ space. Meta-level local optima are shown by the red dots. \textbf{(c,d) Background contour: Task-average loss} after taking $100$ gradient steps from each point. The darker the better quality of the initialization point. \textbf{Lines: Meta-learning trajectories} of $\phi$ obtained from the baselines and our algorithm.  }
	\vspace{-0.15in}
    \label{fig:synthetic_figs2}
\end{figure*} 

\vspace{-0.025in}
\subsection{Meta-level curriculum learning with increasing $k$}
\vspace{-0.025in}

\label{sec:curriculum}
Recall from Section~\ref{sec:ours} that our method computes each meta-update with the gradually increasing $k$, the number of inner-gradient steps. The original motivation of gradually increasing $k$ came from interleaving every inner-optimization step with a meta-update, but we find that it introduces another benefit: a regularization effect. This is because our algorithm could be considered as an instance of curriculum learning at the meta-level. Curriculum learning~\cite{BengioLCW09} is a learning strategy where we present training examples from easy to more difficult ones, thereby sequentially controlling the complexity of the loss landscape. It has been empirically shown that the strategy improves the speed of convergence and the quality of local optima. 

In our case, the number of inner-gradient steps $k$ used to compute each meta-gradient determines the complexity of the meta-training loss landscape. Starting from $k=1$, the meta-learner first seeks to find a slightly better initialization point $\phi$ than the old one based on the very limited information about the task learning trajectories due to the short horizon bias~\cite{wu2018understanding}. The bias simplifies the meta-level loss landscape and thus lowers the risk of falling into bad local minima, which is especially beneficial for the early stage of meta-training (See Figure~\ref{fig:concept_ours}, left). After alleviating the risk, the meta-learner gradually increases $k$ to have more complex loss surfaces and find more informative local minima with longer horizons (See Figure~\ref{fig:concept_ours}, right). It partly explains how our method finds better initialization parameters than those by the previous meta-learning with a fixed length of inner learning trajectories. See Figure~\ref{fig:bar_5}, \ref{fig:bar_80}, and Section~\ref{sec:synthetic} for the discussions with real examples.

\vspace{-0.125in}
\paragraph{Curriculum learning and approximation error.} The curriculum learning effect also partly explains how our model performs well even with the fairly large approximation error. Note that the risk of bad local optima is significant at the beginning of the meta-training. 
Since our approximation is relatively accurate when $k$ is small, our model can enjoy the curriculum learning effect even if the approximation error goes up as $k$ increases. 

\begin{figure*}[t]
    \vspace{-0.15in}
	\centering
	\hspace{-0.1in}
	\hfill
	\subfigure[Meta-convergence]{
	    \raisebox{-0.05in}{
	    \includegraphics[height=3.8cm]{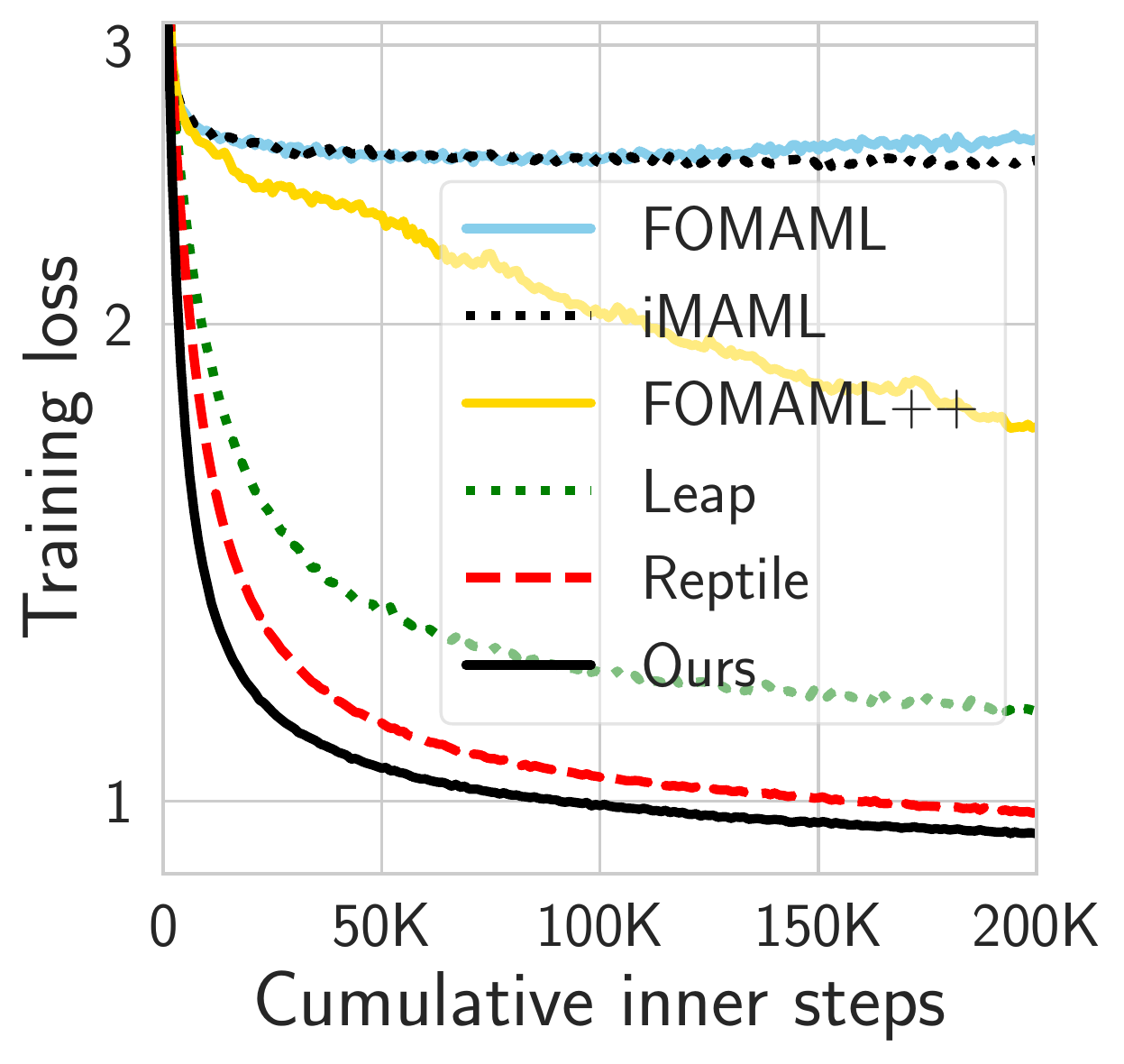}
	    \label{fig:meta_convergence}
	    }
	}
	\hspace{-0.2in}
	\hfill
	\subfigure[Meta-testing performance of the baselines]{
		\raisebox{-0.05in}{
	    \includegraphics[height=3.8cm]{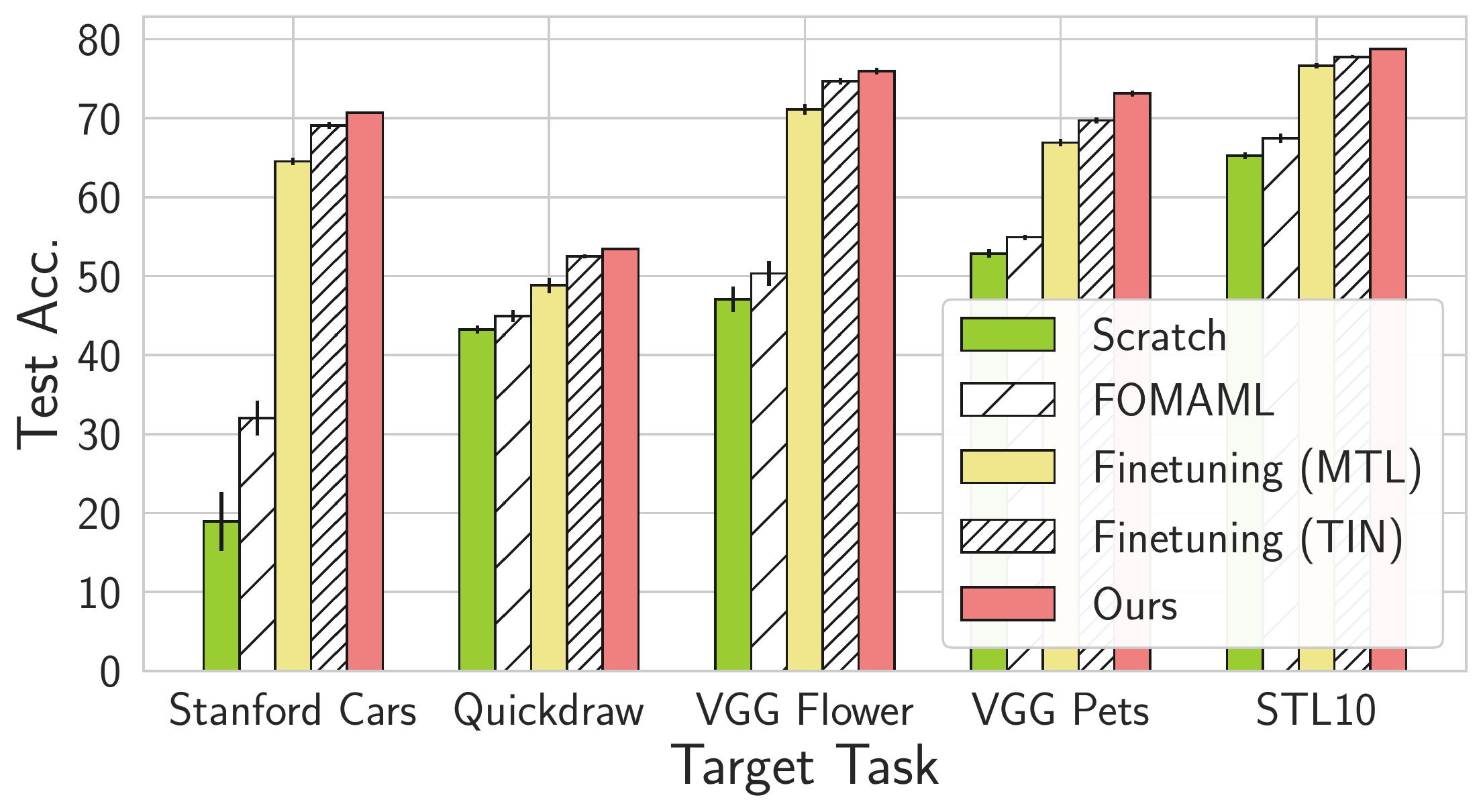}
	    \label{fig:bar}
	    }
	}
	\hspace{-0.2in}
	\hfill
	\subfigure[VGG Pets]{
	    \raisebox{-0.05in}{
	    \includegraphics[height=3.8cm]{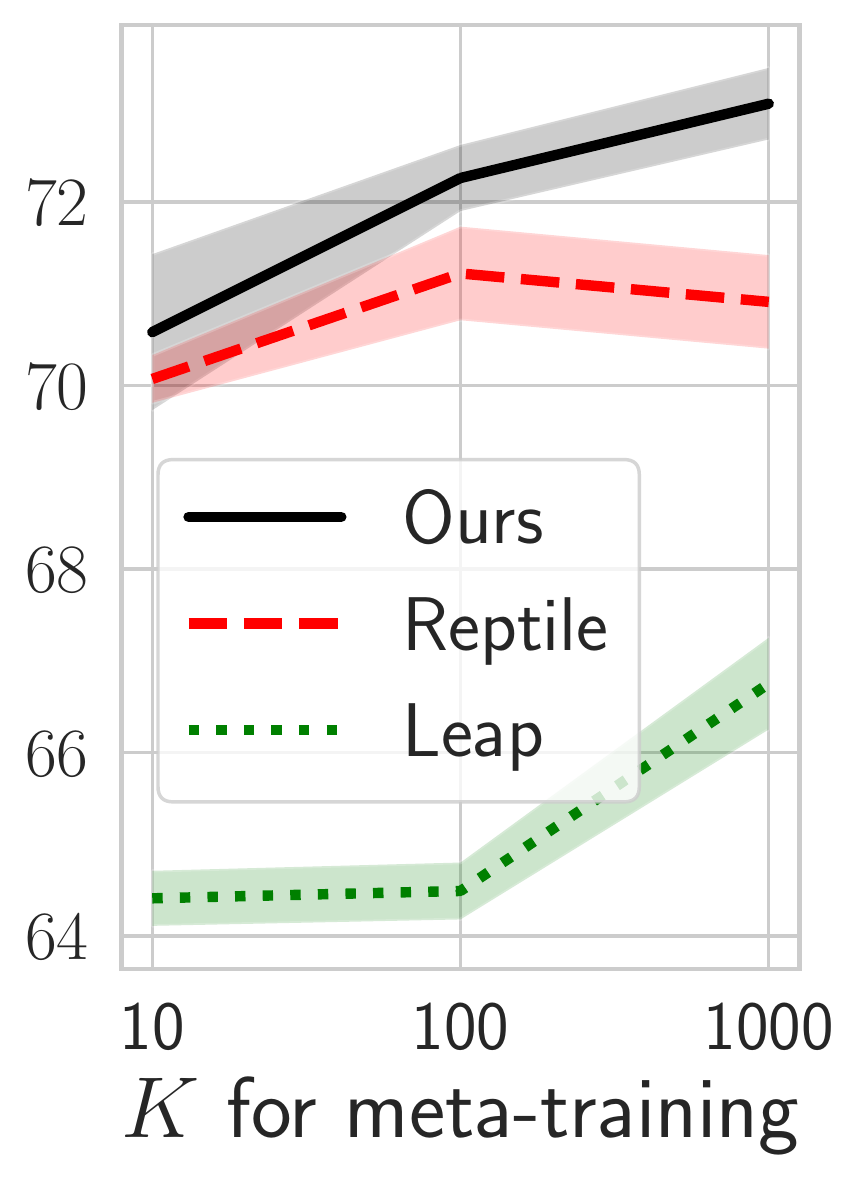}
	    \label{fig:varyingK_pets}
	    }
	}
	\hspace{-0.2in}
	\hfill
	\subfigure[STL10]{
	    \raisebox{-0.05in}{
	    \includegraphics[height=3.8cm]{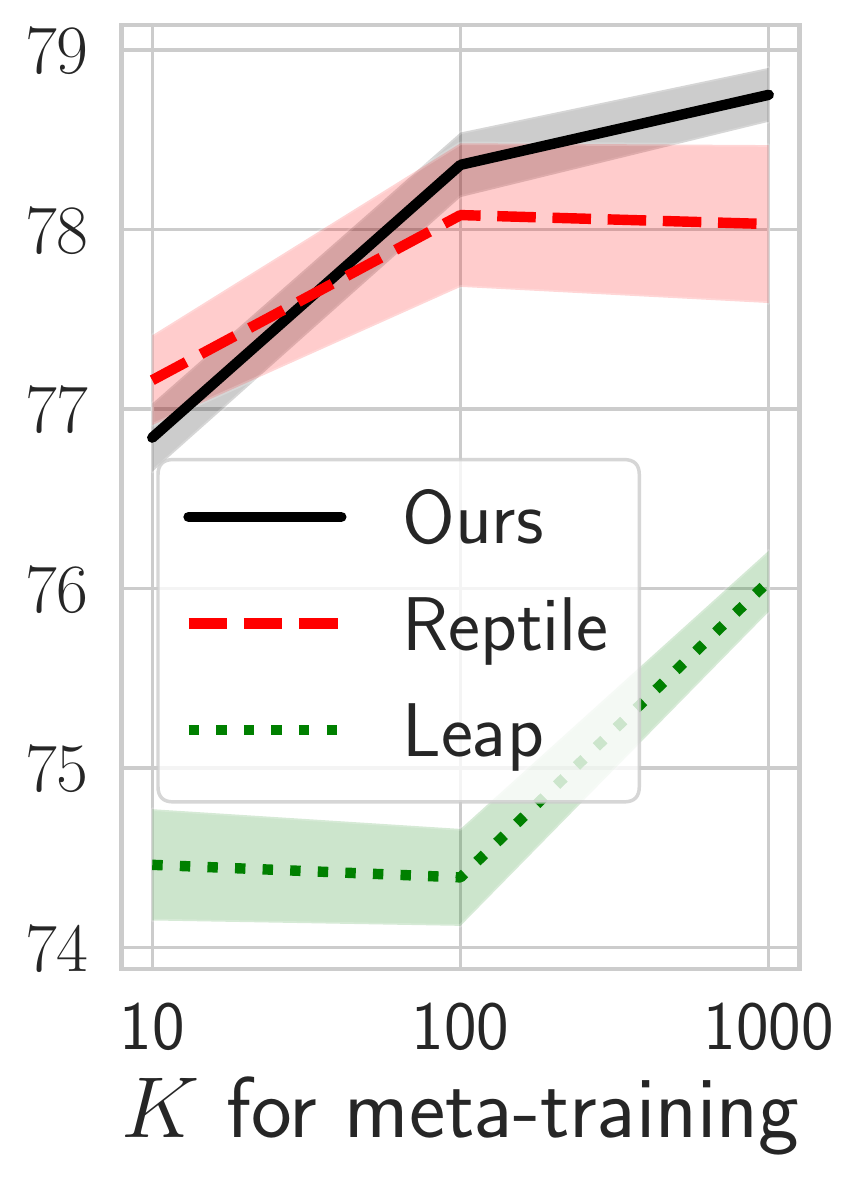}
	    \label{fig:varyingK_stl10}
	    }
	}
	\hspace{-0.1in}
	\hfill
	\vspace{-0.15in}
	\caption{\textbf{(a) Meta-training convergence} measured as task-average training loss vs. cumulative inner-gradient steps.  \textbf{(b) Meta-testing performance} of the baselines. \textbf{(c) Meta-testing performance} vs. $K$ used for meta-training.}
	\vspace{-0.2in}
\end{figure*}

\begin{figure*}[t]
	\centering
	\hspace{-0.2in}
	\hfill
	\subfigure[Stanford Cars]{
	    \raisebox{-0.05in}{
	    \includegraphics[height=3.8cm]{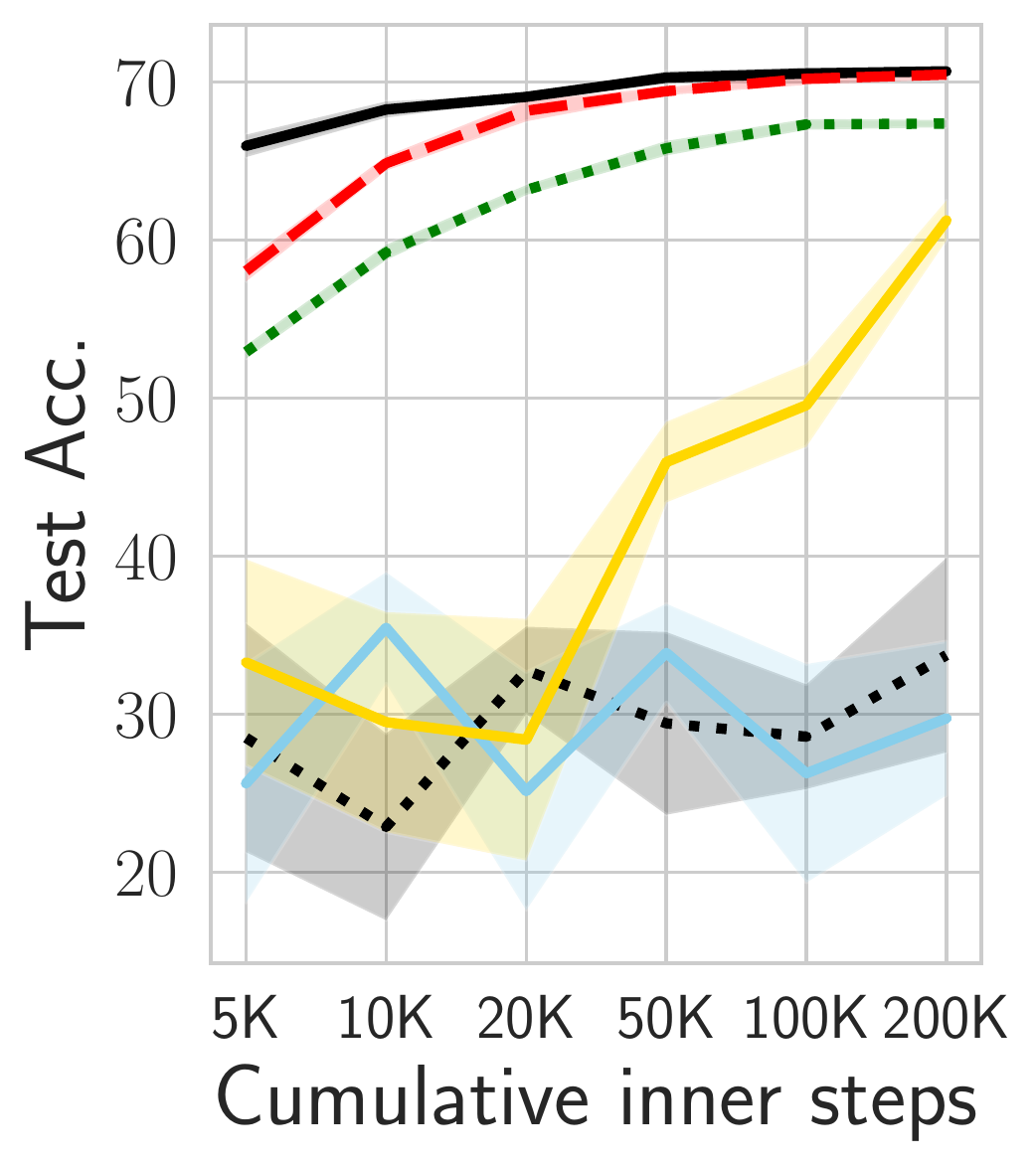}
	    }
	}
	\hspace{-0.2in}
	\hfill
    \subfigure[Quickdraw]{
	    \raisebox{-0.05in}{
	    \includegraphics[height=3.8cm]{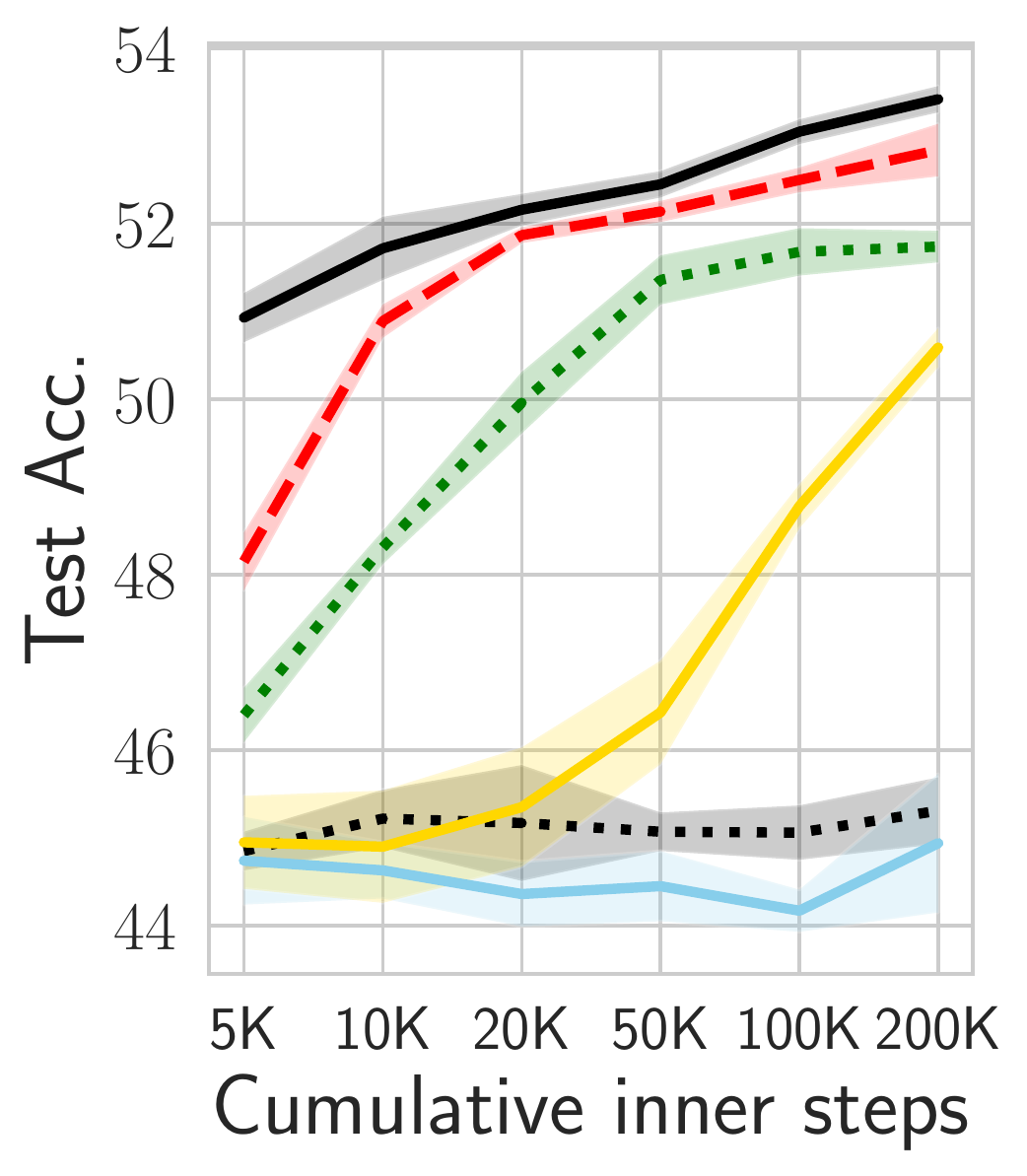}
	    }
	}
	\hspace{-0.2in}
	\hfill
	\subfigure[VGG Flowers]{
	    \raisebox{-0.05in}{
	    \includegraphics[height=3.8cm]{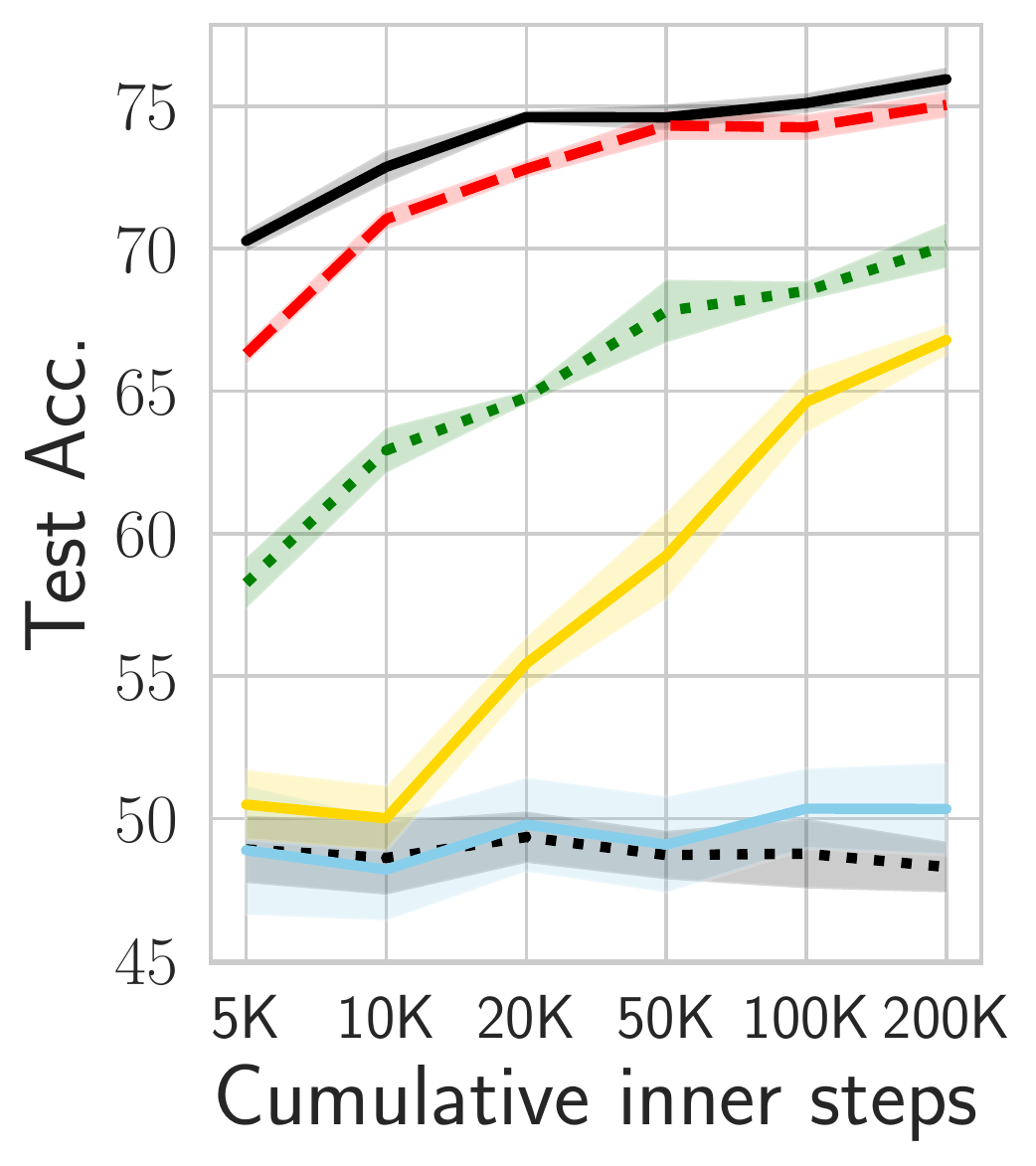}
	    }
	}
	\hspace{-0.2in}
	\hfill
    \subfigure[VGG Pets]{
	    \raisebox{-0.05in}{
	    \includegraphics[height=3.8cm]{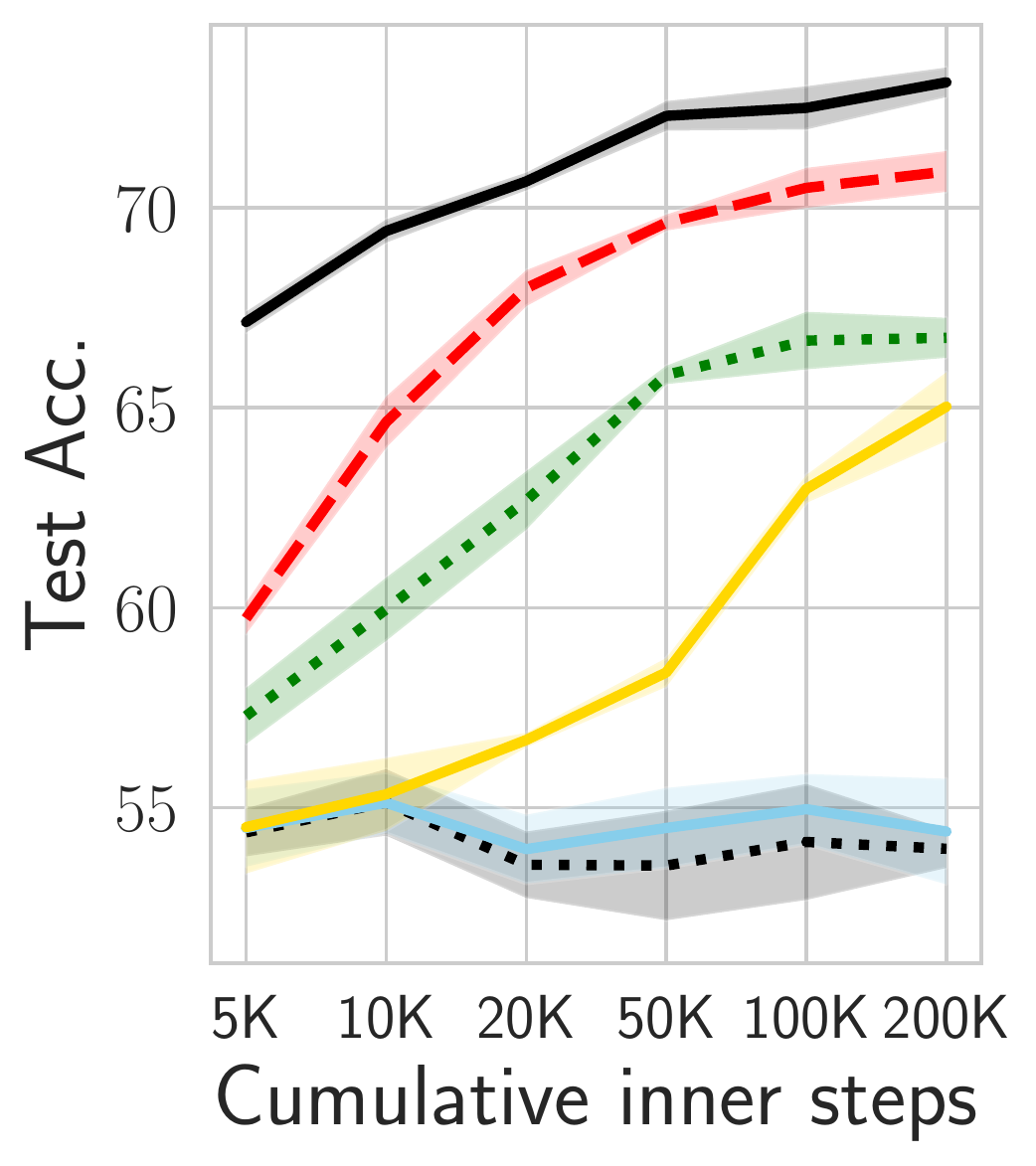}
	    }
	}
	\hspace{-0.2in}
	\hfill
	\subfigure[STL10]{
	    \raisebox{-0.05in}{
	    \includegraphics[height=3.8cm]{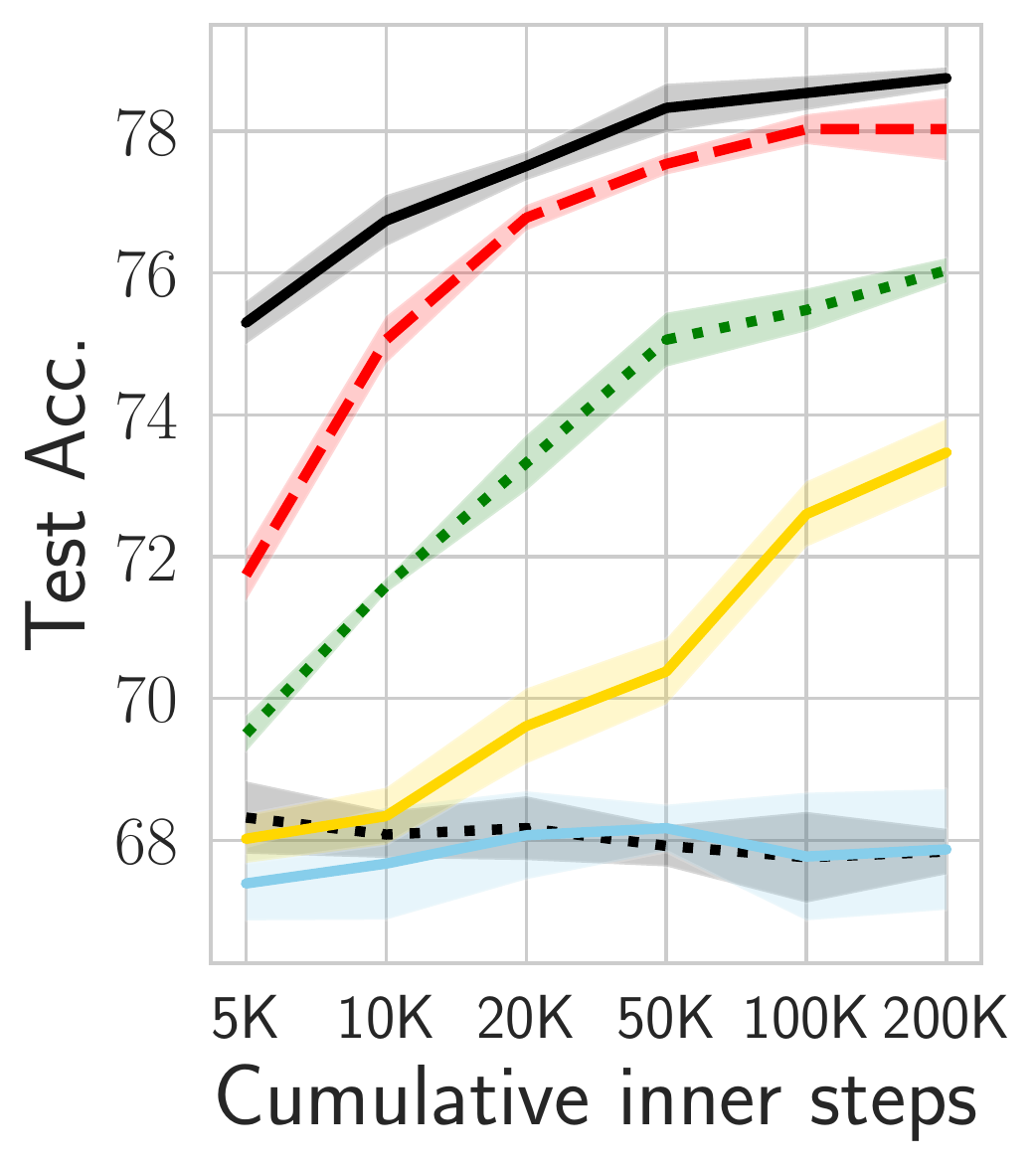}
	    }
	}
	\hspace{-0.2in}
	\hfill
	\vspace{-0.15in}
	\caption{\textbf{Meta-testing performance} to show how efficient each method is in terms of cumulative inner-gradient steps spent for meta-training. We report mean and $95\%$ confidence intervals over $5$ runs. }
	\label{fig:new6}
	\vspace{-0.15in}
\end{figure*} 

\vspace{-0.05in}
\section{Experiments}
\vspace{-0.025in}
\label{experiments}

We first examine how and why our method outperforms the baselines with synthetic experiments. We then verify the effectiveness of our method on a set of large-scale heterogeneous tasks, comparing to finetuning baselines and first-order meta-learning algorithms. 

\vspace{-0.05in}
\subsection{Synthetic experiments}
\label{sec:synthetic}
\vspace{-0.05in}
We first experiment with a synthetic task distribution to provide insights on how our algorithm works.
\vspace{-0.125in}
\paragraph{Task distribution.} We define a 2D function $f(x,y) = \left\{(x^2 -10x + y + 9)^2 + (x + y^2-10y+13)^2 \right\}/3$ which has four global minima (Figure~\ref{fig:task}). We shift this template function toward each of the red dots in Figure~\ref{fig:task1}, which form a circle centered at $(5,5)$, and randomly rotate around each dot to generate eight task losses $\mathcal{L}^{(1)}\dots\mathcal{L}^{(8)}$. Although the tasks share the same loss surface shape, they are heterogeneous since the rotations are random. We use all the eight tasks for meta-training to analyze the meta-convergence of different methods.

\vspace{-0.125in}
\paragraph{Baselines.} We compare \textbf{Ours} with \textbf{Reptile}~\cite{reptile} and \textbf{Ours Accurate}. Ours Accurate computes each of the meta-updates $\Delta_1,\dots,\Delta_{k}$ without the approximation errors for the task-specific parameters in Eq.~\eqref{eq:error_main}. Specifically, Ours Accurate directly computes $U_1(\phi),U_2(\phi+\Delta_1),\dots,U_k(\phi+\Delta_1+\cdots+\Delta_{k-1})$ by repeatedly reinitiating the inner-learning processes after each meta-update, which is computationally far more inefficient than Ours. See Figure~\ref{fig:thetable} for the computational cost for each method. Note that we let the methods perform the same number of total meta-updates.
\textbf{Experimental setup:} We use $\alpha=0.05$, $\beta=0.1$, $K=100$, and $M=3$. We set the inner-optimizer to SGD with momentun ($\mu=0.9$). See \textbf{the supplementary file} for more information. 

\vspace{-0.125in}
\paragraph{Results and analysis.} We make the following observations from the synthetic experiment. Firstly, we should use a large $K$ for meta-training if we want to use a large $K$ for meta-testing. Figure~\ref{fig:largeK} demonstrates the existence of the short horizon bias. It shows that the optimal initialization $\phi^*$ obtained with small $K$ (e.g. $K=25$) cannot provide good performance even if we take a sufficient number of gradient steps from there (e.g. $K=100$).

Also, if we use a large $K$ during meta-training, we can allow the meta-learner to avoid bad local optima at the early stage. It is done by gradually increasing the trajectory length $k$ from $1$ to maximum $K$, which is used to compute each meta-gradient. In order to demonstrate this effect, we visualize the meta loss landscape over $\phi$ with various $k$ in Figure~\ref{fig:bar_5} and \ref{fig:bar_80}, by simply collecting the meta-learning trajectories starting from various points in the spatial grid of $\phi$.
We see from Figure~\ref{fig:bar_80} that there exist many local minima for large $k$. It is because the longer horizons make the task-specific parameters to sensitively react to a small change in the initialization $\phi$, making the direction of meta-gradient frequently change over the space of $\phi$.
As a result, comparing the local optima in Figure~\ref{fig:bar_80} with the map of initialization quality in Figure~\ref{fig:quality_phi1}, we see that many of the the local optima are of low quality and also attract the meta-learner even from the beginning. See Figure~\ref{fig:quality_phi1} and~\ref{fig:quality_phi2} that Reptile gets stuck in a bad local minimum, whereas Ours and Ours Accurate can circumvent it. It shows that Ours and Ours Accurate actually make use of the much simpler loss landscape provided by smaller $k$, effectively lowering the risk of bad local minima. Note that the short horizon bias introduced by smaller $k$ is only temporary as we gradually increase $k$ up to maximum $K$ over the course of inner-optimization processes.

Lastly, Figure~\ref{fig:quality_phi1} and~\ref{fig:quality_phi2} show that although Ours and Ours Accurate reveal dissimilar meta-learning trajectories in general, the early part of the trajectories are quite similar to each other. It means that the early part of Ours is accurate enough to enjoy the curriculum learning effect. The approximation error would increase as $k$ grows, but the figures show that it does not necessarily lead to worse solutions. It explains why the performances of Ours remain robust to the approximation error. 



\vspace{-0.05in}
\subsection{Image classification}
\vspace{-0.05in}
Next, we verify our method on a realistic large-scale and heterogeneous task distribution with multiple datasets.
\vspace{-0.1in}
\paragraph{Datasets.} We consider large-scale datasets with the number of instances roughly ranging from $5,000$ up to $100,000$. For images larger than $84 \times 84$, we resize their width and height into one of $\{28, 32, 64, 84\}$ for faster training. See \textbf{the supplementary file} for more information. For \textbf{meta-training}, we use $7$ datasets: \textbf{Tiny ImageNet}~\cite{tinyimagenet}, \textbf{CIFAR100}~\cite{krizhevsky2009learning}, \textbf{Stanford Dogs}~\cite{stanford_dogs}, \textbf{Aircraft}~\cite{maji2013fine}, \textbf{CUB}~\cite{WahCUB_200_2011}, \textbf{Fashion-MNIST}~\cite{xiao2017}, and \textbf{SVHN}~\cite{netzer2011reading}). Tiny ImageNet (TIN) and CIFAR100 are benchmark classification datasets of general categories. We class-wisely divide TIN into two splits. Other datasets include fine-grained classifications that would require a sufficient amount of task-specific adaptations (e.g. Aircraft), and grey-scale images (Fashion-MNIST). 
We \textbf{meta-test} the trained model on $5$ datasets: \textbf{Stanford Cars}, \textbf{QuickDraw}~\cite{DBLP:journals/corr/HaE17}, \textbf{VGG Flowers}~\cite{Nilsback08}, \textbf{VGG Pets}~\cite{parkhi12a}, and \textbf{STL10}, which are also highly heterogeneous. 

\vspace{-0.1in}
\paragraph{Experimental setup.} 
We use ResNet20 frequently used for images of size $32 \times 32$ (e.g. CIFAR datasets). We use random cropping and horizontal flipping as data augmentations, following the convention. For \textbf{meta-training}, we use the same $\alpha=0.01$, $K=1,000$, and $M=200$ for all the baselines and our model, except for $\beta$ that we found in the range of $\{10^{-3},10^{-2},10^{-1},10^{0},10^{1}\}$. We use SGD with momentum ($\mu=0.9$) and weight decay ($\lambda=0.0005$) as the inner optimizer. For \textbf{meta-testing}, we train $K=1,000$ steps for each dataset. We use SGD with Nesterov momentum optimizer ($\mu=0.9$) with an appropriate learning rate scheduling. The starting learning rate is $\alpha=0.1$ and we use $\lambda=0.0005$. See \textbf{the supplementary file} for more detail. The code is also publicly available\footnote{\url{https://github.com/JWoong148/ContinualTrajectoryShifting}}.

\vspace{-0.1in}
\paragraph{Baselines.}
We first compare with \textbf{Finetuning} baselines. We consider finetuning from the initialization obtained with multi-headed \textbf{multi-task learning (MTL)}, where we pretain a single shared feature extractor across the source tasks while the final dense layers are exclusive from one task to the others. We also consider finetuning from the initialization obtained by learning \textbf{only with Tiny ImageNet (TIN)} dataset in order to alleviate the negative transfer issue that may come with MTL. We next consider the following \textbf{meta-learning} methods. \textbf{FOMAML:} Meta-gradient of this method~\cite{finn2017model} is simply the last-step inner gradient. \textbf{FOMAML++:}~\cite{antoniou2018how} is a variant of FOMAML which periodically accumulates the intermediate meta-gradients (Multi-Step Loss Optimization in MAML++). \textbf{iMAML:}~\cite{rajeswaran2019meta} compute meta-gradient by estimating local curvature at the last step, based on Implicit Function Theorem. \textbf{Reptile:} Meta-gradient of this method~\cite{reptile} is defined as the average of differences between initialization and task-specific parameters. \textbf{Leap:} This method~\cite{leap} defines meta-objective as a sum of the task trajectory lengths, and its meta-gradient is computed with the similar first-order approximation.

\begin{figure}[t]
	\centering
	\vspace{-0.05in}
	\hspace{-0.2in}
	\hfill
	\subfigure[Meta-convergence]{
	    \raisebox{-0.05in}{
	    \includegraphics[height=3.8cm]{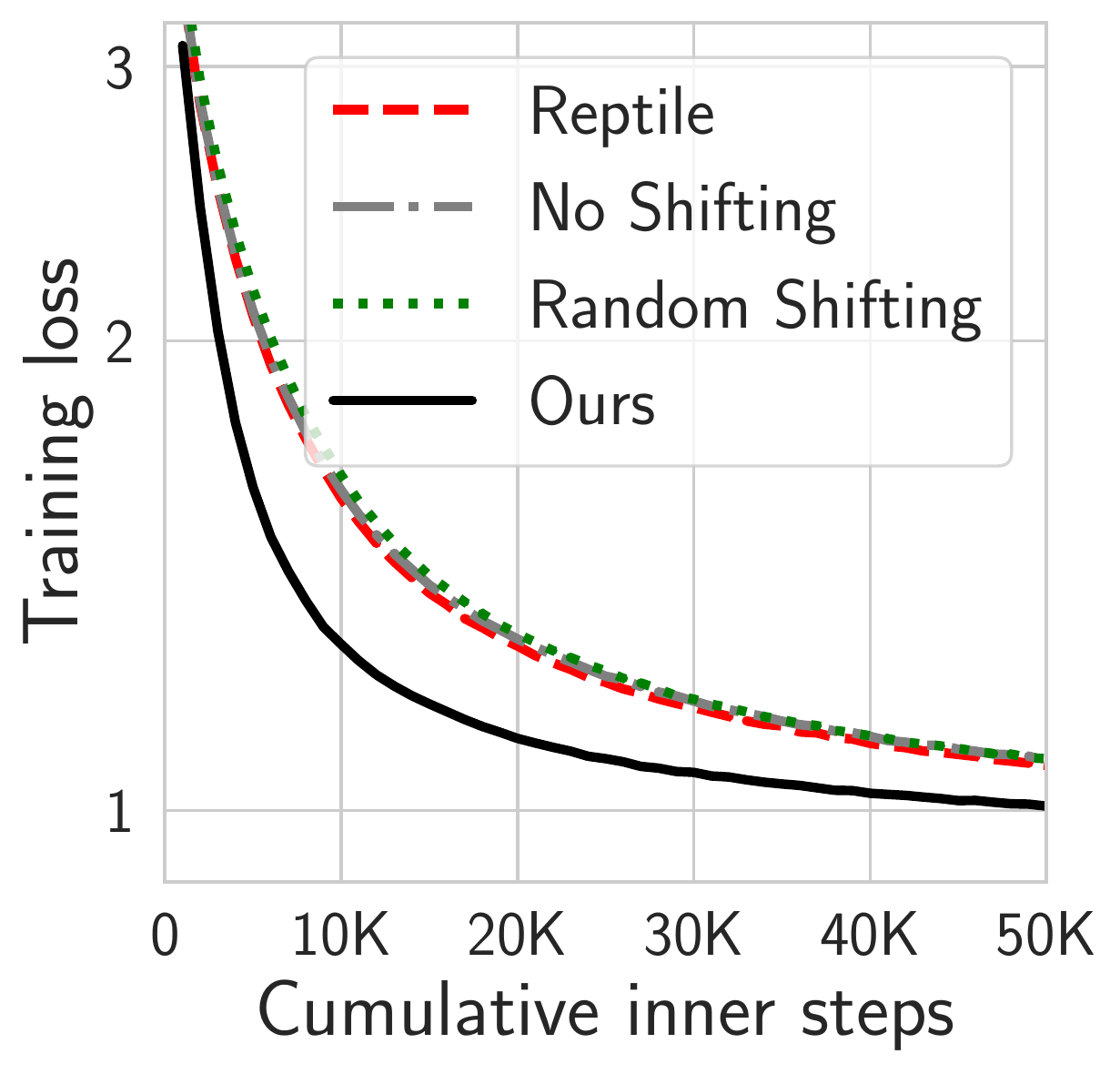}
	    }
	}
	\hspace{-0.2in}
	\hfill
    \subfigure[Meta-testing performance]{
	    \raisebox{-0.05in}{
	    \includegraphics[height=3.8cm]{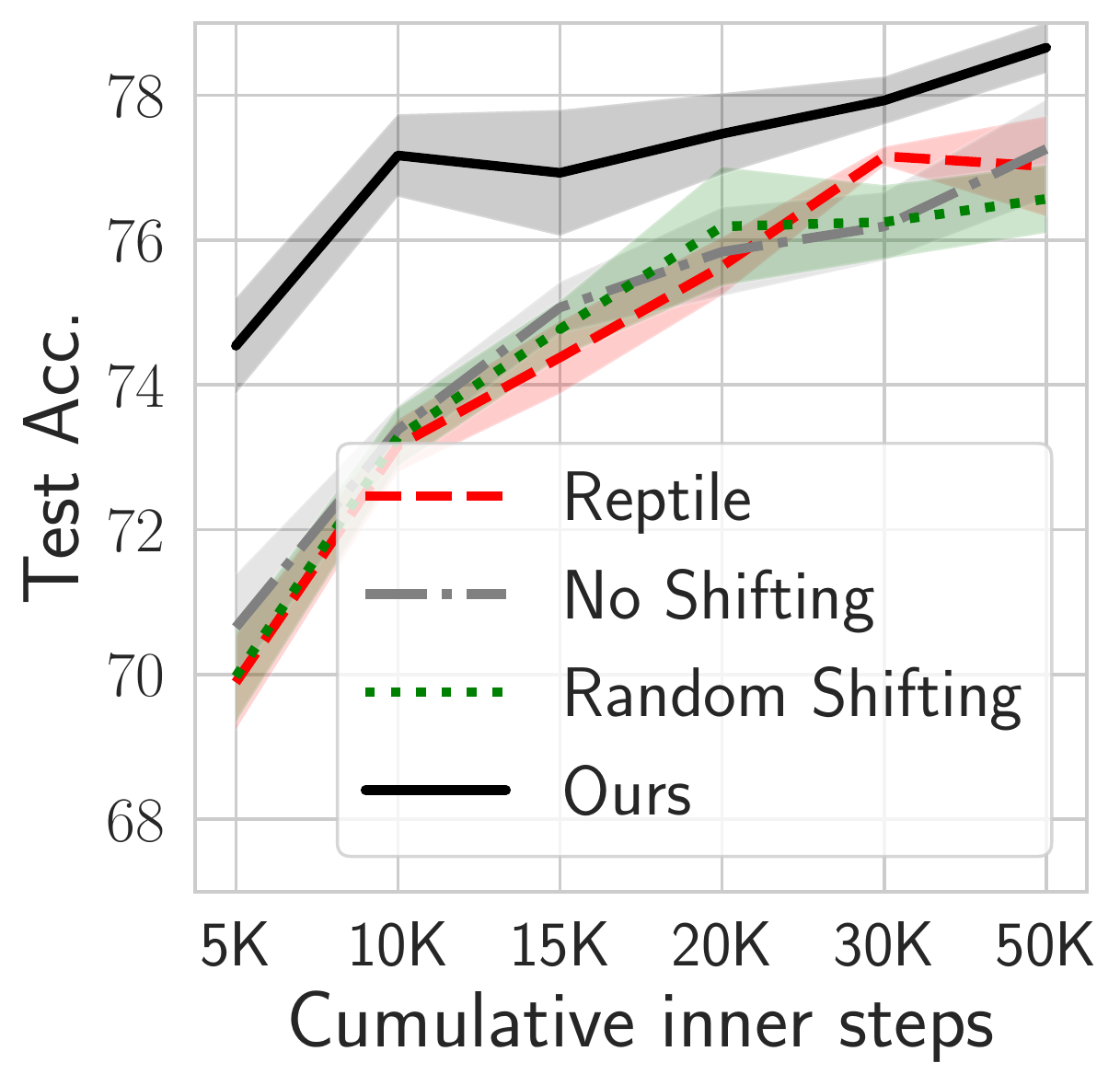}
	    }
	}
	\hspace{-0.2in}
	\hfill
	\vspace{-0.17in}
	\caption{\textbf{Ablation study}. (b) VGG Pets }
	\vspace{-0.2in}
	\label{fig:ablation}
\end{figure}

\begin{table*}[t]
\centering
\vspace{-0.05in}
\small
  \caption{\small\textbf{Classification accuracies obtained with various pre-training methods (\%)} when the target dataset contains 1,000 images. We report the mean accuracies and the $95\%$ confidence intervals over 5 runs.}
  \vspace{0.05in}
  \resizebox{1.\linewidth}{!}{
 \begin{tabular}{c | c c c c c c c c c |c}
 \hline
  ImageNet & \multicolumn{9}{c|}{Target dataset} & \\
pre-training & CIFAR100 & CIFAR10 & SVHN & Dogs & Pets & Flowers & Food & CUB & DTD & Avg.\\
 \hline
 \hline
+ None
& 41.95\tiny$\pm$0.29 
& 81.60\tiny$\pm$0.28 
& 60.09\tiny$\pm$0.98 
& 55.56\tiny$\pm$0.29 
& 83.48\tiny$\pm$0.15 
& 87.01\tiny$\pm$0.38 
& 36.95\tiny$\pm$0.37 
& 34.32\tiny$\pm$0.46 
& 59.39\tiny$\pm$0.53 
& 60.04
\\

+ MTL
& 42.79\tiny$\pm$0.54 
& 82.33\tiny$\pm$0.20 
& 59.05\tiny$\pm$1.09 
& 55.00\tiny$\pm$0.28 
& 83.29\tiny$\pm$0.25 
& 87.04\tiny$\pm$0.34 
& 36.84\tiny$\pm$0.37 
& 34.19\tiny$\pm$0.88 
& 58.86\tiny$\pm$0.49 
& 59.93
\\

+ Reptile
& 47.98\tiny$\pm$0.14 
& \bf 84.58\tiny$\pm$0.12 
& 62.39\tiny$\pm$0.72 
& 56.97\tiny$\pm$0.12 
& 84.25\tiny$\pm$0.22 
& 87.22\tiny$\pm$0.31 
& 37.35\tiny$\pm$0.22 
& 35.44\tiny$\pm$0.48 
& 58.98\tiny$\pm$0.59 
& 61.68
\\

\hline
\bf + Ours
& \bf 48.34\tiny$\pm$0.21 
& 84.42\tiny$\pm$0.15 
& \bf 62.82\tiny$\pm$0.56 
& \bf 57.53\tiny$\pm$0.48 
& \bf 84.65\tiny$\pm$0.11 
& \bf 87.54\tiny$\pm$0.21 
& \bf 37.84\tiny$\pm$0.20 
& \bf 36.40\tiny$\pm$0.20 
& \bf 59.53\tiny$\pm$0.25 
& \bf 62.12

\\

\hline
\end{tabular}
}
  \label{tbl:imagenet}
  \vspace{-0.18in}
\end{table*}

\vspace{-0.1in}
\paragraph{Results and analysis.}
First of all, we see from Figure~\ref{fig:meta_convergence} that Ours achieve much faster meta-convergence than other meta-learning methods, thanks to more frequent meta-updates with the proposed continual trajectory shifting. Our method thus achieves competitive meta-testing performances significantly faster than other meta-learning methods (Figure~\ref{fig:new6}(a-e)\footnote{Note that x-axis is cumulative inner steps at meta-training, not training steps at meta-testing.}). 
Note that Reptile significantly outperforms Leap in our experiments. We carefully tuned the meta-learning rate of each method, and found that Reptile allows much greater meta-learning rate ($\beta=1.0$) than Leap ($\beta=0.1$). 

We also compare with the other baselines in Figure~\ref{fig:bar}. FOMAML, FOMAML++, and iMAML perform much worse than other baselines. For FOMAML, its meta-gradient is simply the last-step inner-gradient, which can be arbitrarily uninformative for the meta-learner~\cite{leap}. iMAML estimates the meta gradient at the last step by implicitly incorporating the learning trajectory based on Implicit Function Theorem, but the results tell that the method is not as effective as explicit methods such as Reptile. FOMAML++ outperforms FOMAML, demonstrating the importance of considering the whole inner-trajectory when computing the meta-gradients~\cite{leap}. 

For the finetuning baselines, finetuning with MTL significantly underperforms the finetuning with only TIN. It demonstrates the effect of negative transfer problem that frequently happens when we jointly learn with multiple heterogeneous datasets. On the other hand, our method outperforms both of the finetuning baselines, indicating that meta-learning of the shared initialization can be an effective alternative for the negative transfer problem, instead of finding a jointly optimal feature extractor for all the tasks. Lastly, Figure~\ref{fig:varyingK_pets} and~\ref{fig:varyingK_stl10} shows that the performance improves as we increase the inner trajectory length up to $K=1,000$, demonstrating the effect of short horizon bias (See also Figure~\ref{fig:largeK}). 

\vspace{-0.1in}
\paragraph{Ablation study.}
We perform the ablation study whether the proposed shifting for the task-learning trajectories (Line 13 in Algorithm~\ref{algo:ours}) is the source of performance improvements. We see from Figure~\ref{fig:ablation} that our model without shifting (\textbf{No Shifting}) or the shifting with the same magnitude but with random direction (\textbf{Random Shifting}) performs almost the same as Reptile, demonstrating the effectiveness of the proposed shifting rule.

\vspace{-0.05in}
\subsection{Improving on ImageNet Pre-trained Model}
\vspace{-0.05in}
We demonstrate that our method is capable of improving on the ImageNet finetuning under limited data regime. 

\vspace{-0.1in}
\paragraph{Datasets.} For \textbf{meta-training}, we construct a heterogeneous data distribution by class-wisely dividing the original ImageNet dataset into $8$ subsets based on the WordNet class hierarchy. We then meta-train the model over the obtained subsets. See Figure 4 and Table 4 in the \textbf{Supplementary file} for more information. We then \textbf{meta-test} with the $9$ benchmark image classification datasets described in Table 5 in the \textbf{Supplementary file}. 

\begin{figure}[t!]
    \vspace{-0.05in}
	\centering
    \hskip -0.2in
	\hfill
	\subfigure[Pets]{
	    \includegraphics[height=3.3cm]{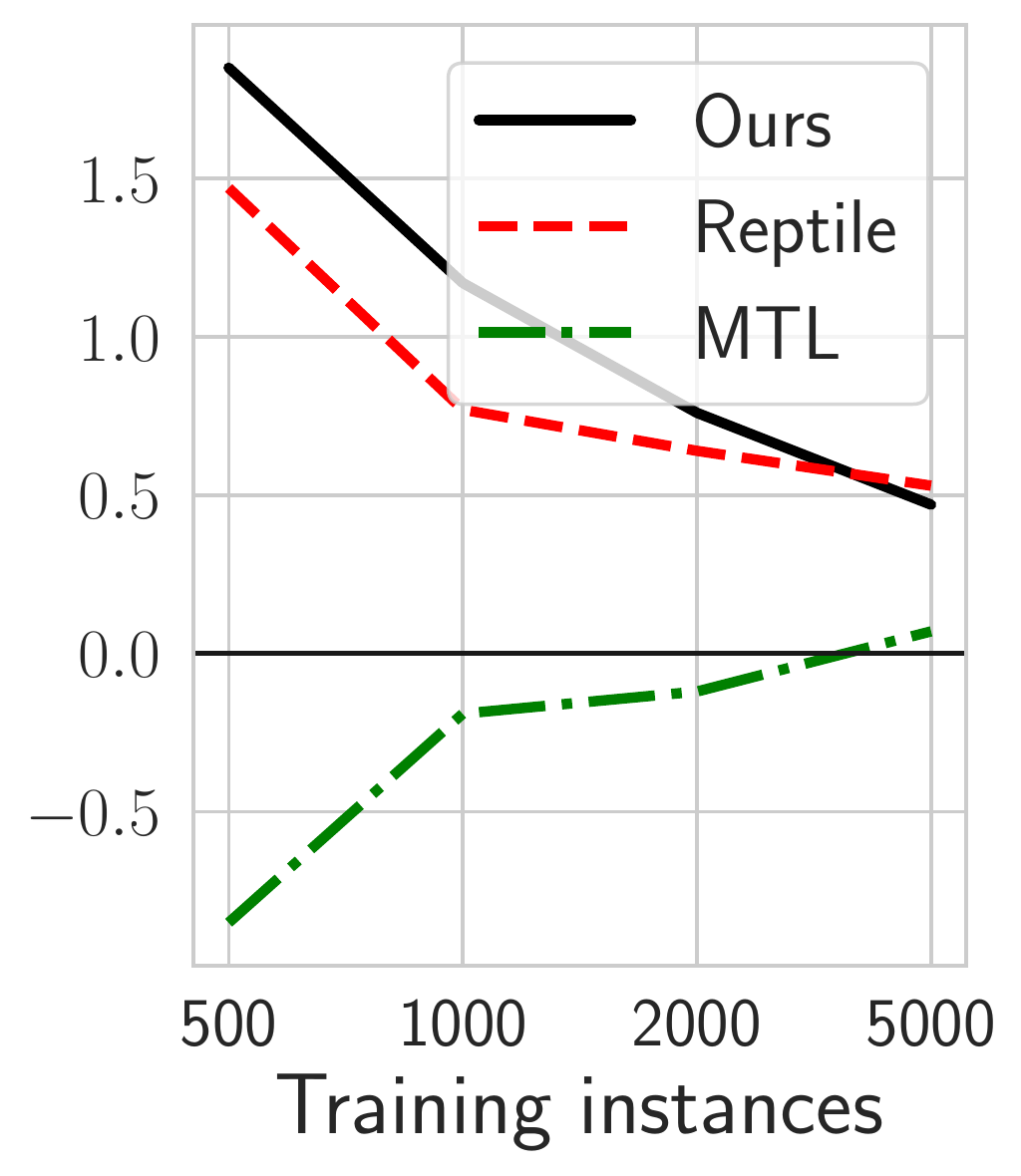}
    }
    \hskip -0.2in
	\hfill
	\subfigure[Food]{
	    \includegraphics[height=3.3cm]{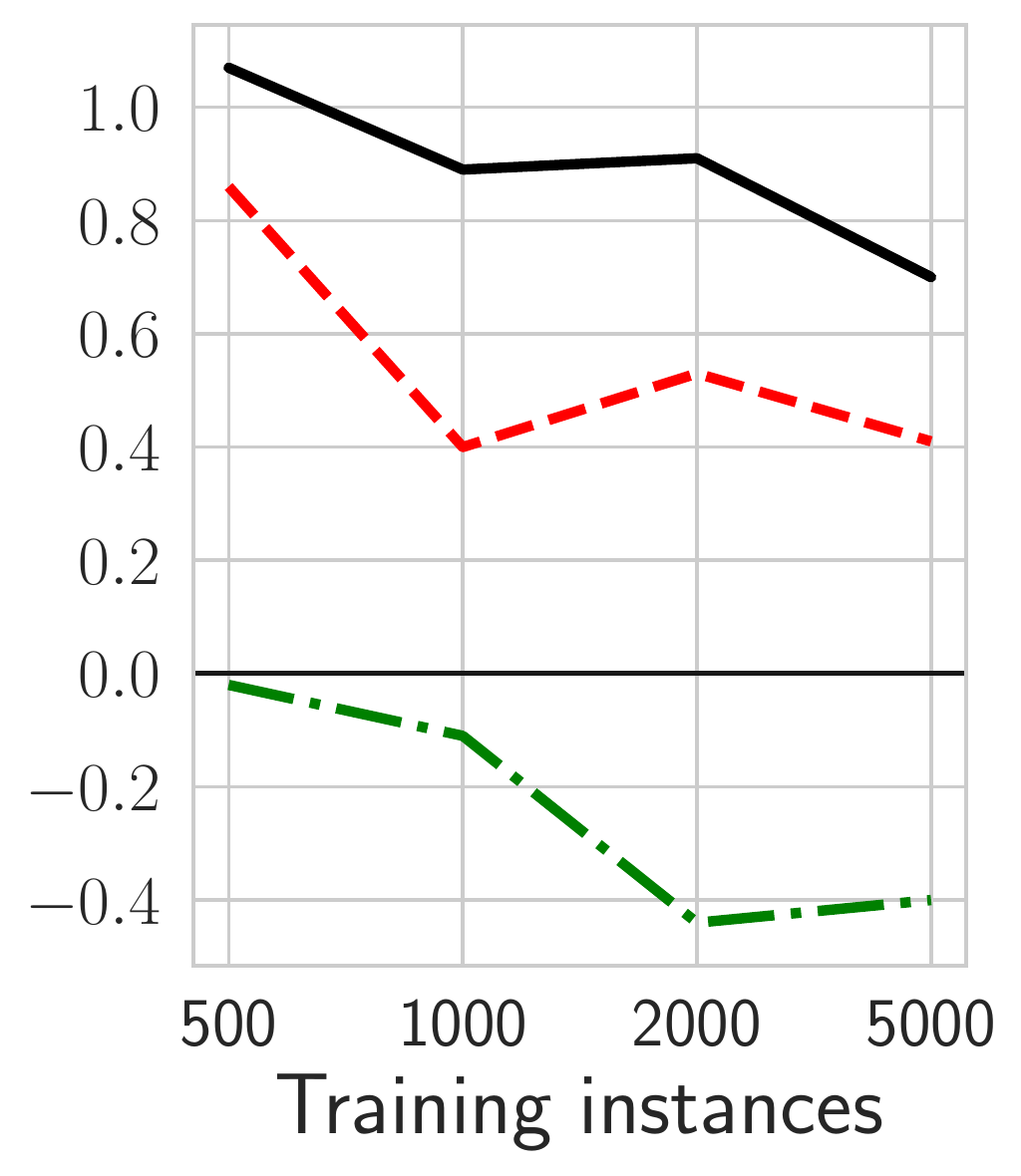}
    }
    \hskip -0.2in
	\hfill
	\subfigure[CUB]{
	    \includegraphics[height=3.3cm]{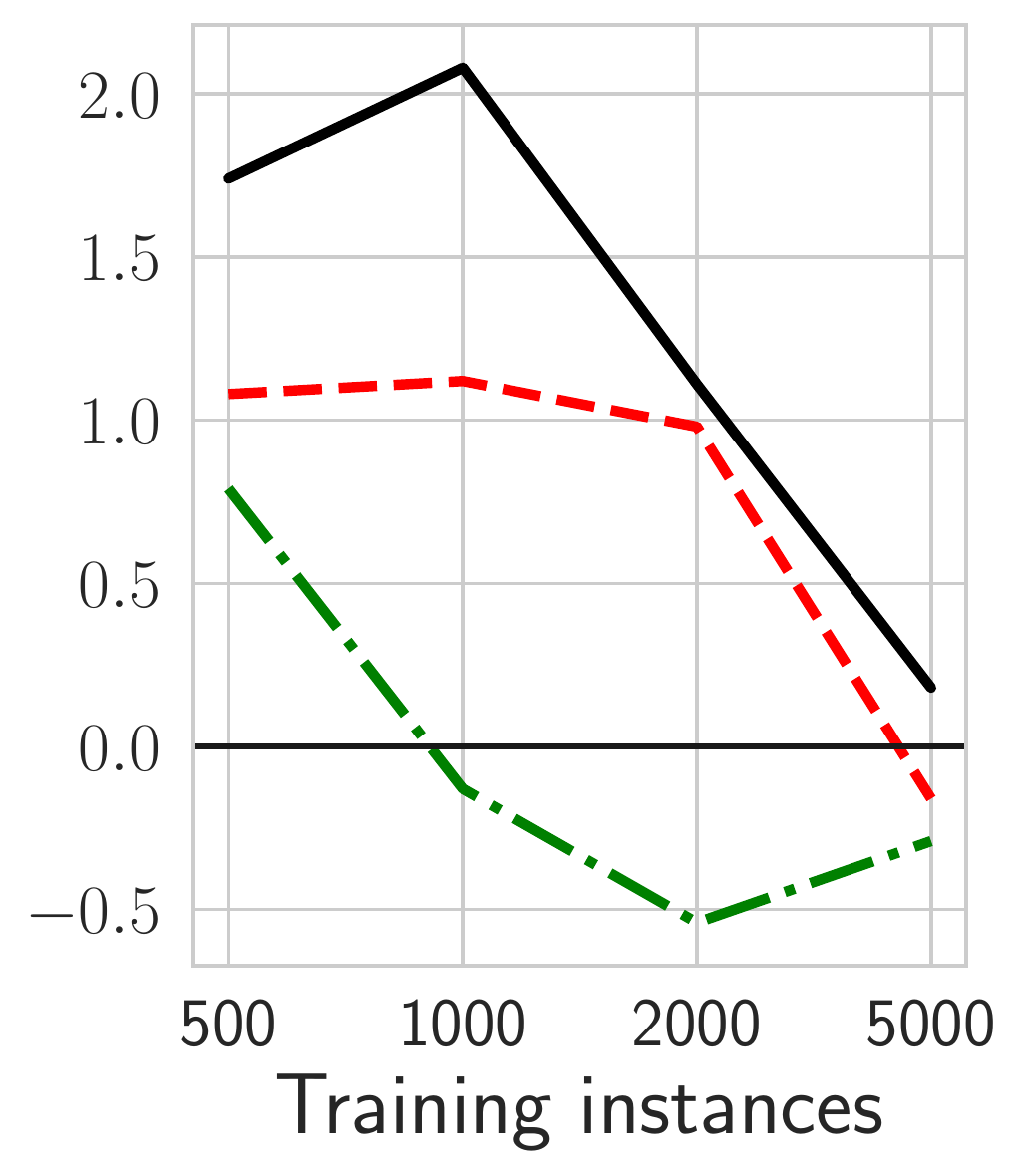}
    }
    \hskip -0.2in
	\hfill
	\vspace{-0.15in}
	\caption{\small Accuracy improvements (\%) over ImageNet Finetuning. }
	\vspace{-0.15in}
    \label{fig:imagenet_instances}
\end{figure} 

\vspace{-0.1in}
\paragraph{Experimental setup.} We use ResNet18~\cite{he2016deep} suitable for the images of size $224\times 224$. We use random cropping and horizontal flipping as data augmentations. \newline\textbf{Meta-training:} For MTL, Reptile and our model, we start from the ImageNet pretrained model for the meta-training to converge faster and reach a better solution than meta-training from scratch. We use SGD with momentum ($\mu=0.9$) as the inner-optimizer without applying the weight decay ($\lambda=0$). We set the batch size to $256$.  For \textbf{MTL}, we set the learning rate to $0.01$ and train for $50,000$ steps. For \textbf{Reptile} and \textbf{our model}, we use $\alpha=0.01$, $K=1,000$, and $M=50$, but use different meta-learning rate ($\beta=1$ for Reptile and $\beta=0.001$ for our model), since the optimal meta-learning rate differs across two methods. We meta-train for $50,000$ steps. \textbf{Meta-testing:} We subsample each target dataset such that each contains $1,000$ training datapoints. We normalize each input image with the mean and standard deviation of RGB channels across the whole training datapoints. We set the batch size to $256$. We train $K=1,000$ steps with Nesterov momentum optimizer ($\mu=0.9$). The starting learning rate is $\alpha=0.01$, and we step-wisely decay $\alpha$ at $400$, $700$, and $900$ steps by multiplying $0.2$. We use the weight decay $\lambda=0.0001$.
    
    
    \vspace{-0.15in}
    \paragraph{Results and analysis} Table~\ref{tbl:imagenet} shows the results. We empirically observe that our method is more effective when the task-specific learning suffers from overfitting. In order to clearly see this effect, we subsample each target dataset to contain only $1,000$ images and compare the performances. Table~\ref{tbl:imagenet} shows that our method consistently outperforms the baselines across most of the datasets when each target dataset has a limited number of instances. Figure~\ref{fig:imagenet_instances} confirms that on most of the target datasets, the performance improvements from the base ImageNet finetuning increase as the size of each dataset gets smaller. We conjecture that the performance improvements comes from the \emph{smoother} initial model parameter learned with our meta-learning algorithm, which may correspond to a stronger prior over the model parameters that can effectively regularize the task-specific learning with the small datasets. 

\vspace{-0.02in}
\section{Conclusion}
\vspace{-0.0in}
In this paper, we tackled the challenging problem of large-scale meta-learning. We first showed that a large number of inner-gradient steps allows to capture the structure of large-scale meta-learning well. We then improve the meta-learning efficiency with the \emph{continual trajectory shifting}, which continuously shifts the inner-learning trajectories w.r.t. the frequent update of the initialization point. By doing so, unlike the previous meta-learning algorithms, the task-learners no longer need to reinitiate the learning trajectory for every meta-update, thereby allowing to arbitrarily increase the meta-update frequency. 
We investigated why and how our model works well with synthetic experiment and also validated the effectiveness of our method on the large-scale experiments with image datasets. We believe that our work make a meaningful step toward applying meta-learning to large-scale real-world tasks. 
\vspace{-0.05in}
\section*{Acknowledgement}
\vspace{-0.05in}
    This work was supported by Google AI Focused Research Award, the Engineering Research Center Program through the National Research Foundation of Korea (NRF) funded by the Korean Government MSIT (NRF-2018R1A5A1059921), and the Institute of Information \& communications Technology Planning \& Evaluation (IITP) grant funded by the Korea government(MSIT) (No.2019-0-00075, Artificial Intelligence Graduate School Program(KAIST))

\bibliography{icml2021}

\begin{thebibliography}{42}
\providecommand{\natexlab}[1]{#1}
\providecommand{\url}[1]{\texttt{#1}}
\expandafter\ifx\csname urlstyle\endcsname\relax
  \providecommand{\doi}[1]{doi: #1}\else
  \providecommand{\doi}{doi: \begingroup \urlstyle{rm}\Url}\fi

\bibitem[Antoniou et~al.(2018)Antoniou, Edwards, and Storkey]{antoniou2018how}
Antoniou, A., Edwards, H., and Storkey, A.
\newblock How to train your maml.
\newblock In \emph{International Conference on Learning Representations}, 2018.

\bibitem[Balduzzi et~al.(2017{\natexlab{a}})Balduzzi, Frean, Leary, Lewis, Ma,
  and McWilliams]{balduzzi17b}
Balduzzi, D., Frean, M., Leary, L., Lewis, J., Ma, K. W.-D., and McWilliams, B.
\newblock The shattered gradients problem: If resnets are the answer, then what
  is the question?
\newblock In \emph{International Conference on Machine Learning}, pp.\
  342--350. PMLR, 2017{\natexlab{a}}.

\bibitem[Balduzzi et~al.(2017{\natexlab{b}})Balduzzi, McWilliams, and
  Butler-Yeoman]{balduzzi17c}
Balduzzi, D., McWilliams, B., and Butler-Yeoman, T.
\newblock Neural taylor approximations: Convergence and exploration in
  rectifier networks.
\newblock In \emph{International Conference on Machine Learning}, pp.\
  351--360. PMLR, 2017{\natexlab{b}}.

\bibitem[Bengio et~al.(2009)Bengio, Louradour, Collobert, and
  Weston]{BengioLCW09}
Bengio, Y., Louradour, J., Collobert, R., and Weston, J.
\newblock Curriculum learning.
\newblock In \emph{Proceedings of the 26th annual international conference on
  machine learning}, pp.\  41--48, 2009.

\bibitem[Coates et~al.(2011)Coates, Ng, and Lee]{coates2011analysis}
Coates, A., Ng, A., and Lee, H.
\newblock An analysis of single-layer networks in unsupervised feature
  learning.
\newblock In \emph{Proceedings of the fourteenth international conference on
  artificial intelligence and statistics}, pp.\  215--223. JMLR Workshop and
  Conference Proceedings, 2011.

\bibitem[Dhillon et~al.(2019)Dhillon, Chaudhari, Ravichandran, and
  Soatto]{Dhillon2020A}
Dhillon, G.~S., Chaudhari, P., Ravichandran, A., and Soatto, S.
\newblock A baseline for few-shot image classification.
\newblock In \emph{International Conference on Learning Representations}, 2019.

\bibitem[Finn et~al.(2017)Finn, Abbeel, and Levine]{finn2017model}
Finn, C., Abbeel, P., and Levine, S.
\newblock Model-agnostic meta-learning for fast adaptation of deep networks.
\newblock In \emph{International conference on machine learning}, pp.\
  1126--1135. PMLR, 2017.

\bibitem[Flennerhag et~al.(2018)Flennerhag, Moreno, Lawrence, and
  Damianou]{leap}
Flennerhag, S., Moreno, P.~G., Lawrence, N.~D., and Damianou, A.
\newblock Transferring knowledge across learning processes.
\newblock In \emph{International Conference on Learning Representations}, 2018.

\bibitem[Flennerhag et~al.(2019)Flennerhag, Rusu, Pascanu, Visin, Yin, and
  Hadsell]{Flennerhag2020Meta-Learning}
Flennerhag, S., Rusu, A.~A., Pascanu, R., Visin, F., Yin, H., and Hadsell, R.
\newblock Meta-learning with warped gradient descent.
\newblock In \emph{International Conference on Learning Representations}, 2019.

\bibitem[Ha \& Eck(2017)Ha and Eck]{DBLP:journals/corr/HaE17}
Ha, D. and Eck, D.
\newblock A neural representation of sketch drawings.
\newblock \emph{arXiv preprint arXiv:1704.03477}, 2017.

\bibitem[He et~al.(2016)He, Zhang, Ren, and Sun]{he2016deep}
He, K., Zhang, X., Ren, S., and Sun, J.
\newblock Deep residual learning for image recognition.
\newblock In \emph{Proceedings of the IEEE conference on computer vision and
  pattern recognition}, pp.\  770--778, 2016.

\bibitem[Jang et~al.(2019)Jang, Lee, Hwang, and Shin]{Jang2019LearningWA}
Jang, Y., Lee, H., Hwang, S.~J., and Shin, J.
\newblock Learning what and where to transfer.
\newblock In \emph{International Conference on Machine Learning}, pp.\
  3030--3039. PMLR, 2019.

\bibitem[Khosla et~al.(2011)Khosla, Jayadevaprakash, Yao, and
  Li]{stanford_dogs}
Khosla, A., Jayadevaprakash, N., Yao, B., and Li, F.-F.
\newblock Novel dataset for fine-grained image categorization: Stanford dogs.
\newblock In \emph{Proc. CVPR Workshop on Fine-Grained Visual Categorization
  (FGVC)}, volume~2. Citeseer, 2011.

\bibitem[Kornblith et~al.(2019)Kornblith, Shlens, and Le]{kornblith2019do}
Kornblith, S., Shlens, J., and Le, Q.~V.
\newblock Do better imagenet models transfer better?
\newblock In \emph{Proceedings of the IEEE/CVF conference on computer vision
  and pattern recognition}, pp.\  2661--2671, 2019.

\bibitem[Krause et~al.(2013)Krause, Stark, Deng, and Fei-Fei]{stanford_cars}
Krause, J., Stark, M., Deng, J., and Fei-Fei, L.
\newblock 3d object representations for fine-grained categorization.
\newblock In \emph{Proceedings of the IEEE international conference on computer
  vision workshops}, pp.\  554--561, 2013.

\bibitem[Krizhevsky(2009)]{krizhevsky2009learning}
Krizhevsky, A.
\newblock Learning multiple layers of features from tiny images.
\newblock 2009.

\bibitem[Lake et~al.(2015)Lake, Salakhutdinov, and Tenenbaum]{lake2015human}
Lake, B.~M., Salakhutdinov, R., and Tenenbaum, J.~B.
\newblock Human-level concept learning through probabilistic program induction.
\newblock \emph{Science}, 350\penalty0 (6266):\penalty0 1332--1338, 2015.

\bibitem[Le \& Yang(2015)Le and Yang]{tinyimagenet}
Le, Y. and Yang, X.~S.
\newblock Tiny imagenet visual recognition challenge.
\newblock 2015.

\bibitem[Lee et~al.(2019)Lee, Maji, Ravichandran, and Soatto]{Lee_2019_CVPR}
Lee, K., Maji, S., Ravichandran, A., and Soatto, S.
\newblock Meta-learning with differentiable convex optimization.
\newblock In \emph{Proceedings of the IEEE/CVF Conference on Computer Vision
  and Pattern Recognition}, pp.\  10657--10665, 2019.

\bibitem[Lee \& Choi(2018)Lee and Choi]{lee2018gradient}
Lee, Y. and Choi, S.
\newblock Gradient-based meta-learning with learned layerwise metric and
  subspace.
\newblock In \emph{International Conference on Machine Learning}, pp.\
  2927--2936. PMLR, 2018.

\bibitem[Liu et~al.(2018)Liu, Lee, Park, Kim, Yang, Hwang, and
  Yang]{liu2018learning}
Liu, Y., Lee, J., Park, M., Kim, S., Yang, E., Hwang, S.~J., and Yang, Y.
\newblock Learning to propagate labels: Transductive propagation network for
  few-shot learning.
\newblock In \emph{International Conference on Learning Representations}, 2018.

\bibitem[Maji et~al.(2013)Maji, Rahtu, Kannala, Blaschko, and
  Vedaldi]{maji2013fine}
Maji, S., Rahtu, E., Kannala, J., Blaschko, M., and Vedaldi, A.
\newblock Fine-grained visual classification of aircraft.
\newblock \emph{arXiv preprint arXiv:1306.5151}, 2013.

\bibitem[Mishra et~al.(2018)Mishra, Rohaninejad, Chen, and Abbeel]{mishra2018a}
Mishra, N., Rohaninejad, M., Chen, X., and Abbeel, P.
\newblock A simple neural attentive meta-learner.
\newblock In \emph{International Conference on Learning Representations}, 2018.

\bibitem[Netzer et~al.(2011)Netzer, Wang, Coates, Bissacco, Wu, and
  Ng]{netzer2011reading}
Netzer, Y., Wang, T., Coates, A., Bissacco, A., Wu, B., and Ng, A.~Y.
\newblock Reading digits in natural images with unsupervised feature learning.
\newblock 2011.

\bibitem[Nichol et~al.(2018)Nichol, Achiam, and Schulman]{reptile}
Nichol, A., Achiam, J., and Schulman, J.
\newblock On first-order meta-learning algorithms.
\newblock \emph{arXiv preprint arXiv:1803.02999}, 2018.

\bibitem[Nilsback \& Zisserman(2008)Nilsback and Zisserman]{Nilsback08}
Nilsback, M.-E. and Zisserman, A.
\newblock Automated flower classification over a large number of classes.
\newblock In \emph{2008 Sixth Indian Conference on Computer Vision, Graphics \&
  Image Processing}, pp.\  722--729. IEEE, 2008.

\bibitem[Oreshkin et~al.(2018)Oreshkin, Rodr{\'\i}guez~L{\'o}pez, and
  Lacoste]{oreshkin18}
Oreshkin, B., Rodr{\'\i}guez~L{\'o}pez, P., and Lacoste, A.
\newblock Tadam: Task dependent adaptive metric for improved few-shot learning.
\newblock \emph{Advances in neural information processing systems}, 31, 2018.

\bibitem[Parkhi et~al.(2012)Parkhi, Vedaldi, Zisserman, and Jawahar]{parkhi12a}
Parkhi, O.~M., Vedaldi, A., Zisserman, A., and Jawahar, C.
\newblock Cats and dogs.
\newblock In \emph{2012 IEEE conference on computer vision and pattern
  recognition}, pp.\  3498--3505. IEEE, 2012.

\bibitem[Rajeswaran et~al.(2019)Rajeswaran, Finn, Kakade, and
  Levine]{rajeswaran2019meta}
Rajeswaran, A., Finn, C., Kakade, S.~M., and Levine, S.
\newblock Meta-learning with implicit gradients.
\newblock \emph{Advances in neural information processing systems}, 32, 2019.

\bibitem[Ravi \& Larochelle(2016)Ravi and Larochelle]{ravi2016optimization}
Ravi, S. and Larochelle, H.
\newblock Optimization as a model for few-shot learning.
\newblock In \emph{International Conference on Learning Representations}, 2016.

\bibitem[Russakovsky et~al.(2015)Russakovsky, Deng, Su, Krause, Satheesh, Ma,
  Huang, Karpathy, Khosla, Bernstein, et~al.]{russakovsky2015imagenet}
Russakovsky, O., Deng, J., Su, H., Krause, J., Satheesh, S., Ma, S., Huang, Z.,
  Karpathy, A., Khosla, A., Bernstein, M., et~al.
\newblock Imagenet large scale visual recognition challenge.
\newblock \emph{International journal of computer vision}, 115\penalty0
  (3):\penalty0 211--252, 2015.

\bibitem[Rusu et~al.(2018)Rusu, Rao, Sygnowski, Vinyals, Pascanu, Osindero, and
  Hadsell]{rusu2018meta}
Rusu, A.~A., Rao, D., Sygnowski, J., Vinyals, O., Pascanu, R., Osindero, S.,
  and Hadsell, R.
\newblock Meta-learning with latent embedding optimization.
\newblock In \emph{International Conference on Learning Representations}, 2018.

\bibitem[Santoro et~al.(2016)Santoro, Bartunov, Botvinick, Wierstra, and
  Lillicrap]{santoro2016meta}
Santoro, A., Bartunov, S., Botvinick, M., Wierstra, D., and Lillicrap, T.
\newblock Meta-learning with memory-augmented neural networks.
\newblock In \emph{International conference on machine learning}, pp.\
  1842--1850. PMLR, 2016.

\bibitem[Schmidhuber(1987)]{schmidhuber1987evolutionary}
Schmidhuber, J.
\newblock \emph{Evolutionary principles in self-referential learning, or on
  learning how to learn: the meta-meta-... hook}.
\newblock PhD thesis, Technische Universit{\"a}t M{\"u}nchen, 1987.

\bibitem[Snell et~al.(2017)Snell, Swersky, and Zemel]{snell2017prototypical}
Snell, J., Swersky, K., and Zemel, R.
\newblock Prototypical networks for few-shot learning.
\newblock \emph{Advances in neural information processing systems}, 30, 2017.

\bibitem[Song et~al.(2019)Song, Gao, Yang, Choromanski, Pacchiano, and
  Tang]{Song2020ES-MAML}
Song, X., Gao, W., Yang, Y., Choromanski, K., Pacchiano, A., and Tang, Y.
\newblock Es-maml: Simple hessian-free meta learning.
\newblock In \emph{International Conference on Learning Representations}, 2019.

\bibitem[Sung et~al.(2018)Sung, Yang, Zhang, Xiang, Torr, and
  Hospedales]{yang2017learning}
Sung, F., Yang, Y., Zhang, L., Xiang, T., Torr, P.~H., and Hospedales, T.~M.
\newblock Learning to compare: Relation network for few-shot learning.
\newblock In \emph{Proceedings of the IEEE conference on computer vision and
  pattern recognition}, pp.\  1199--1208, 2018.

\bibitem[Thrun \& Pratt(2012)Thrun and Pratt]{thrun98}
Thrun, S. and Pratt, L.
\newblock \emph{Learning to learn}.
\newblock Springer Science \& Business Media, 2012.

\bibitem[Vinyals et~al.(2016)Vinyals, Blundell, Lillicrap, Wierstra,
  et~al.]{vinyals2016matching}
Vinyals, O., Blundell, C., Lillicrap, T., Wierstra, D., et~al.
\newblock Matching networks for one shot learning.
\newblock \emph{Advances in neural information processing systems}, 29, 2016.

\bibitem[Wah et~al.(2011)Wah, Branson, Welinder, Perona, and
  Belongie]{WahCUB_200_2011}
Wah, C., Branson, S., Welinder, P., Perona, P., and Belongie, S.
\newblock The caltech-ucsd birds-200-2011 dataset.
\newblock 2011.

\bibitem[Wu et~al.(2018)Wu, Ren, Liao, and Grosse]{wu2018understanding}
Wu, Y., Ren, M., Liao, R., and Grosse, R.
\newblock Understanding short-horizon bias in stochastic meta-optimization.
\newblock In \emph{International Conference on Learning Representations}, 2018.

\bibitem[Xiao et~al.(2017)Xiao, Rasul, and Vollgraf]{xiao2017}
Xiao, H., Rasul, K., and Vollgraf, R.
\newblock Fashion-mnist: a novel image dataset for benchmarking machine
  learning algorithms.
\newblock \emph{arXiv preprint arXiv:1708.07747}, 2017.

\end{thebibliography}


\begin{thebibliography}{19}
\providecommand{\natexlab}[1]{#1}
\providecommand{\url}[1]{\texttt{#1}}
\expandafter\ifx\csname urlstyle\endcsname\relax
  \providecommand{\doi}[1]{doi: #1}\else
  \providecommand{\doi}{doi: \begingroup \urlstyle{rm}\Url}\fi

\bibitem[Bossard et~al.(2014)Bossard, Guillaumin, and Gool]{BossardGG14}
Bossard, L., Guillaumin, M., and Gool, L.~V.
\newblock Food-101--mining discriminative components with random forests.
\newblock In \emph{European conference on computer vision}, pp.\  446--461.
  Springer, 2014.

\bibitem[Cimpoi et~al.(2014)Cimpoi, Maji, Kokkinos, Mohamed, and
  Vedaldi]{cimpoi14describing}
Cimpoi, M., Maji, S., Kokkinos, I., Mohamed, S., and Vedaldi, A.
\newblock Describing textures in the wild.
\newblock In \emph{Proceedings of the IEEE conference on computer vision and
  pattern recognition}, pp.\  3606--3613, 2014.

\bibitem[Coates et~al.(2011)Coates, Ng, and Lee]{coates2011analysis}
Coates, A., Ng, A., and Lee, H.
\newblock An analysis of single-layer networks in unsupervised feature
  learning.
\newblock In \emph{Proceedings of the fourteenth international conference on
  artificial intelligence and statistics}, pp.\  215--223. JMLR Workshop and
  Conference Proceedings, 2011.

\bibitem[Finn et~al.(2017)Finn, Abbeel, and Levine]{finn2017model}
Finn, C., Abbeel, P., and Levine, S.
\newblock Model-agnostic meta-learning for fast adaptation of deep networks.
\newblock In \emph{International conference on machine learning}, pp.\
  1126--1135. PMLR, 2017.

\bibitem[Flennerhag et~al.(2018)Flennerhag, Moreno, Lawrence, and
  Damianou]{leap}
Flennerhag, S., Moreno, P.~G., Lawrence, N.~D., and Damianou, A.
\newblock Transferring knowledge across learning processes.
\newblock In \emph{International Conference on Learning Representations}, 2018.

\bibitem[Ha \& Eck(2017)Ha and Eck]{DBLP:journals/corr/HaE17}
Ha, D. and Eck, D.
\newblock A neural representation of sketch drawings.
\newblock \emph{arXiv preprint arXiv:1704.03477}, 2017.

\bibitem[Khosla et~al.(2011)Khosla, Jayadevaprakash, Yao, and
  Li]{stanford_dogs}
Khosla, A., Jayadevaprakash, N., Yao, B., and Li, F.-F.
\newblock Novel dataset for fine-grained image categorization: Stanford dogs.
\newblock In \emph{Proc. CVPR Workshop on Fine-Grained Visual Categorization
  (FGVC)}, volume~2. Citeseer, 2011.

\bibitem[Kingma \& Ba(2014)Kingma and Ba]{kingma2014adam}
Kingma, D.~P. and Ba, J.
\newblock Adam: A method for stochastic optimization.
\newblock \emph{arXiv preprint arXiv:1412.6980}, 2014.

\bibitem[Krause et~al.(2013)Krause, Stark, Deng, and Fei-Fei]{stanford_cars}
Krause, J., Stark, M., Deng, J., and Fei-Fei, L.
\newblock 3d object representations for fine-grained categorization.
\newblock In \emph{Proceedings of the IEEE international conference on computer
  vision workshops}, pp.\  554--561, 2013.

\bibitem[Krizhevsky(2009)]{krizhevsky2009learning}
Krizhevsky, A.
\newblock Learning multiple layers of features from tiny images.
\newblock 2009.

\bibitem[Le \& Yang(2015)Le and Yang]{tinyimagenet}
Le, Y. and Yang, X.~S.
\newblock Tiny imagenet visual recognition challenge.
\newblock 2015.

\bibitem[Maji et~al.(2013)Maji, Rahtu, Kannala, Blaschko, and
  Vedaldi]{maji2013fine}
Maji, S., Rahtu, E., Kannala, J., Blaschko, M., and Vedaldi, A.
\newblock Fine-grained visual classification of aircraft.
\newblock \emph{arXiv preprint arXiv:1306.5151}, 2013.

\bibitem[Netzer et~al.(2011)Netzer, Wang, Coates, Bissacco, Wu, and
  Ng]{netzer2011reading}
Netzer, Y., Wang, T., Coates, A., Bissacco, A., Wu, B., and Ng, A.~Y.
\newblock Reading digits in natural images with unsupervised feature learning.
\newblock 2011.

\bibitem[Nichol et~al.(2018)Nichol, Achiam, and Schulman]{reptile}
Nichol, A., Achiam, J., and Schulman, J.
\newblock On first-order meta-learning algorithms.
\newblock \emph{arXiv preprint arXiv:1803.02999}, 2018.

\bibitem[Nilsback \& Zisserman(2008)Nilsback and Zisserman]{Nilsback08}
Nilsback, M.-E. and Zisserman, A.
\newblock Automated flower classification over a large number of classes.
\newblock In \emph{2008 Sixth Indian Conference on Computer Vision, Graphics \&
  Image Processing}, pp.\  722--729. IEEE, 2008.

\bibitem[Parkhi et~al.(2012)Parkhi, Vedaldi, Zisserman, and Jawahar]{parkhi12a}
Parkhi, O.~M., Vedaldi, A., Zisserman, A., and Jawahar, C.
\newblock Cats and dogs.
\newblock In \emph{2012 IEEE conference on computer vision and pattern
  recognition}, pp.\  3498--3505. IEEE, 2012.

\bibitem[Sutskever et~al.(2013)Sutskever, Martens, Dahl, and
  Hinton]{sutskever13}
Sutskever, I., Martens, J., Dahl, G., and Hinton, G.
\newblock On the importance of initialization and momentum in deep learning.
\newblock In \emph{International conference on machine learning}, pp.\
  1139--1147. PMLR, 2013.

\bibitem[Wah et~al.(2011)Wah, Branson, Welinder, Perona, and
  Belongie]{WahCUB_200_2011}
Wah, C., Branson, S., Welinder, P., Perona, P., and Belongie, S.
\newblock The caltech-ucsd birds-200-2011 dataset.
\newblock 2011.

\bibitem[Xiao et~al.(2017)Xiao, Rasul, and Vollgraf]{xiao2017fashion}
Xiao, H., Rasul, K., and Vollgraf, R.
\newblock Fashion-mnist: a novel image dataset for benchmarking machine
  learning algorithms.
\newblock \emph{arXiv preprint arXiv:1708.07747}, 2017.

\end{thebibliography}
\bibliographystyle{icml2021}

\end{document}


\onecolumn
\icmltitle{Supplementary File for \\ Large-Scale Meta-Learning with Continual Trajectory Shifting}



\icmlsetsymbol{equal}{*}

\begin{icmlauthorlist}
\icmlauthor{JaeWoong Shin}{equal,kaist}
\icmlauthor{Hae Beom Lee}{equal,kaist}
\icmlauthor{Boqing Gong}{google,kaist}
\icmlauthor{Sung Ju Hwang}{kaist,aitrics}
\end{icmlauthorlist}

\icmlaffiliation{kaist}{Graduate School of AI, KAIST, South Korea}
\icmlaffiliation{google}{Google, LA}
\icmlaffiliation{aitrics}{AITRICS, South Korea}

\icmlcorrespondingauthor{Sung Ju Hwang}{sjhwang82@kaist.ac.kr}

\icmlkeywords{Machine Learning, ICML}

\vskip 0.3in



\printAffiliationsAndNotice{\icmlEqualContribution} 





\newtheorem{thm}{Theorem}[section]
\newtheorem{lem}{Lemma}[section]
\newtheorem{cor}{Corollary}

\newcommand{\dee}{\mathrm{d}}
\newcommand{\grad}{\nabla}

\newcommand{\normal}{\mathcal{N}}
\newcommand{\bbE}{\mathbb{E}}
\newcommand{\calL}{\mathcal{L}}
\newcommand{\calM}{\mathcal{M}}
\newcommand{\calD}{\mathcal{D}}
\newcommand{\card}[1]{\vert {#1} \vert}
\newcommand{\distiid}{\overset{iid}{\dist}}
\newcommand{\distind}{\overset{ind}{\dist}}
\newcommand{\Indicator}[1]{\mathds{1}_{\{#1\}}}
\newcommand{\II}{\mathbb{I}}
\newcommand{\defas}{\vcentcolon=}  
\newcommand{\iid}{i.i.d.}

\newcommand{\bs}[1]{{\boldsymbol{#1}}}
\newcommand{\bepsilon}{{\boldsymbol{\epsilon}}}
\newcommand{\btheta}{{\boldsymbol{\theta}}}
\newcommand{\bTheta}{{\boldsymbol{\Theta}}}
\newcommand{\bbeta}{{\boldsymbol{\beta}}}
\newcommand{\Real}{{\mathbb{R}}}
\newcommand{\calP}{{\mathcal{P}}}
\newtheorem{observation}{Observation}

\newcommand{\D}{\mathcal{D}}
\newcommand{\tr}{\text{tr}}
\newcommand{\va}{\text{va}}
\newcommand{\te}{\text{te}}
\newcommand{\bx}{\mathbf{x}}
\newcommand{\bz}{\mathbf{z}}
\newcommand{\by}{\mathbf{y}}
\newcommand{\ba}{\mathbf{a}}
\newcommand{\bX}{\mathbf{X}}
\newcommand{\bY}{\mathbf{Y}}

\newcommand{\bzero}{\mathbf{0}}
\newcommand{\bone}{\mathbf{1}}
\newcommand{\bb}{\mathbf{b}}
\newcommand{\bH}{\mathbf{H}}
\newcommand{\bof}{\mathbf{f}}
\newcommand{\bhy}{\widehat{\by}}
\newcommand{\bw}{\mathbf{w}}
\newcommand{\bI}{\mathbf{I}}
\newcommand{\be}{\mathbf{e}}
\newcommand{\bG}{\mathbf{G}}
\newcommand{\bh}{\mathbf{h}}
\newcommand{\bhW}{{\widehat{\bW}}}
\newcommand{\btW}{{\widetilde{\bW}}}
\newcommand{\bV}{\mathbf{V}}
\newcommand{\bv}{\mathbf{v}}
\newcommand{\bphi}{{\boldsymbol{\phi}}}
\newcommand{\bpi}{{\boldsymbol{\pi}}}
\newcommand{\noise}{{\boldsymbol{\varepsilon}}}

\newcommand{\ts}{{(t)}}

\newcommand{\bee}{\begin{eqnarray}}
\newcommand{\eee}{\end{eqnarray}}

\appendix

This supplementary file consists of the following contents:
\vspace{-0.1in}
\begin{itemize}
    \item \textbf{Section~\ref{sec:inner_optimizer}:} We show that the type of inner-optimizer (e.g. SGD with momentum or Adam) can largely affect the quality of the initialization parameters at convergence.
    \item \textbf{Section~\ref{sec:trajectory_shifting}:} We visualize the effect of trajectory shifting with the synthetic experiments.
    \item \textbf{Section~\ref{sec:experimental_setup}:} We provide detailed description of the experimental setup for each experiment in the main paper, including the synthetic experiment, the image classification, the ImageNet experiment, and the empirical error analysis.
    \item \textbf{Section~\ref{sec:error_complexity}:} We derive Eq. (5) in the main paper, which is the complexity of the approximation error caused by the proposed continual trajectory shifting.
    \item \textbf{Section~\ref{sec:momentum}:} We prove that we can use the same shifting rule even with the momentum optimizer and weight decaying.
\end{itemize}

\section{Effect of Inner-optimizer Type}
\label{sec:inner_optimizer}

We briefly discuss if we can add in weight decay or change the type of optimizers in defining $U_k(\phi)$, without changing the results of Eq. (5) in the main paper. We also discuss which optimizer works relatively better over the others. The type of inner-optimizer is highly relevant to the quality of $\phi$ at convergence. Specifically, inner-optimizers with faster convergence result in faster meta-convergence as well, showing the strong dependency between the inner- and meta- optimization. Also, inner-optimizers that exhibit oscillating behavior helps the meta learner to escape from bad local minima. Momentum optimizer~\cite{sutskever13} is the one with all those properties. 

\begin{figure}[t]
    \vspace{-0.05in}
	\centering
	\hfill
	\subfigure[Starting point: (-5, 5)]{
	    \includegraphics[height=3.95cm]{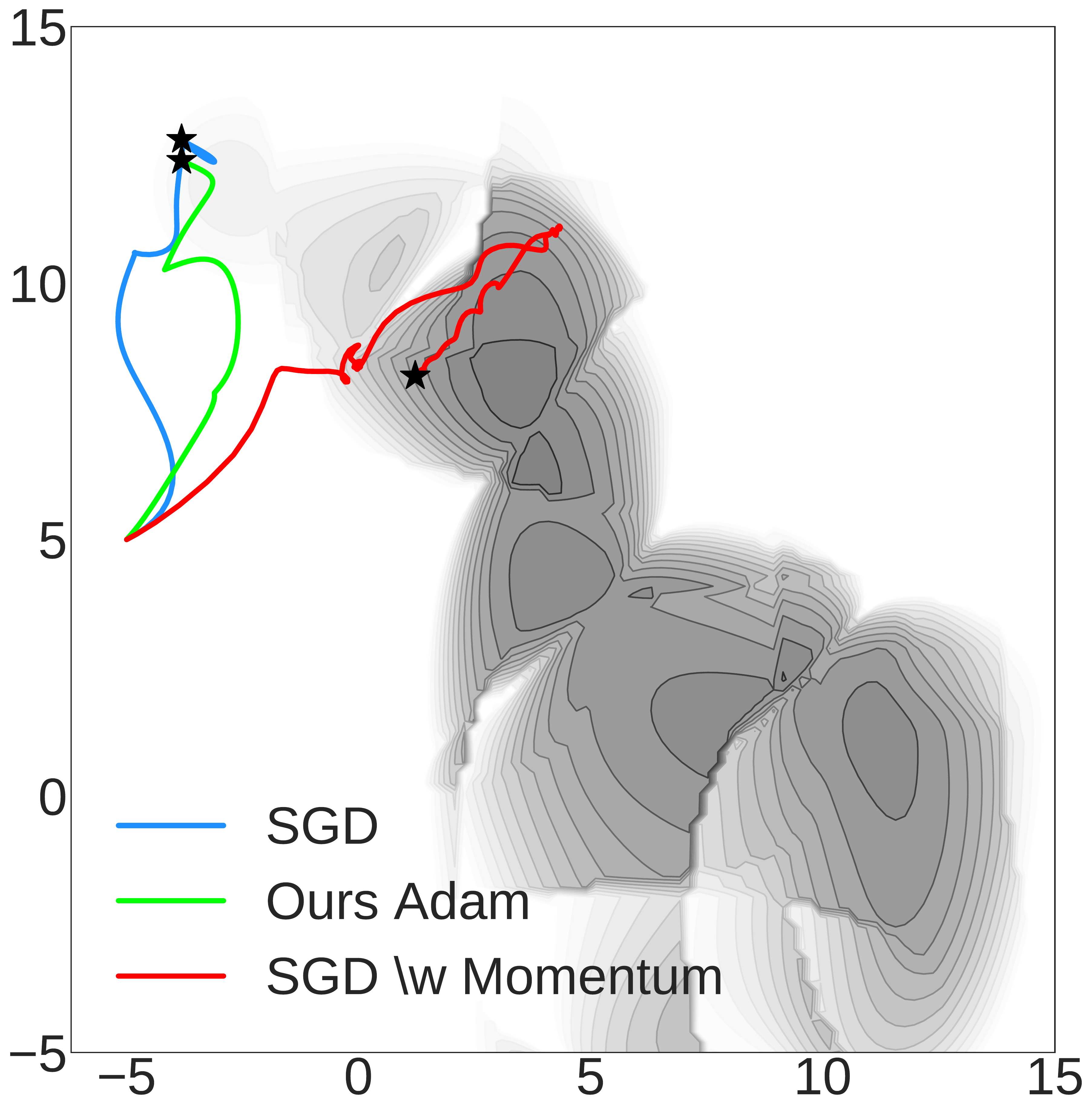}
	}
	\hfill
	\subfigure[Starting point: (15, 5)]{
	    \includegraphics[height=3.95cm]{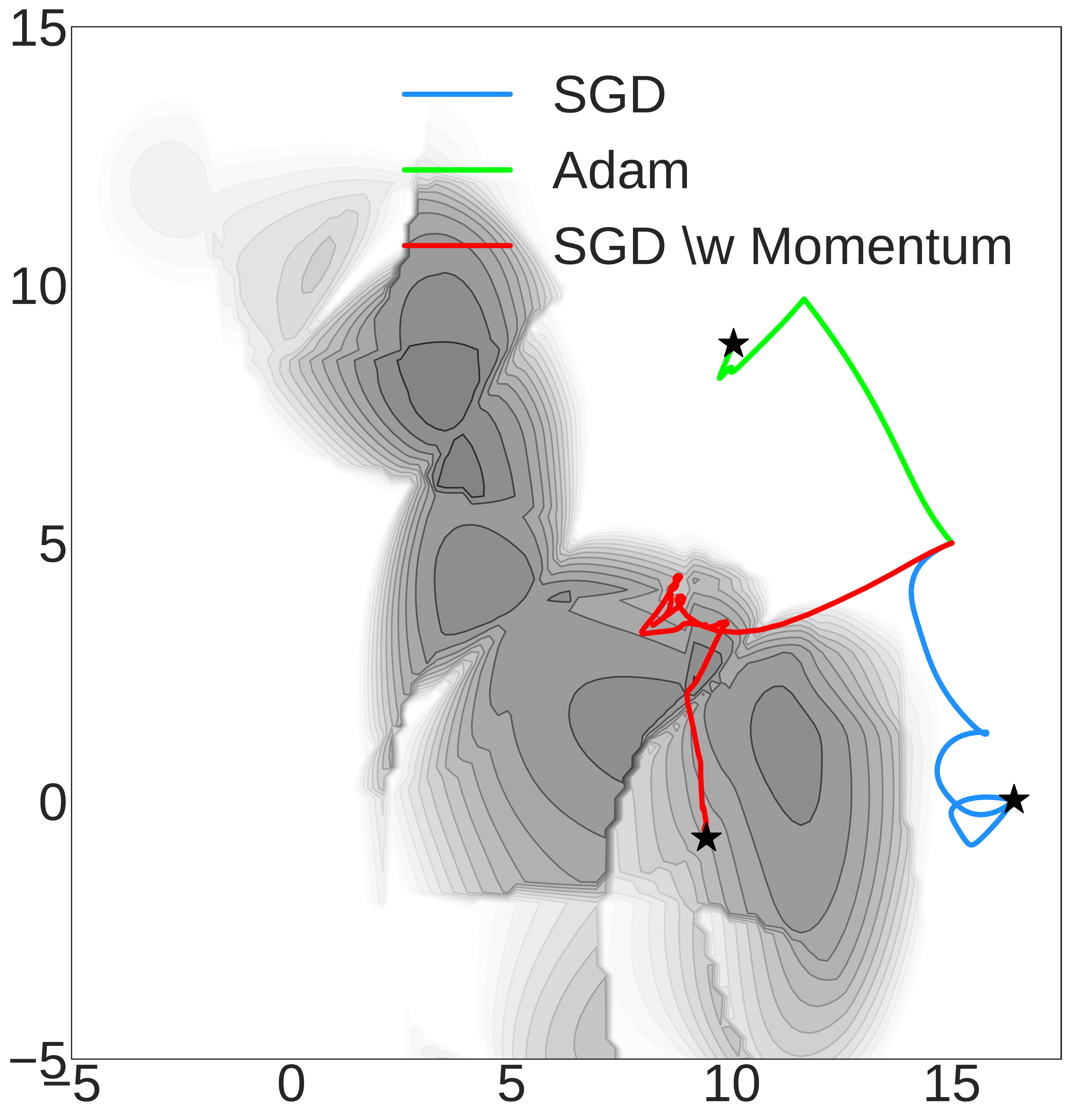}
    }
	\hfill
	\subfigure[Starting point: (5, -5)]{
	    \includegraphics[height=3.95cm]{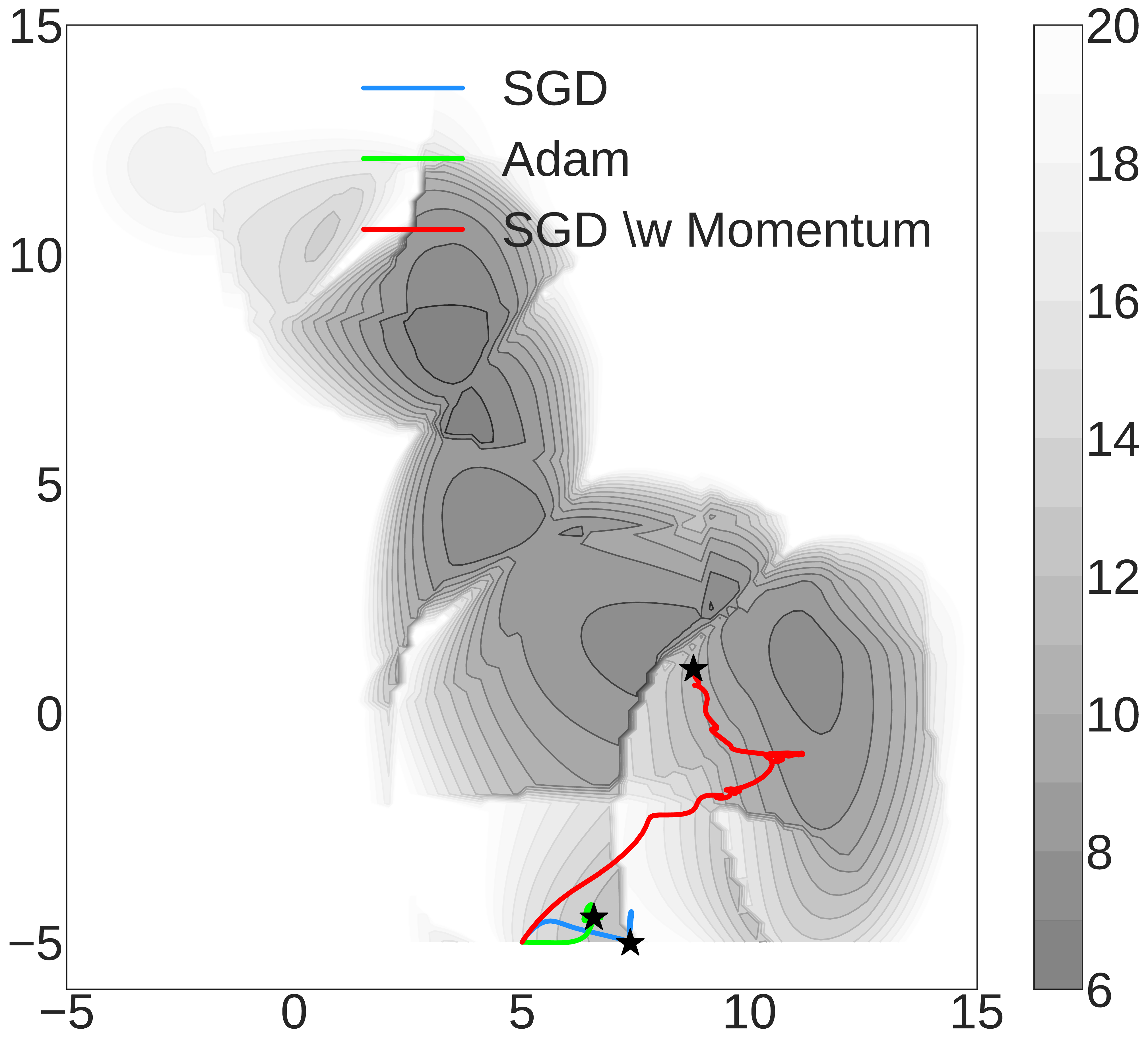}
    }
	\hfill
	\vspace{-0.1in}
	\caption{\small \textbf{Lines:} Meta-training trajectory of our method with various types of inner-optimizers. \textbf{Background contour:} Task-average loss after 100 gradient steps. The darker the background contour, the better quality of the initialization point. }
	\vspace{-0.1in}
    \label{fig:inner_opt}
\end{figure}

\paragraph{Momentum and weight decay} 
Given the momentum $\mu \in [0,1]$ and weight decay $\lambda \geq 0$, we can show that we can apply the same shifting rule $\theta_k \leftarrow \theta_k + \Delta_k$ introduced in the main paper. This will only result in higher approximation error compared to the vanilla SGD case. See Section~\ref{sec:momentum} for the derivation of the following results:
\begin{align}
    U_k\left(\phi + \sum_{i=1}^{k-1}\Delta_i\right) &=U_1(\cdots U_1(U_1(\phi)+\Delta_1) \cdots + \Delta_{k-1}) + O\left(\beta \alpha (h+2\lambda) k^2 + \beta^2 k\right)
\end{align}  for $k\ge2$.
\paragraph{Adam} Unfortunately, the analogous derivation for Adam optimizer~\cite{kingma2014adam} requires to differentiate very complicated expression involving element-wise square root division. Therefore, although we may use the same shifting rule $\theta_k \leftarrow \theta_k + \Delta_k$ together with the Adam optimizer, we cannot expect that the approximation error will be bounded in any reasonable way. However, we do not have to consider Adam as an inner-optimizer in context of meta-training because oscillating property of momentum optimizer is preferable over stable learning trajectory provided by Adam optimizer. In our synthetic experiment, we tried applying Adam, but obtained much worse initial model parameters than using momentum. See Figure~\ref{fig:inner_opt} for the actual meta-learning trajectories obtained with the various optimizers. 

\section{Visualization of Trajectory Shifting}
\label{sec:trajectory_shifting}
\begin{figure}[H]
    \vspace{-0.1in}
	\centering
	\subfigure[Task 1]{
	    \includegraphics[height=3.95cm]{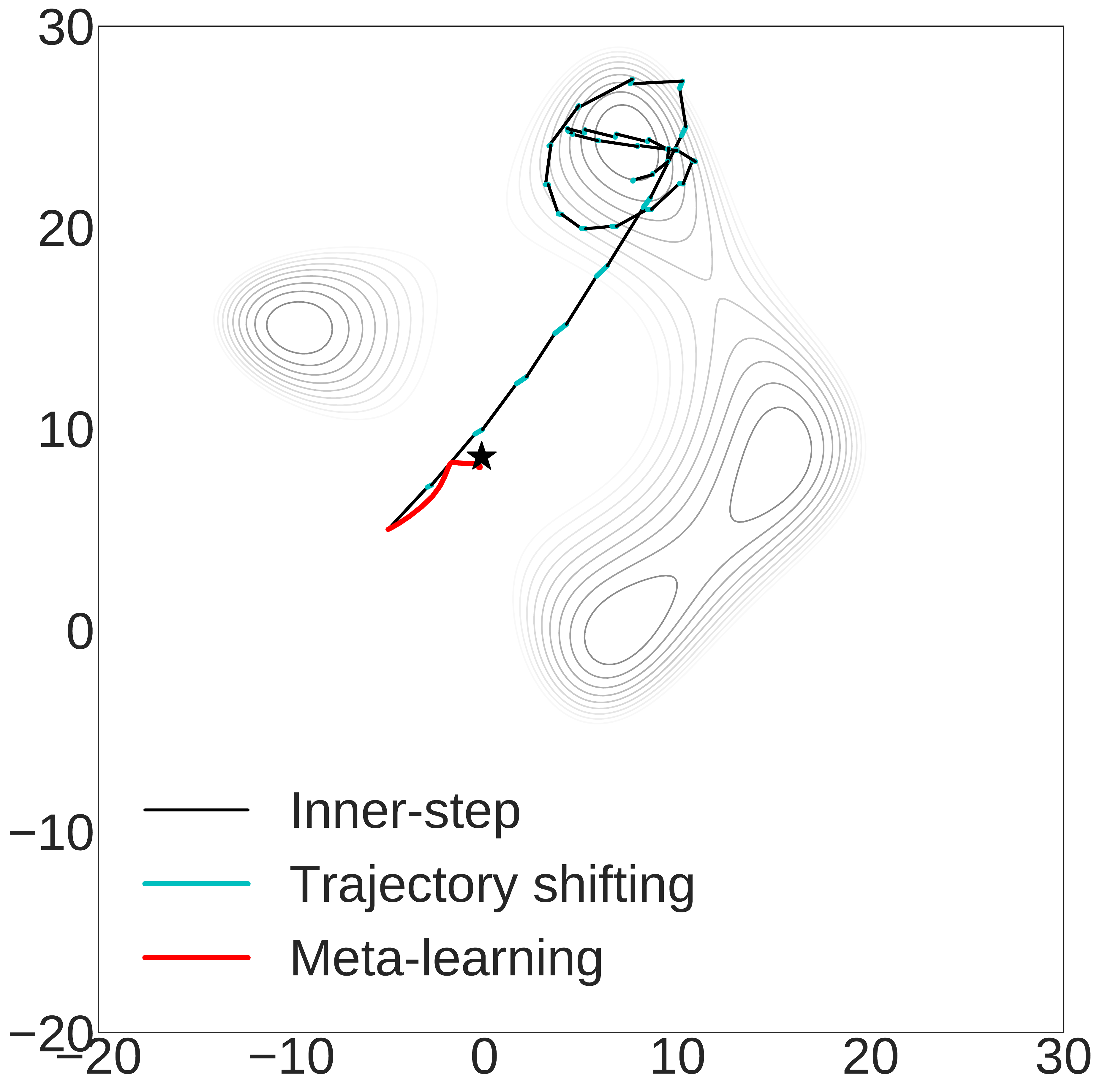}
	}
	\subfigure[Task 2]{
	    \includegraphics[height=3.95cm]{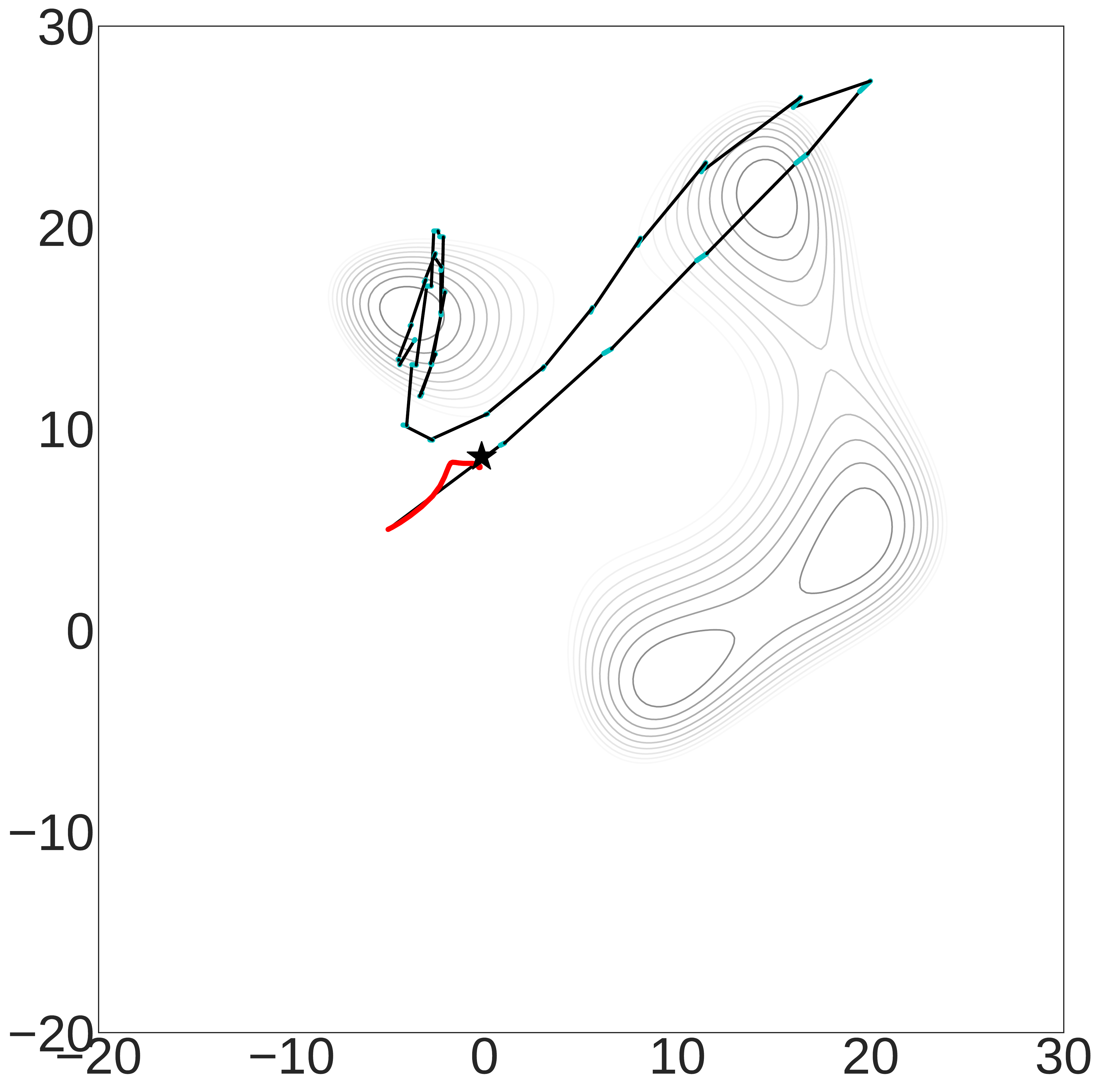}
	}
	\subfigure[Task 3]{
	    \includegraphics[height=3.95cm]{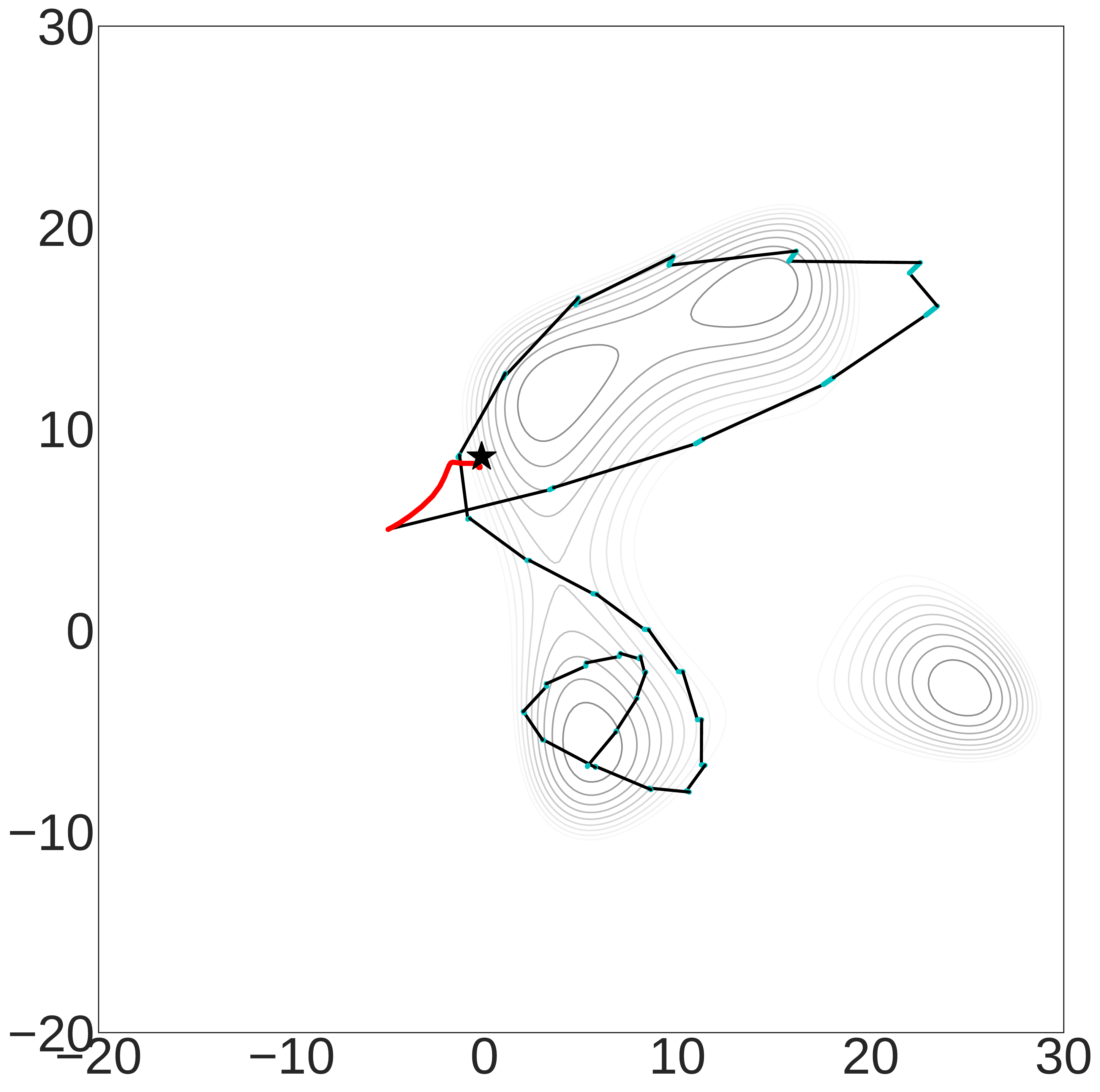}
	}
	\subfigure[Task 4]{
	    \includegraphics[height=3.95cm]{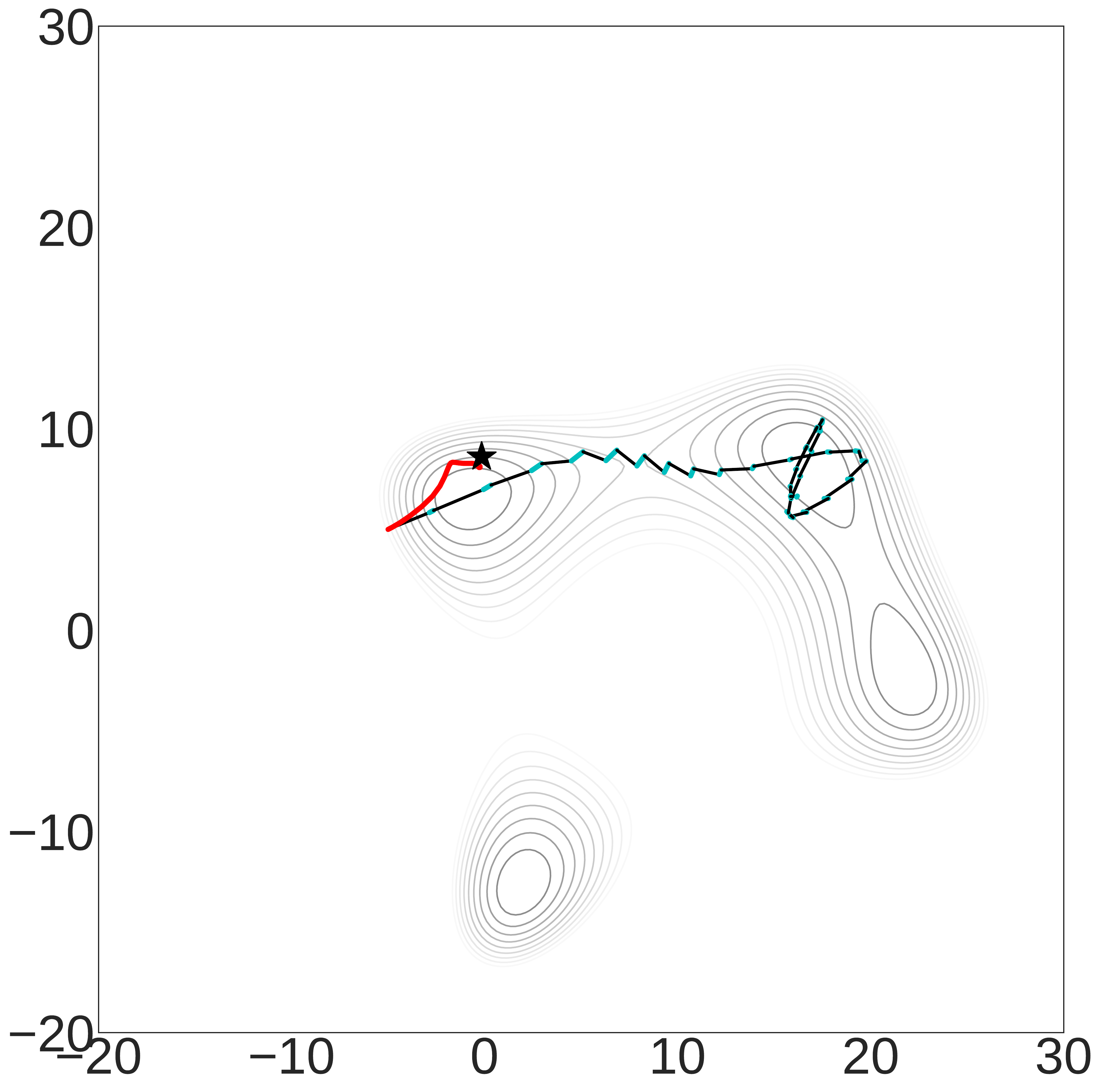}
	}
	\subfigure[Task 1]{
	    \includegraphics[height=3.95cm]{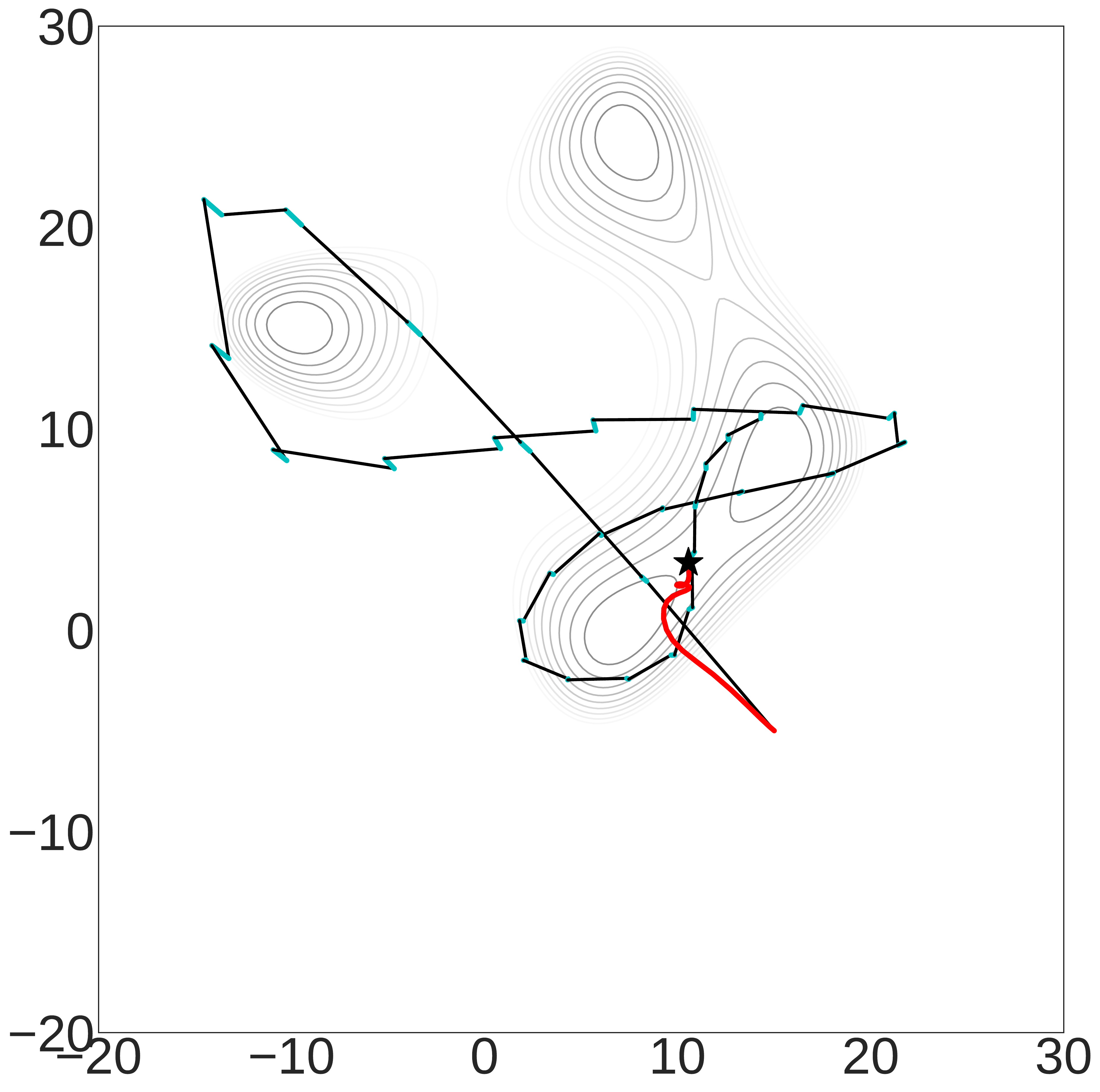}
	}
	\subfigure[Task 2]{
	    \includegraphics[height=3.95cm]{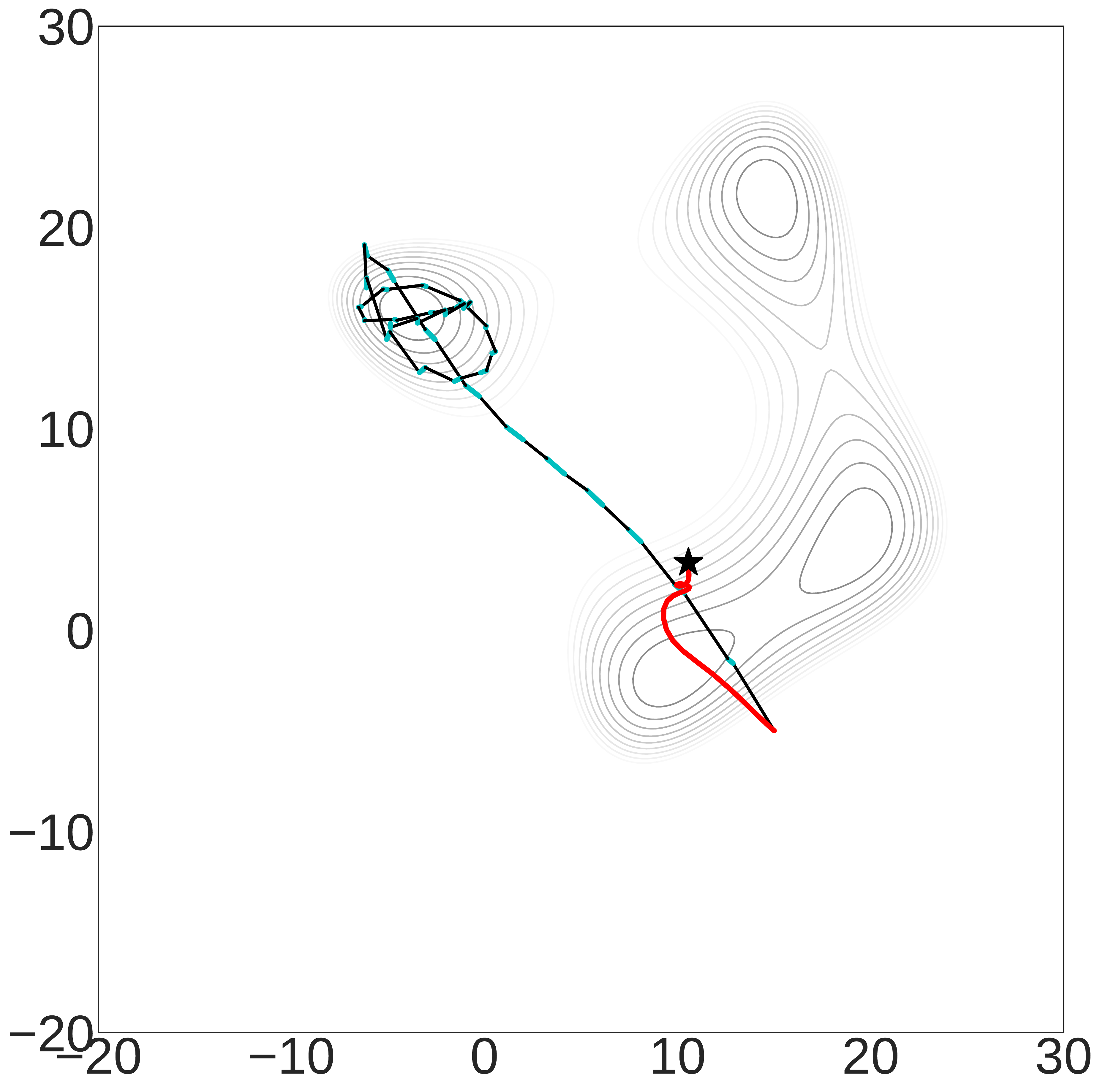}
	}
	\subfigure[Task 3]{
	    \includegraphics[height=3.95cm]{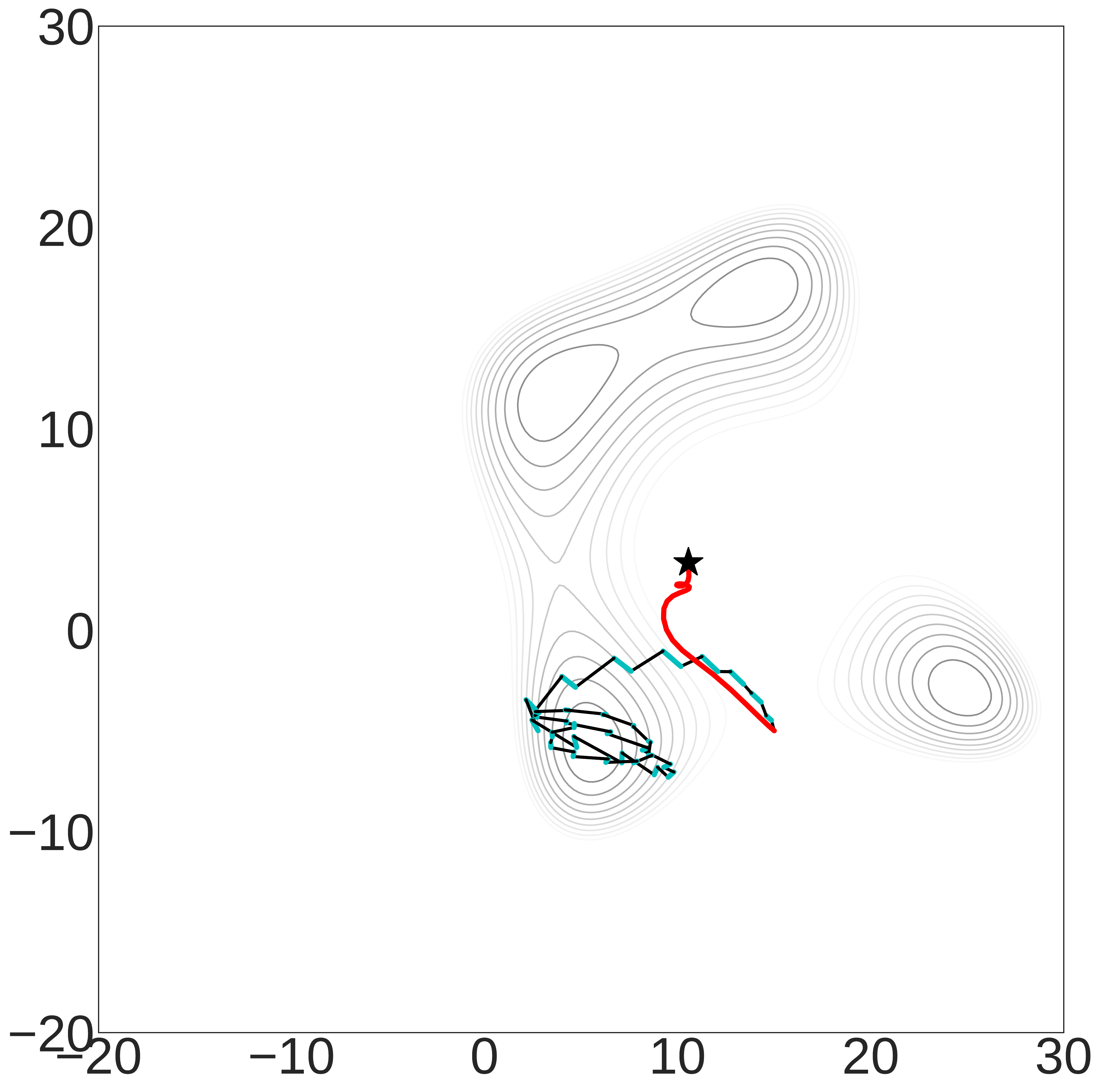}
	}
	\subfigure[Task 4]{
	    \includegraphics[height=3.95cm]{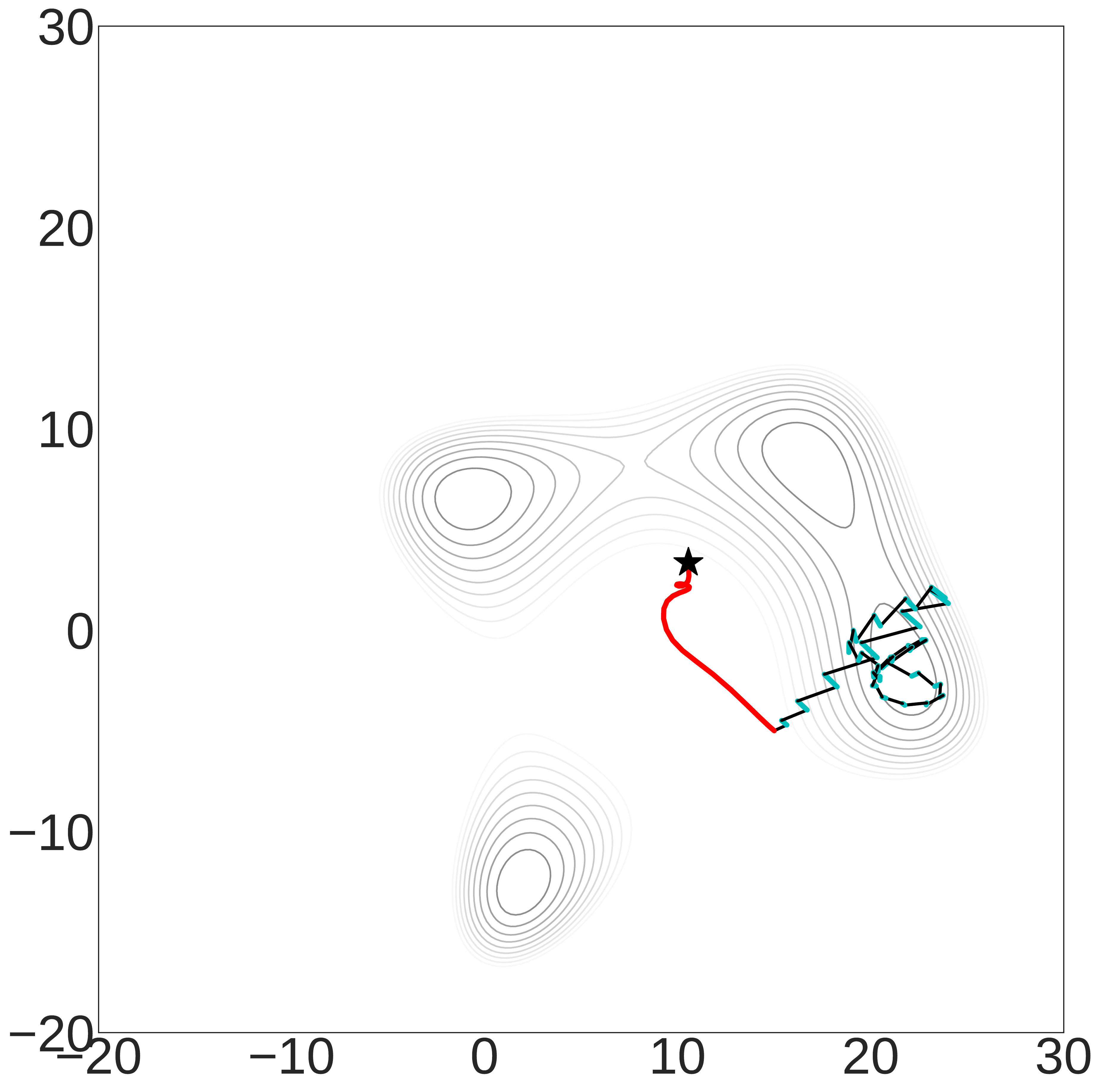}
	}
	\vspace{-0.15in}
	\caption{\small Visualization of the trajectory shifting with the four tasks (Task 1 - Task 4) from the synthetic experiments. \textbf{Top row:} starting point: (-5, 5). \textbf{Bottom row:} starting point: (15, 5). }
	\vspace{-0.1in}
    \label{fig:trajectory_shifting}
\end{figure}

In Figure~\ref{fig:trajectory_shifting}, we visualize the actual trajectory shifting with the synthetic experiments. We can see how each of the inner-learning trajectories is interleaved with a sequence of meta-updates.

\section{Experimental Setup}
\label{sec:experimental_setup}
In this section, we provide the detailed experimental setup for the synthetic experiments, the image classifications, the ImageNet experiments, and the empirical error analysis.
\subsection{Synthetic experiments}
\begin{figure}[H]
    \vspace{-0.20in}
	\centering
	\subfigure[Task 1]{
	    \includegraphics[height=3.95cm]{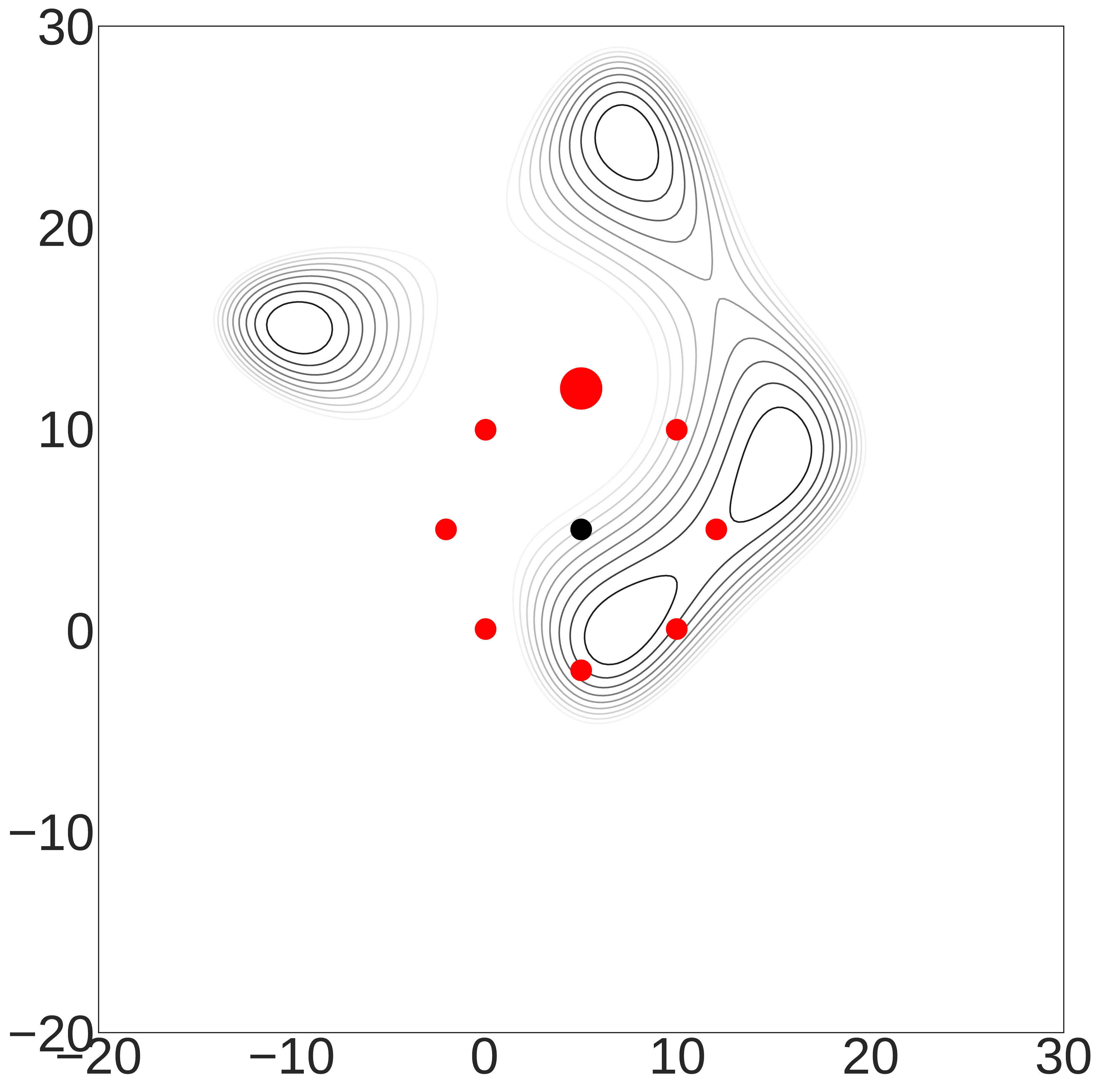}
	}
	\subfigure[Task 2]{
	    \includegraphics[height=3.95cm]{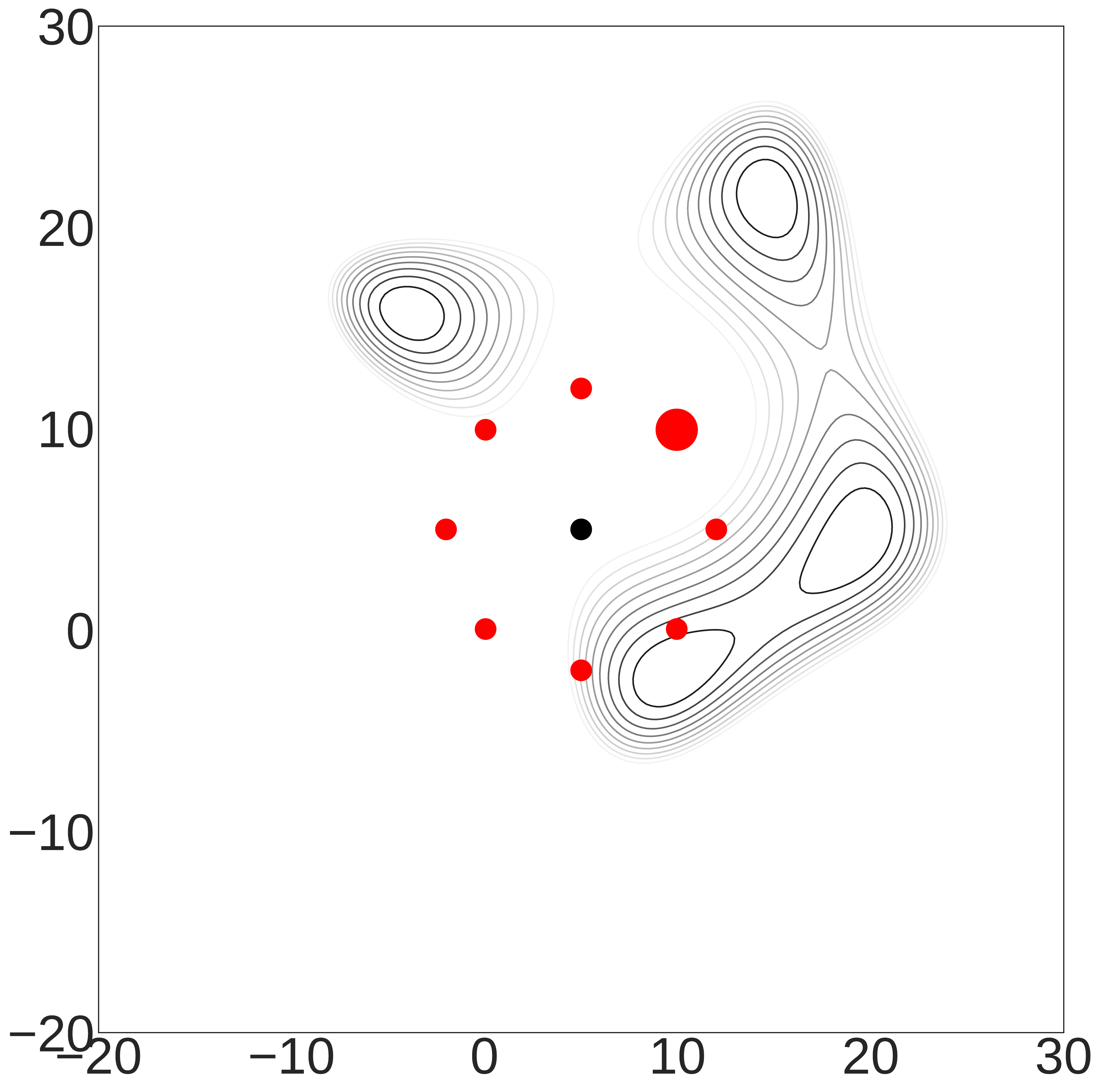}
	}
	\subfigure[Task 3]{
	    \includegraphics[height=3.95cm]{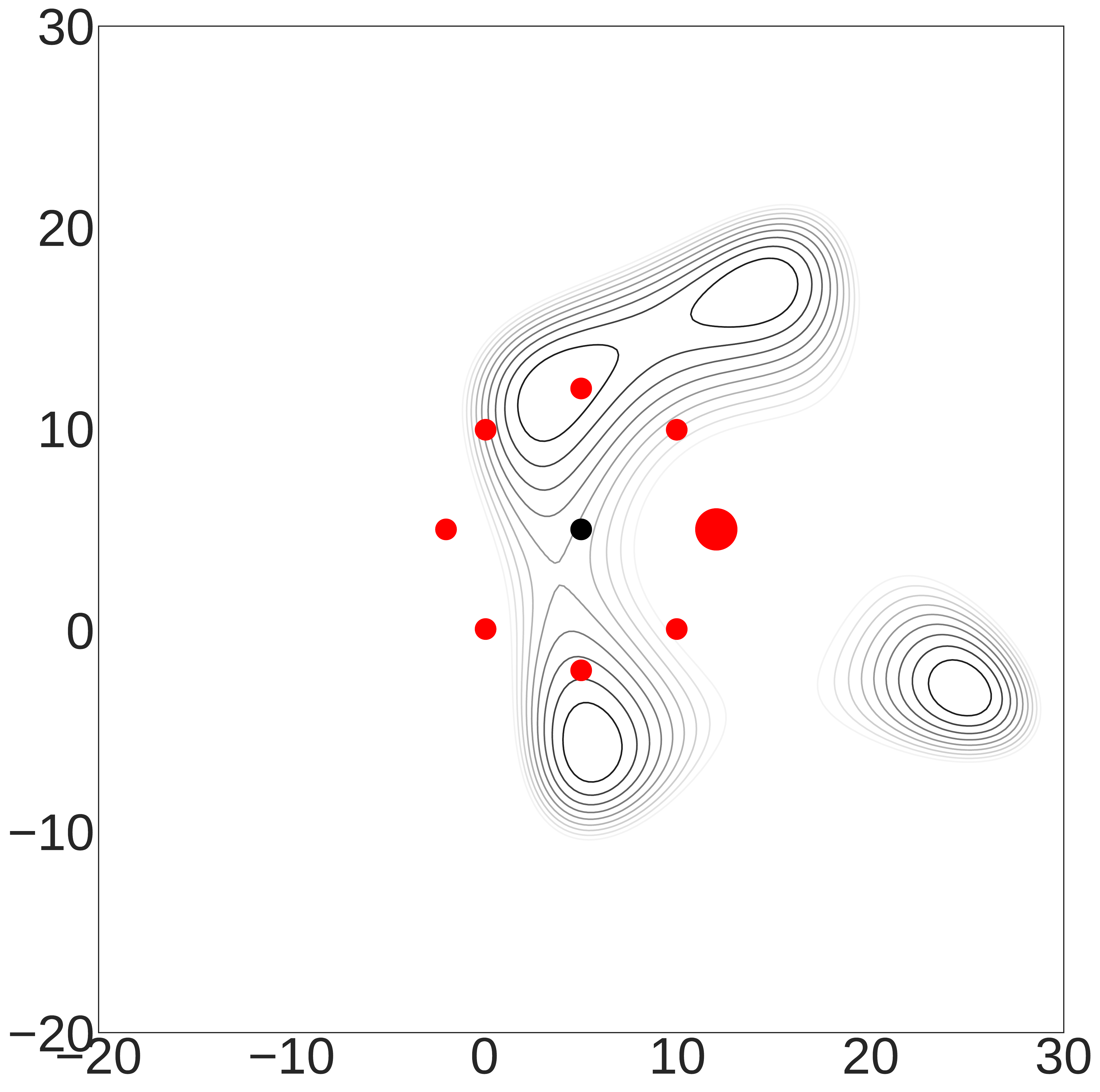}
	}
	\subfigure[Task 4]{
	    \includegraphics[height=3.95cm]{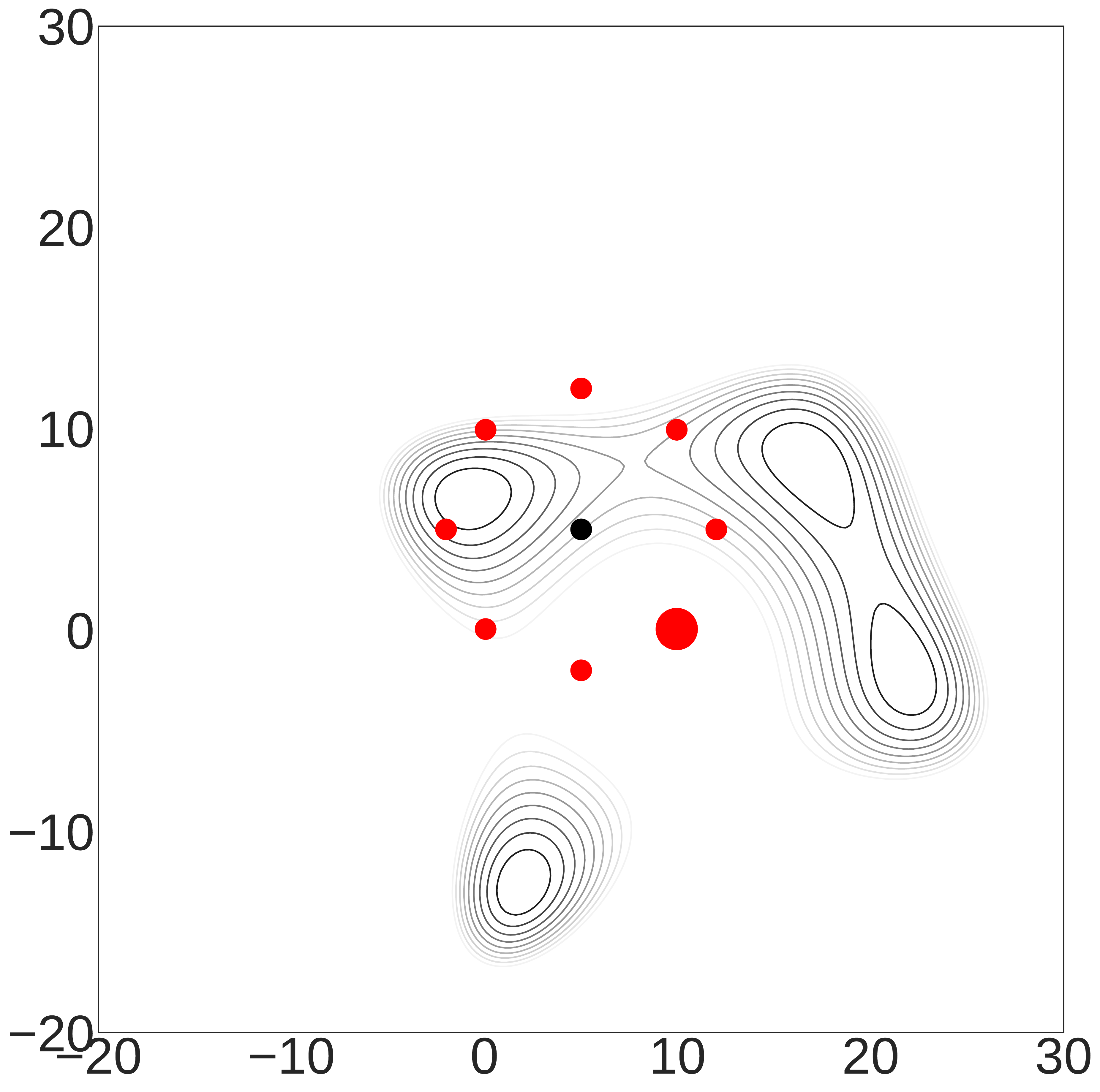}
	}
	\subfigure[Task 5]{
	    \includegraphics[height=3.95cm]{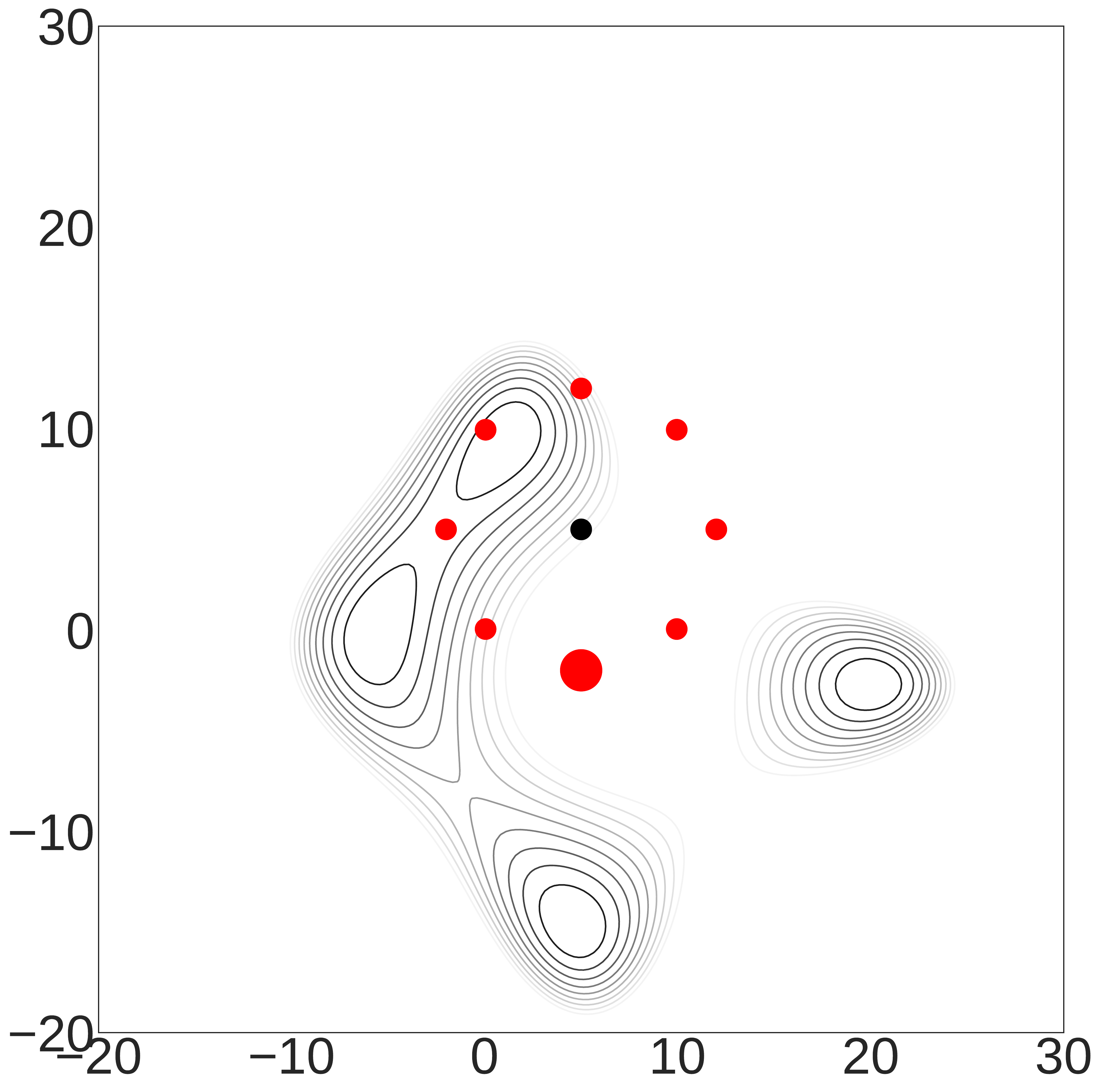}
	}
	\subfigure[Task 6]{
	    \includegraphics[height=3.95cm]{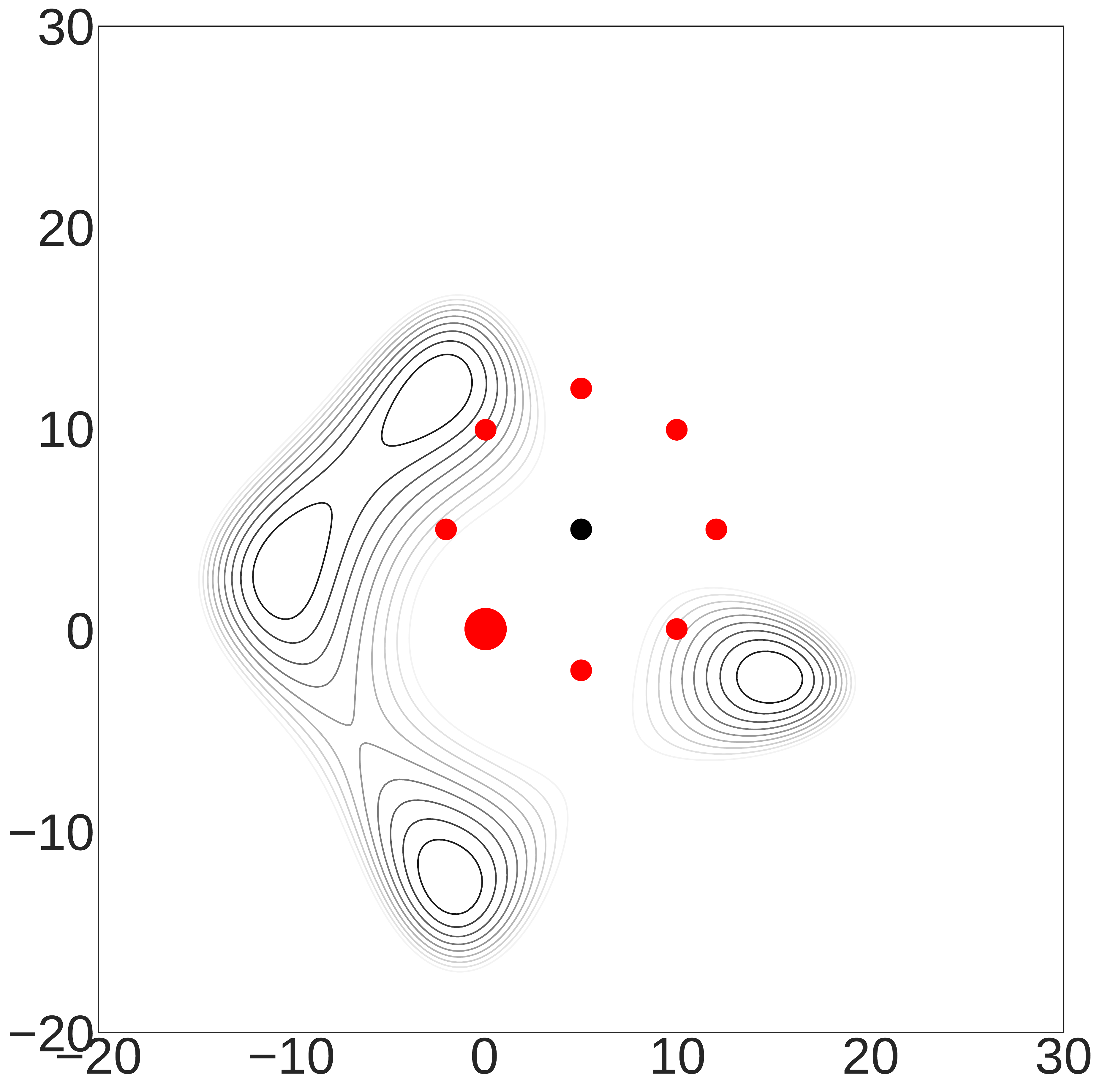}
	}
	\subfigure[Task 7]{
	    \includegraphics[height=3.95cm]{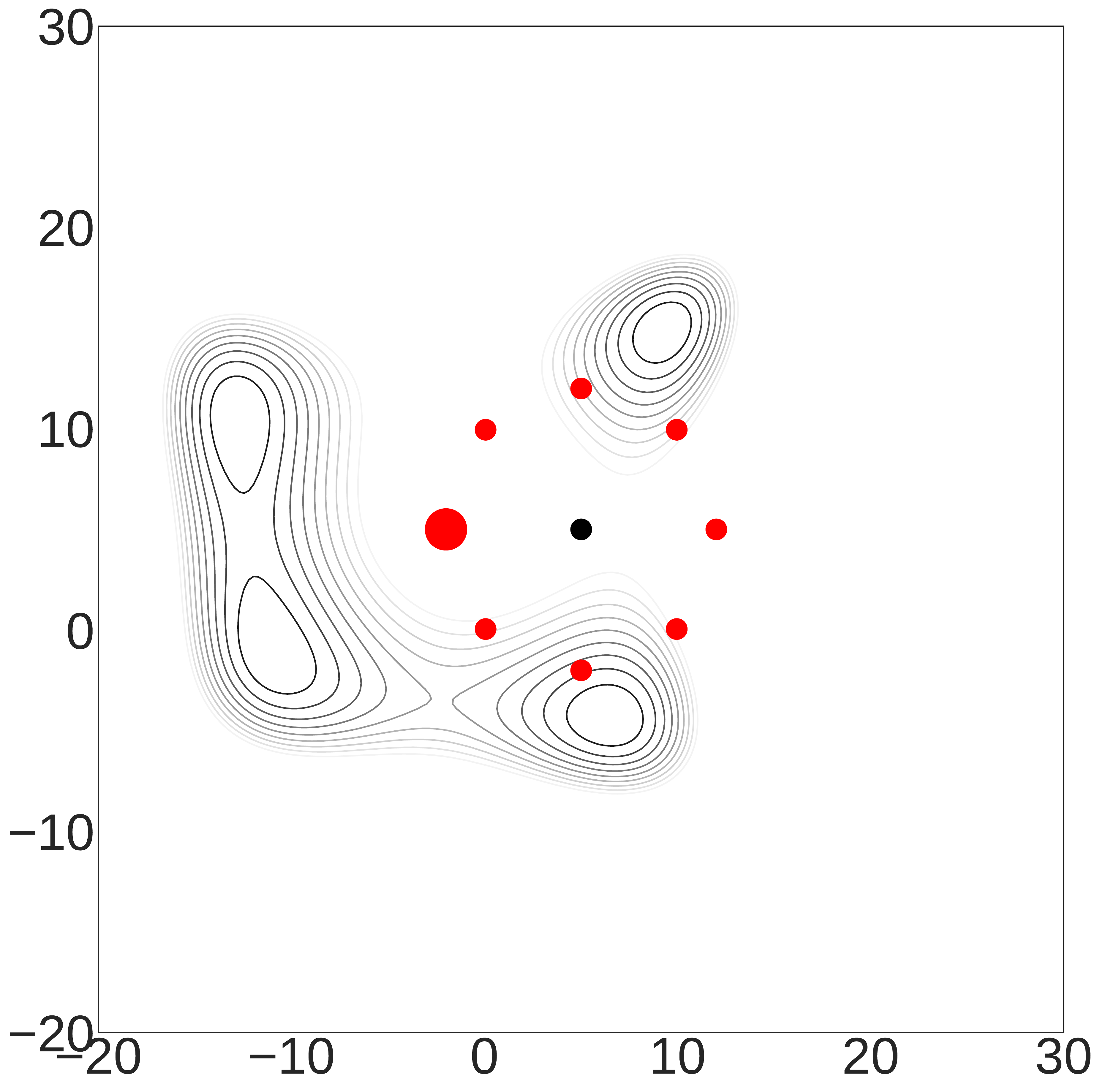}
	}
	\subfigure[Task 8]{
	    \includegraphics[height=3.95cm]{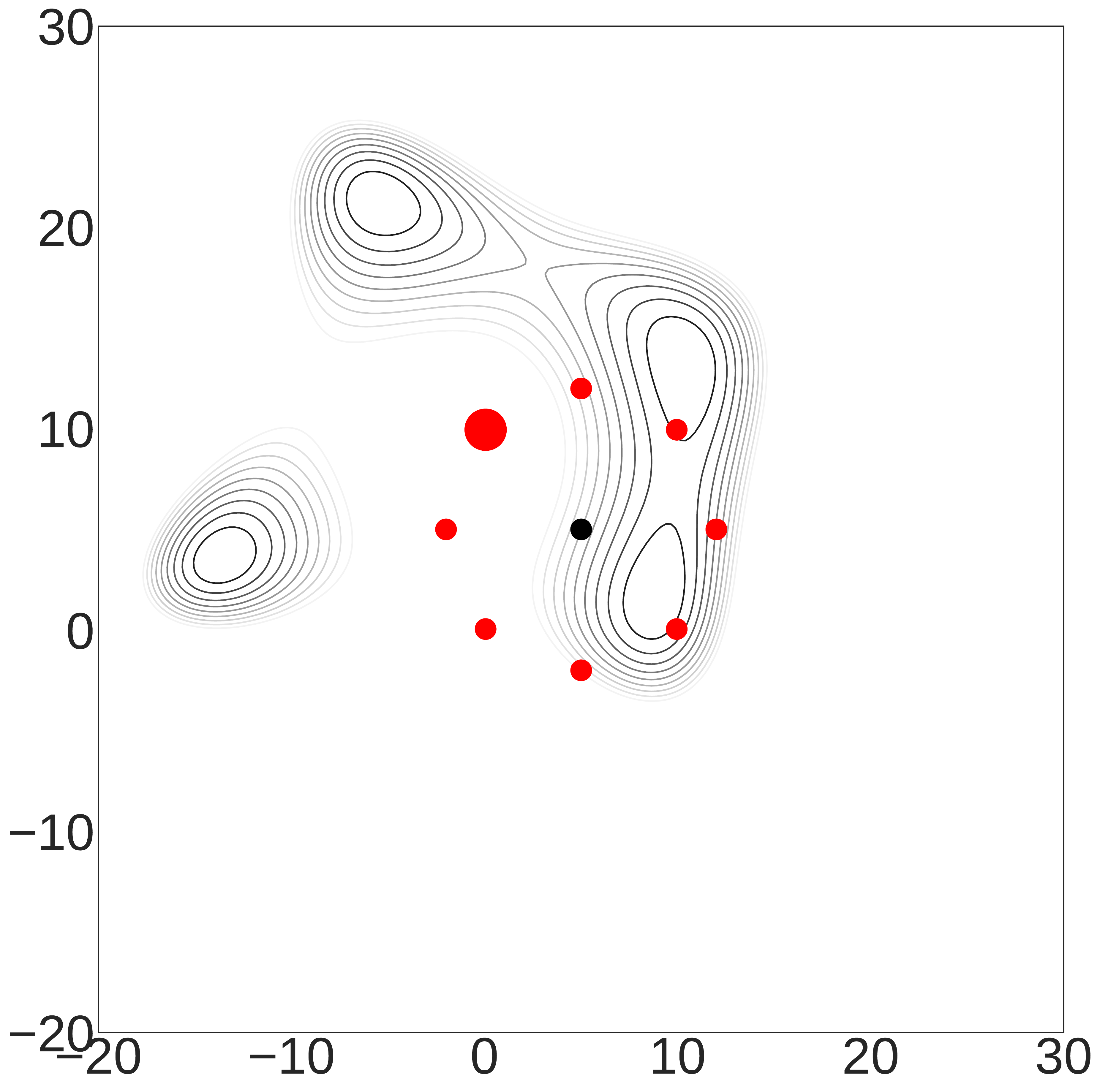}
	}
	\vspace{-0.15in}
	\caption{\small Loss surfaces of the eight tasks. }
	\vspace{-0.20in}
    \label{fig:synthetic_tasks}
\end{figure}

We visualize in Figure~\ref{fig:synthetic_tasks} the loss surfaces of all the eight tasks used for the synthetic experiments.

\subsection{Image classifications}
We provide some additional information about the experimental setup for the image classification experiments. 
\begin{table}[H]
\vspace{-0.1in}
\centering
\small
  \caption{\small \textbf{Meta-training datasets} for the image classification experiments.}
  \vspace{0.05in}
 \begin{tabular}{c | c c c c | c}
 \hline
Dataset & \# training instances & \# test instances & \# classes & Image size & Note\\
 \hline
 \hline
 Tiny ImageNet split 1~\cite{tinyimagenet}
 & 50,000
 & 5,000
 & 100
 & 64
 & Class 1-100
\\
Tiny ImageNet split 2~\cite{tinyimagenet}
 & 50,000
 & 5,000
 & 100
 & 64
 & Class 101-200
\\
CIFAR100~\cite{krizhevsky2009learning}
 & 50,000
 & 10,000
 & 100
 & 32
 & 
\\
Stanford Dogs~\cite{stanford_dogs}
 & 11,999
 & 8,580
 & 120
 & 84
 & 
\\
Aircraft~\cite{maji2013fine}
 & 6,667
 & 3,333
 & 100
 & 84
 & 
\\
CUB~\cite{WahCUB_200_2011}
 & 5,994
 & 5,794
 & 200
 & 84
 & 
\\
Fashion-MNIST~\cite{xiao2017fashion}
 & 60,000
 & 10,000
 & 10
 & 28
 & Grey-scale
\\
SVHN~\cite{netzer2011reading}
 & 73,257
 & 26,032
 & 10
 & 32
 & 
\\

\hline
\end{tabular}
\vspace{-0.25in}
  \label{tbl:meta_tr_datasets}
\end{table}
\begin{table}[H]
\centering
\small
  \caption{\small\textbf{Target datasets} for the image classification experiments.}
  \vspace{0.05in}
 \begin{tabular}{c | c c c c | c}
 \hline
Dataset & \# training instances & \# test instances & \# classes & Image size & Note\\
 \hline
 \hline
 Stanford Cars~\cite{stanford_cars}
 & 8,144
 & 8,041
 & 196
 & 84
 & 
\\
QuickDraw~\cite{DBLP:journals/corr/HaE17}
 & 34,500
 & 34,500
 & 345
 & 28
 & Grey-scale
\\
VGG Flowers~\cite{Nilsback08}
 & 2,040
 & 6,149
 & 102
 & 84
 & 
\\
VGG Pets~\cite{parkhi12a}
 & 3,680
 & 3,669
 & 37
 & 84
 & 
\\
STL10~\cite{coates2011analysis}
 & 5,000
 & 8,000
 & 10
 & 32
 & 
\\
\hline
\end{tabular}
\vspace{-0.25in}
  \label{tbl:meta_te_datasets}
\end{table}
\begin{table}[H]
\centering
\small
  \caption{\small The value of $\beta$ used for the image classification experiments.}
  \vspace{0.05in}
 \begin{tabular}{c | c c c }
 \hline
& \multicolumn{3}{c}{$K$} \\\
Method & 10 & 100 & 1000 \\
 \hline
 \hline
FOMAML~\cite{finn2017model}
 & 0.5
 & 0.2
 & 0.1
\\
Leap~\cite{leap}
 & 0.1
 & 0.1
 & 0.1
\\
Reptile~\cite{reptile}
 & 5
 & 2
 & 1
\\
Ours
 & 1
 & 0.1
 & 0.01
\\
\hline
\end{tabular}
\vspace{-0.15in}
  \label{tbl:beta}
\end{table}
\vspace{-0.1in}
\begin{itemize}
\item See Table~\ref{tbl:meta_tr_datasets} for more information about the datasets used for meta-training and Table~\ref{tbl:meta_te_datasets} for meta-testing. 


\item We carefully tuned the meta-learning rate $\beta$ for all the meta-learning baselines. Notably, we found that the optimal $\beta$ should increase as we reduce the length of inner-optimization trajectory $K$. See Table~\ref{tbl:beta} for the actual value of $\beta$ we used in the experiments.

\item The last fully connected layer (classifier) is a part of $\theta$, but not included in $\phi$. Batch norm parameters (i.e. scale and shift) are included in both $\theta$ and $\phi$. However, batch statistics (i.e. running mean and running variance) are neither a part of $\theta$ nor $\phi$.

\item Recall from the Algorithm 1 and 2 in the main paper that we repeat the inner-optimization process $M$ times, and we reset the task-specific parameters before starting each process. Note that we \textbf{do not} reset the following information: the statistics for the optimizer, the parameters for the last fully connected layer, and the batch norm statistics. Conceptually, it would be natural to reset the above information as well because we do not transfer them to meta-testing. However, we found that it makes no difference in terms of meta-testing performance.
\end{itemize}

\subsection{ImageNet experiments}
\begin{figure}[H]
    \vspace{-0.25in}
	\centering
	\includegraphics[width=1.\linewidth]{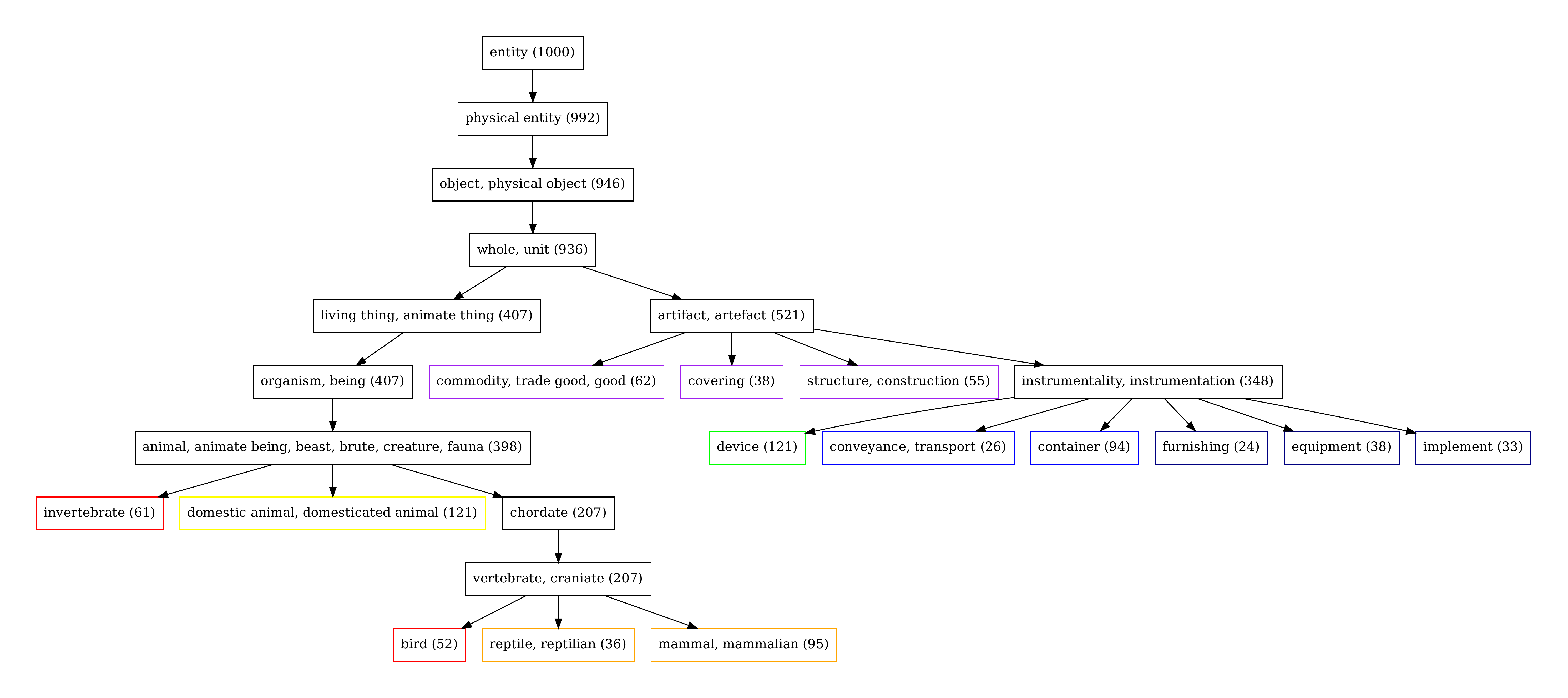}
	\vspace{-0.4in}
	\caption{\small \textbf{ImageNet splits based on the WordNet hierarchy}. Each node correspones to a WordNet label and the number of its subclasses is shown in the parenthesis. The nodes of the same split are shown in the same color. We do not show the nodes of the last split 8 for better visualization. }
	\vspace{-0.2in}
    \label{fig:hierarchy}
\end{figure}

We next provide the detailed description about the datasets used for the ImageNet experiments in Figure~\ref{fig:hierarchy}, Table~\ref{tbl:imagenet_split}, and Table~\ref{tbl:imagenet_dataset}.
\begin{table}[H]
\centering
\small
  \caption{\small ImageNet splits for \textbf{meta-training}}
  \vspace{0.05in}
 \begin{tabular}{c | c c | c  c }
 \hline
Split & \# training instances & \# test instances & \# classes & WordNet label (\# classes)\\
 \hline
 \hline
\multirow{2}{*}{1}  & \multirow{2}{*}{146,741} & \multirow{2}{*}{5,650} & \multirow{2}{*}{113}  & bird (52)         \\ 
                    &                   &                   &                       & invertebrate (61) \\
\hline
\multirow{2}{*}{2}  & \multirow{2}{*}{169,582} & \multirow{2}{*}{6,550} & \multirow{2}{*}{131}  & reptile (36)      \\ 
                    &                   &                   &                       & mammal (95)       \\
\hline
3                   & 151,773                  &  6,050                 & 121                   & domestic animal(121) \\
\hline
4                   & 154,047                  & 6,050                  & 121                   & device (121)  \\
\hline
\multirow{2}{*}{5}  & \multirow{2}{*}{154,373} & \multirow{2}{*}{6,000} & \multirow{2}{*}{120}  & conveyance, transport (26) \\
                    &                   &                   &                       & container (94) \\
\hline
\multirow{3}{*}{6}  & \multirow{3}{*}{121,344} & \multirow{3}{*}{4,750} & \multirow{3}{*}{95}  & furnishing (24) \\
                    &                   &                   &                       & equipment (38) \\
                    &                   &                   &                       & implement (33) \\
\hline
\multirow{3}{*}{7}  & \multirow{3}{*}{198,314} & \multirow{3}{*}{7,750} & \multirow{3}{*}{155} & commodity, trade good, good (62) \\
                    &                       &                       &                   & covering (38) \\
                    &                       &                       &                   & structure, construction (55) \\
\hline
8                   & 184,993                      &  7,200                     & 144                  & etc \\
\hline
\end{tabular}
  \label{tbl:imagenet_split}
\end{table}
\begin{table}[H]
\centering
\small
  \caption{\small \textbf{Target datasets} for the ImageNet experiments.}
  \vspace{0.05in}
   \resizebox{1.\linewidth}{!}{
 \begin{tabular}{c | c c c c | c}
 \hline
Dataset & \# training instances & \# test instances & \# classes & Image size & Note\\
 \hline
 \hline
 
CIFAR100~\cite{krizhevsky2009learning}
 & 50,000
 & 10,000
 & 100
 & 128
 & Upscaled from 32$\times$32
\\

CIFAR10~\cite{krizhevsky2009learning}
 & 50,000
 & 10,000
 & 10
 & 128
 & Upscaled from 32$\times$32
\\

SVHN~\cite{netzer2011reading}
 & 73,257
 & 26,032
 & 10
 & 128
 & Upscaled from 32$\times$32
\\
 
Stanford Dogs~\cite{stanford_dogs}
 & 11,999
 & 8,580
 & 120
 & 224
 &
\\


VGG Pets~\cite{parkhi12a}
 & 3,680
 & 3,669
 & 37
 & 224
 &
\\

VGG Flowers~\cite{Nilsback08}
 & 2,040
 & 6,149
 & 102
 & 224
 &
\\

Food-101~\cite{BossardGG14}
 & 75,750
 & 25,250
 & 101
 & 224
 &
\\


CUB~\cite{WahCUB_200_2011}
 & 5,994
 & 5,794
 & 200
 & 224
 &
\\


DTD~\cite{cimpoi14describing}
 & 4,230
 & 1,410
 & 47
 & 224
 & Texture dataset
\\
\hline
\end{tabular}
 }
  \label{tbl:imagenet_dataset}
\end{table}

\subsection{Empirical error analysis}
We explain the experimental setup for \textbf{the Figure 3 in the main paper}. We define the approximation error as:
\begin{equation}
\begin{aligned}
    \varepsilon &:= U_k(\phi + \Delta_1 + \cdots + \Delta_{k-1} ) - U_1(\cdots U_1(U_1(\phi)+\Delta_1) \cdots + \Delta_{k-1}) \label{eq:error}
\end{aligned}
\end{equation}
and report $\log_{10} \|\varepsilon\|_2$ versus inner-learning rate $\alpha$, meta-learning rate $\beta$, length of inner-learning trajectory $k$, and type of network activations (ReLU vs. Softplus). 
\vspace{-0.1in}
\begin{itemize}
    \item We use ResNet20.
    \item We use the first split of TinyImageNet dataset to generate a sequence of losses $\mathcal{L}_0,\dots,\mathcal{L}_{k-1}$. Batch size is set to 128.
    \item We use vanilla SGD for $U_k$, such that $U_k(\phi) := \phi - \alpha \sum_{i=0}^{k-1} \grad_{\theta} \mathcal{L}_{i}|_{\theta=\theta_{i}}$. 
    \item Note that $\Delta_i = \beta \cdot \mathsf{MetaGrad}_i$. In order to compute the approximation error $\varepsilon$ in a feasible amount of time, we assume that $\mathsf{MetaGrad}_0,\dots,\mathsf{MetaGrad}_{k-1}$ are sampled from the following synthetic distribution: $\mathsf{MetaGrad}_i = x_i/{\|x_i\|_2}$, where $x_i \sim N(0,I)$ for $i=0,\dots,k-1$. Therefore, we have $\|\Delta_0\|_2 = \cdots = \|\Delta_{k-1}\|_2 = \beta$, i.e. the size of meta-update is fixed as $\beta$, but the direction is randomized. We use the same sequence of $\Delta_0,\dots,\Delta_{k-1}$ in computing the first and second term of Eq.~\eqref{eq:error}.
\end{itemize}

\section{Derivation of the Error Complexity}
\label{sec:error_complexity}
In this section, we derive \textbf{Eq. (5) in the main paper}, the complexity of the approximation error. We first recap the notations:
\begin{itemize}
    \item $\phi$: Shared initial model parameters that we meta-learn
    \item $\alpha$: Inner-learning rate
    \item $\beta$: Meta-learning rate
    \item $U_k(\omega)$: Task-specific parameters after $k$ SGD steps from $\omega$, i.e. 
    \begin{align*}
    U_k(\omega) = {U_{k-1}(\omega) - \alpha\grad_{\omega}\mathcal{L}_{k-1}|_{\omega=U_{k-1}(\omega)}} = \omega - \alpha \sum_{i=0}^{k-1} \grad_{\omega} \mathcal{L}_{i}
    \end{align*}
    \item $\theta_k$: Task-specific parameters after $k$ SGD steps from $\phi$, i.e. $U_k(\phi)$
    \item $H_k$: Hessian of loss function at $\theta_k$
    \item $\Delta_k$: Meta-update (or trajectory shifting) at step $k$, i.e. \[\Delta_k = -\beta \cdot \mathsf{MetaGrad}(\phi;\theta_k)\]
\end{itemize}

Our derivation is based on the following assumptions:
\begin{enumerate}
    \item $U_k$ is infinitely differentiable.
    \item Norm of the Hessian is bounded by $h$ at everywhere, i.e. $\|H\|=O(h)$.
    \item Norm of meta-update is bounded by $\beta$ for every step, i.e. $\|\Delta_k\|=O(\beta)$ for every $k$.
\end{enumerate}

\begin{thm} For $k\ge 1$ and any $\Delta$ whose norm is sufficiently small, 
    \[U_k(\phi + \Delta) = U_k(\phi) + \Delta + O(\alpha h k\|\Delta\| + \|\Delta\|^2)\]
    \label{thm1}
\end{thm}
\begin{proof}
    Using the Talyor approximation,
    \begin{align*}
    U_k(\phi + \Delta) &= U_k(\phi) + \frac{\partial U_k(\phi)}{\partial \phi}\Delta + \frac{1}{2} \Delta^\top \frac{\partial^2 U_k(\phi)}{\partial \phi^2} \Delta + \cdots \nonumber \\ 
    &= U_k(\phi) + \frac{\partial U_k(\phi)}{\partial \phi}\Delta + O(\|\Delta\|^2)
    \end{align*}
    On the other hand,
    \begin{align*}
        \frac{\partial U_k(\phi)}{\partial\phi} &= \frac{\partial U_k(\phi)}{\partial U_{k-1}(\phi)} \cdots
        \frac{\partial U_1(\phi)}{\partial\phi} = \prod_{i=0}^{k-1} \left(I-\alpha H_{i}\right)  \\ 
        &= I - \sum_{i=0}^{k-1} \alpha H_i + \sum_{i=0}^{k-1}\sum_{j=i+1}^{k-1}\alpha^2 H_i H_j+\cdots = I + O(\alpha h k )
    \end{align*}
    
    Combining the two approximations,
    \begin{align*}
        U_k(\phi + \Delta) &=U_k(\phi) + ( I + O(\alpha h k ))\Delta + O(\|\Delta\|^2) \\
        &= U_k(\phi) + \Delta + O(\alpha h k \|\Delta\| + \|\Delta\|^2) 
    \end{align*}
\end{proof}

\begin{thm} For $k\ge 1$ and any $\{\Delta_i\}$ with $\|\Delta_i\|=O(\beta)$,
\begin{align*}
    U_k\left(\phi + \sum_{i=1}^{k}\Delta_i\right) =U_1(\cdots U_1(U_1(\phi)+\Delta_1) \cdots + \Delta_{k-1}) + \Delta_k + O(\beta \alpha h k^2 + \beta^2 k)
    \end{align*}
    \label{thm2}
\end{thm}

\begin{proof}
    For $k=1$, 
    \begin{align*}
        U_1(\phi + \Delta_1) = U_1(\phi)+\Delta_1 + O(\beta \alpha h  + \beta^2)&&\text{(Theorem \ref{thm1})}
    \end{align*} With the assumption at step $k$,
    \begin{align*}
        U_{k+1}\left(\phi + \sum_{i=1}^{k+1}\Delta_i\right) &=U_{k+1}\left(\phi+ \sum_{i=1}^{k}\Delta_i\right)+\Delta_{k+1} + O\left(\alpha h (k+1)\|\Delta_{k+1}\| + \|\Delta_{k+1}\|^2\right) \\
        &\omit\hfill\text{(Theorem \ref{thm1})}\\
        &= U_1\left(U_{k}\left(\phi+ \sum_{i=1}^{k}\Delta_i\right)\right) +\Delta_{k+1}+ O(\beta\alpha h (k+1) + \beta^2) \\
        &= U_1\left(U_1\left(\cdots U_1(U_1(\phi)+\Delta_1\right) \cdots + \Delta_{k-1}) + \Delta_{k} + O(\beta\alpha h k^2  + \beta^2k)\right) \\
        &\qquad+\Delta_{k+1}+ O(\beta\alpha h (k+1) + \beta^2) \\
        &\omit\hfill\text{(Assumption at step $k$)}\\
        &= U_1\left(U_1(\cdots U_1(U_1(\phi)+\Delta_1) \cdots + \Delta_{k-1}) + \Delta_k\right) +O(\beta\alpha h k^2 + \beta^2k) \\
        &\qquad+O(\alpha h (\beta\alpha h k^2 + \beta^2k)+(\beta\alpha h k^2 + \beta^2k)^2) \\
        &\qquad+\Delta_{k+1}+ O(\beta\alpha h (k+1) + \beta^2) \\
        &\omit\hfill \text{(Theorem \ref{thm1})}\\
        &= U_1(U_1(\cdots U_1(U_1(\phi)+\Delta_1) \cdots + \Delta_{k-1}) + \Delta_k) +\Delta_{k+1}+ O(\beta\alpha h (k+1)^2 + \beta^2(k+1))
    \end{align*}
\end{proof}
\begin{cor} Asymptotic approximation error of proposed \textbf{continual trajectory shifting} is as follows:
    \begin{align*}
    U_k\left(\phi + \sum_{i=1}^{k-1}\Delta_i\right) &=U_1(\cdots U_1(U_1(\phi)+\Delta_1) \cdots + \Delta_{k-1})  + O(\beta \alpha h k^2 + \beta^2 k)
    \end{align*} for $k\ge2$.
    \label{cor1}
\end{cor}
\begin{proof}
    \begin{align*}
        U_{k}\left(\phi + \sum_{i=1}^{k-1}\Delta_i\right) &=U_1\left(U_{k-1}\left(\phi + \sum_{i=1}^{k-1}\Delta_i\right)\right) \\
        &= U_1(U_1(\cdots U_1(U_1(\phi)+\Delta_1) \cdots) + \Delta_{k-1} + O(\beta\alpha h (k-1)^2 + \beta^2(k-1))) \\
        &\omit\hfill\text{(Theorem \ref{thm2})}\\
        &= U_1(\cdots U_1(U_1(\phi)+\Delta_1) \cdots + \Delta_{k-1})+ O(\beta\alpha h (k-1)^2  + \beta^2(k-1)) \\
        &\qquad+ O(\alpha h (\beta\alpha h (k-1)^2  + \beta^2(k-1)) + (\beta\alpha h (k-1)^2  + \beta^2(k-1))^2) \\
        &\omit\hfill\text{(Theorem \ref{thm1})}\\
        &=U_1(\cdots U_1(U_1(\phi)+\Delta_1) \cdots + \Delta_{k-1}) + O(\beta \alpha h k^2 + \beta^2 k)
    \end{align*}
\end{proof}

\section{Derivation for the momentum optimizer}
In this section, we prove that we can use the same shifting rule even with the momentum optimizer and weight decaying. See Section~\ref{sec:inner_optimizer} for more discussion about the empirical effect of the type of inner-optimizer.
\label{sec:momentum}

\paragraph{Momentum}
Note that the update function of SGD with momentum $\mu$ is:
\begin{align*}
    U_k(\omega) &= U_{k-1}(\omega) - \alpha \cdot g_{k}(\omega) \\
    \text{where } g_k(\omega)&=\mu\cdot g_{k-1}(\omega) + \grad_\omega\mathcal{L}_{k-1}|_{\omega=U_{k-1}(\omega)}.
\end{align*}
Then, the following lemma holds:
\begin{lem}
The approximation of Jacobian is
\[
    \frac{\partial U_k(\phi)}{\partial\phi} = I + O\left(\frac{\alpha hk}{1-\mu}\right)
\]
\end{lem}
\begin{proof}
Let the approximation error of Jacobian at step $k$ be $\epsilon_k$, i.e. \[\frac{\partial U_k(\phi)}{\partial\phi} = I + \epsilon_k\] 
For $k\ge 2$, 
\begin{align*}
    U_k(\phi) &= U_{k-1}(\phi) -\alpha\cdot g_k(\phi) \\
    & = U_{k-1}(\phi) - \alpha\cdot(\mu\cdot g_{k-1}(\phi) + \grad_\phi\mathcal{L}_{k-1}|_{\phi=U_{k-1}(\phi)}) \\
    & = U_{k-1}(\phi) - \mu\cdot\alpha\cdot g_{k-1}(\phi) - \alpha\cdot\grad_\phi\mathcal{L}_{k-1}|_{\phi=U_{k-1}(\phi)} \\
    & = U_{k-1}(\phi) - \mu\cdot\left(U_{k-1}(\phi)-U_{k-2}(\phi)\right) - \alpha \cdot \grad_\phi\mathcal{L}_{k-1}|_{\phi=U_{k-1}(\phi)}
\end{align*}
Then, we compute the Jacobian as
\begin{align*}
    \frac{\partial U_k(\phi)}{\partial \phi} &= \frac{\partial U_{k-1}(\phi)}{\partial \phi} - \mu\cdot\left(\frac{\partial U_{k-1}(\phi)}{\partial \phi}-\frac{\partial U_{k-2}(\phi)}{\partial \phi}\right)- \alpha \cdot\frac{\partial \grad_\phi\mathcal{L}_{k-1}|_{\phi=U_{k-1}(\phi)}}{\partial \phi} \\
    & = (I + \epsilon_{k-1}) + \mu\cdot(\epsilon_{k-1} -\epsilon_{k-2}) - \alpha\cdot\frac{\partial \grad_\phi\mathcal{L}_{k-1}|_{\phi=U_{k-1}(\phi)}}{\partial U_{k-1}(\phi)}\cdot\frac{\partial U_{k-1}(\phi)}{\partial \phi} \\
    & = I + \epsilon_{k-1} + \mu\cdot(\epsilon_{k-1} -\epsilon_{k-2}) - \alpha H_{k-1}\cdot(I+\epsilon_{k-1})
\end{align*}
Thus,
\begin{align*}
    \epsilon_k&=\epsilon_{k-1} + \mu\cdot(\epsilon_{k-1} -\epsilon_{k-2}) + O(\alpha h(1+ \epsilon_{k-1}))
\end{align*}
We can say $\epsilon_0=0$ and for $k=1$,
    \[
        \frac{\partial U_1(\phi)}{\partial \phi} = \frac{\partial (\phi - \alpha \cdot  \grad_\phi\mathcal{L}_{0}|_{\phi=\phi})}{\partial \phi}=I-\alpha H_0 \qquad\rightarrow\qquad
        \epsilon_1 = O(\alpha h)
    \]
    
    Then, 
\begin{align*}
    \epsilon_k - \epsilon_{k-1} &= \mu\cdot(\epsilon_{k-1} -\epsilon_{k-2}) + O(\alpha h (1 + \epsilon_{k-1})) \\
    &= \mu^2\cdot(\epsilon_{k-2} -\epsilon_{k-3}) + O(\alpha h(1+\epsilon_{k-1}+\mu(1+\epsilon_{k-2})))\\ 
    &=\cdots\\
    &= \mu^{k-1}\cdot(\epsilon_{1} -\epsilon_{0}) + O\left(\alpha h\left(\sum_{i=0}^{k-2}\mu^i(1+\epsilon_{k-1-i})\right)\right)\\
    &= O\left(\alpha h\left(\sum_{i=0}^{k-1}\mu^i+\sum_{i=0}^{k-2}\mu^i\epsilon_{k-1-i})\right)\right)
\end{align*}
Since the second term $O(\alpha h\sum_{i=0}^{k-2}\mu^i\epsilon_{k-1-i})$ is quadratic to $\alpha h$, dropping the term,
\begin{align*}
    \epsilon_k &= \epsilon_{k-1} + O\left(\alpha h\sum_{i=0}^{k-1}\mu^i\right)\\
    &=\epsilon_{k-1} + O\left(\alpha h\frac{1-\mu^k}{1-\mu}\right) \\ 
    &=\epsilon_{k-2} + O\left(\alpha h\left(\frac{1-\mu^k}{1-\mu} + \frac{1-\mu^{k-1}}{1-\mu}\right)\right) \\
    &= \cdots \\
    & = \epsilon_{0} + O\left(\alpha h\sum_{i=1}^{k}\frac{1-\mu^i}{1-\mu}\right) \\
    & = O\left(\frac{\alpha h}{1-\mu}\left(k+1 - \frac{1-\mu^{k+1}}{1-\mu}\right)\right)= O\left(\frac{\alpha hk}{1-\mu}\right)
\end{align*}

\end{proof}

Then, as Theorem \ref{thm1}, approximation error between $U_k(\phi+\Delta)$ and $U_k(\phi)+\Delta$ is as follows:

\begin{thm}
For any $\Delta$ that with sufficiently small norm, 
    \[U_k(\phi + \Delta) = U_k(\phi) + \Delta + O\left(\frac{\alpha h k}{1-\mu}\|\Delta\| + \|\Delta\|^2\right)\]

\end{thm}
and as Corollary \ref{cor1}, 
\begin{cor} Asymptotic approximation error of proposed \textbf{Continual Correction} with SGD and momentum $\mu$ is as follows:
    \begin{align*}
    U_k\left(\phi + \sum_{i=1}^{k-1}\Delta_i\right) &=U_1(\cdots U_1(U_1(\phi)+\Delta_1) \cdots + \Delta_{k-1})  + O\left(\frac{\beta \alpha h k^2}{1-\mu} + \beta^2 k\right)
    \end{align*} for $k\ge2$.
\end{cor}

We omit the proofs since only the coefficients are different.

\paragraph{Weight decay}
Denote the update function of SGD with weight decay $\lambda$ by:
\begin{align*}
    U_k(\omega) &= U_{k-1}(\omega) - \alpha \cdot\left(\grad_\omega\left(\mathcal{L}_{k-1}|_{\omega=U_{k-1}(\omega)} + \lambda\|U_{k-1}(\omega)\|^2\right)\right).
\end{align*}
Then, following lemma holds:
\begin{lem}
Approximation of Jacobian is
\[
    \frac{\partial U_k(\phi)}{\partial\phi} = I + O\left(\alpha( h+2\lambda)k\right)
\]
\end{lem}
\begin{proof}
    \begin{align*}
        \frac{\partial U_k(\phi)}{\partial\phi} &= \frac{\partial U_k(\phi)}{\partial U_{k-1}(\phi)} \cdots
        \frac{\partial U_1(\phi)}{\partial\phi} = \prod_{i=0}^{k-1} \left(I-\alpha H_{i}+2\lambda\right)  \\ 
        &= I + O(\alpha (h+2\lambda) k )
    \end{align*}
\end{proof}
Then, as previous,

\begin{cor} Asymptotic approximation error of proposed \textbf{continual trajectory shifting} with SGD and weight decay $\lambda$ is as follows:
    \begin{align*}
    U_k\left(\phi + \sum_{i=1}^{k-1}\Delta_i\right) &=U_1(\cdots U_1(U_1(\phi)+\Delta_1) \cdots + \Delta_{k-1}) + O\left(\beta \alpha (h+2\lambda) k^2 + \beta^2 k\right)
    \end{align*} for $k\ge2$.
\end{cor}



\bibliography{icml2021_supp}
\bibliographystyle{icml2021_supp}